\documentclass{article}

\usepackage{microtype}
\usepackage{graphicx}
\usepackage{booktabs} 

\usepackage{hyperref}
\usepackage[font=small,labelfont=bf]{caption}
\usepackage{subcaption}
\usepackage{wrapfig}
\usepackage{adjustbox}
\usepackage{makecell}
\usepackage{bbm}
\usepackage{rotating}
\usepackage[retainorgcmds]{IEEEtrantools}

\renewcommand{\cite}[1]{\citep{#1}}

\usepackage[accepted]{icml2025}

\usepackage[cmex10]{amsmath}
\usepackage{amssymb}
\usepackage{mathtools}
\usepackage{amsthm}
\usepackage{multirow}
\usepackage{tabularx}

\usepackage[capitalize,noabbrev]{cleveref}

\theoremstyle{plain}
\newtheorem{theorem}{Theorem}[section]
\newtheorem{proposition}[theorem]{Proposition}
\newtheorem{lemma}[theorem]{Lemma}

\theoremstyle{definition}
\newtheorem{definition}[theorem]{Definition}

\theoremstyle{remark}

\usepackage[textsize=tiny]{todonotes}

\icmltitlerunning{On Self-Adaptive Perception Loss Function for Sequential Lossy Compression}

\begin{document}

\twocolumn[
\icmltitle{On Self-Adaptive Perception Loss Function for Sequential Lossy Compression}



\icmlsetsymbol{equal}{*}

\begin{icmlauthorlist}
\icmlauthor{Sadaf Salehkalaibar}{1}
\icmlauthor{Buu Phan}{2}
\icmlauthor{Likun Cai}{2}
\icmlauthor{Joao Atz Dick}{2}
\icmlauthor{Wei Yu}{2}
\icmlauthor{Jun Chen}{3}
\icmlauthor{Ashish Khisti}{2}

\end{icmlauthorlist}

\icmlaffiliation{1}{University of Manitoba}
\icmlaffiliation{2}{University of Toronto}
\icmlaffiliation{3}{McMaster University}

\icmlcorrespondingauthor{Sadaf Salehkalaibar}{Sadaf.Salehkalaibar@umanitoba.ca}

\icmlkeywords{Machine Learning, ICML}

\vskip 0.3in
]



\printAffiliationsAndNotice{}  

\begin{abstract}
We consider causal, low-latency, sequential lossy compression, with mean squared-error (MSE) as the distortion loss, and a perception loss function (PLF) to enhance the realism of reconstructions. 
As the main contribution, we propose and analyze a new PLF that considers the joint distribution between the current source frame and the previous reconstructions. We establish the theoretical rate-distortion-perception function for first-order Markov sources and analyze the Gaussian model in detail. From a qualitative perspective,  the proposed metric can simultaneously avoid the error-permanence phenomenon and also better exploit the temporal correlation between high-quality reconstructions. The proposed metric is referred to as self-adaptive perception loss function (PLF-SA), as its behavior adapts to the quality of reconstructed frames. We provide a detailed comparison of the proposed perception loss function with previous approaches through both information theoretic analysis as well as experiments involving moving MNIST and UVG datasets.

\end{abstract}

\section{Introduction}

In recent years, the topic of lossy compression for videos has received significant attention, driven by the growing demand for producing visually appealing reconstructions even at lower bitrates. Early versions of compression algorithms relied on distortion measures, e.g., mean squared error (MSE), MS-SSIM \cite{SSIM, PSNR3, PSNR4} and PSNR \cite{PSNR1, PSNR2, PSNR3, PSNR4}. However, these metrics often resulted in outputs that were perceived as blurry and lacking \emph{realism}. Consequently, there have been efforts to incorporate \emph{perception}-based loss functions into compression systems to improve visual quality. These loss functions aim to quantify the divergence between the distributions of the source and the reconstruction, where achieving \emph{perfect} perceptual quality means that the two distributions match with each other. Blau and Michaeli \cite{blau2019rethinking} explored the rate-distortion-perception (RDP) tradeoff from a theoretical perspective. Subsequently, Zhang et al. \cite{Jun-Ashish2021} introduced universal representations, wherein the representation remains fixed during encoding, and only the decoder can be adjusted to attain near-optimal performance.

With the multitude of frames in a video, there is no unique perception loss function (PLF) that is suitable in all cases. In fact, at least two different PLFs have been proposed in prior work. One choice is a PLF metric based on the framewise marginal distributions (FMD) of the source and reconstruction \cite{video1}, where the perception loss function only preserves the marginal distribution of the reconstructed frames instead of the joint distribution. In contrast to this choice, other works \cite{video-joint} have proposed PLF based on the entire joint distribution (JD) of source frames. A recent study \cite{Jun-Ashish2023} establishes the rate-distortion-perception (RDP) trade-off for both metrics. It is shown that at low bitrates, PLF-JD encounters \emph{error permanence phenomenon}, where errors propagate across all future reconstructions, leaving distortion unchanged across frames. On the other hand, at higher bit rates PLF-JD is more desirable, as PLF-FMD does not address temporal consistency between frames.

\begin{figure*}[ht]
    \centering
    \begin{subfigure}{0.29\linewidth}
        \setlength{\lineskip}{1pt}
        \centering
        \includegraphics[width=.95\linewidth]{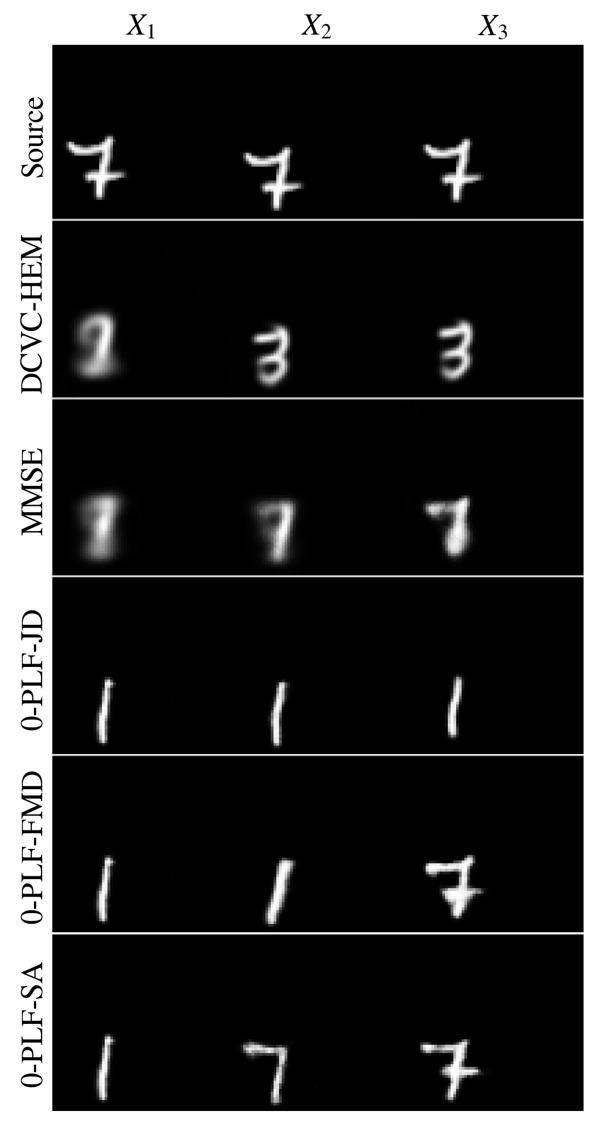}
        \captionsetup{width=0.9\linewidth}
        \subcaption{}
    \end{subfigure}
    \begin{subfigure}{0.29\linewidth}
        \setlength{\lineskip}{1pt}
        \centering
        \includegraphics[width=.95\linewidth]{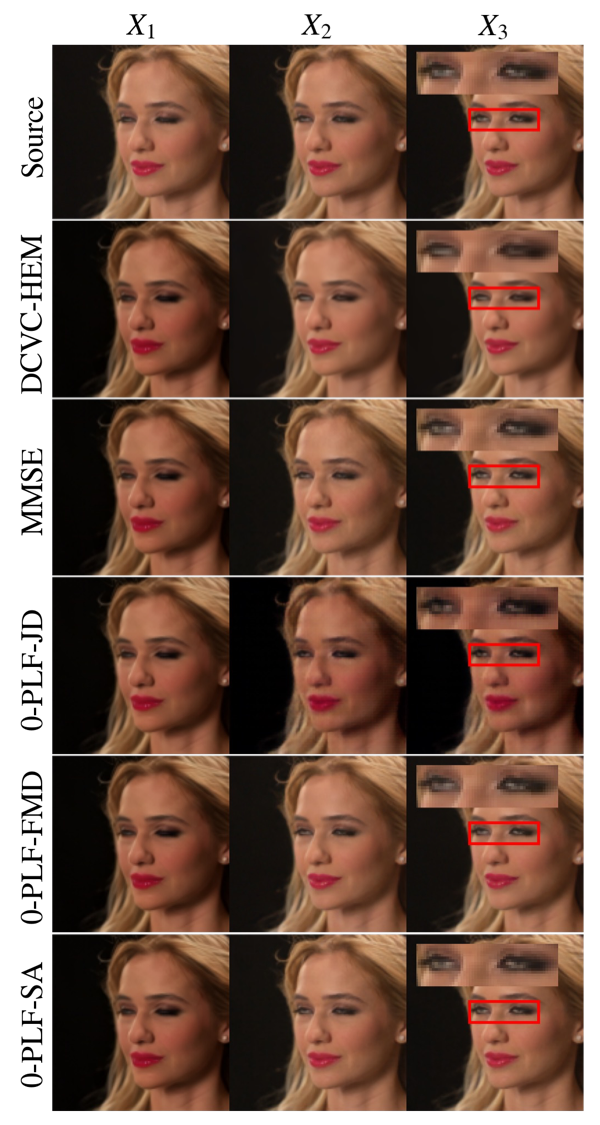}
        \captionsetup{width=0.9\linewidth}
        \subcaption{ }
    \end{subfigure}
    \begin{subfigure}{0.373\linewidth}
        \centering
        \includegraphics[width=.95\linewidth]{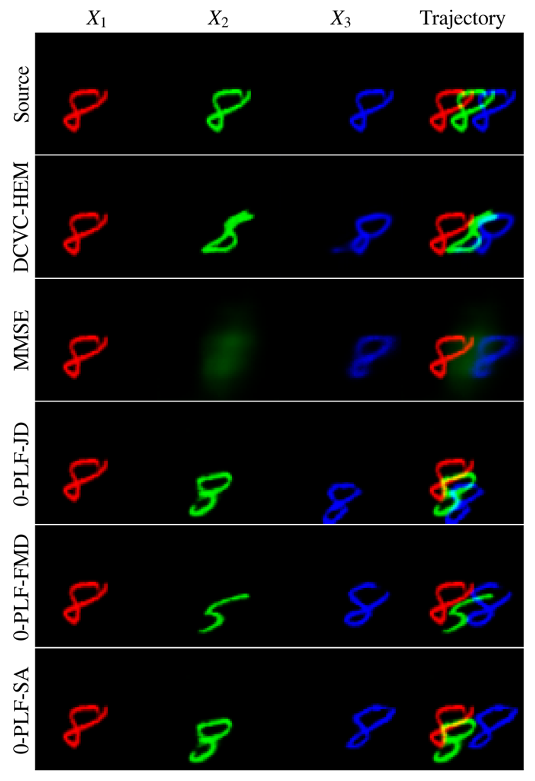}
        \captionsetup{width=0.9\linewidth}
        \subcaption{}
        \label{fig:mnist_2_fast}
    \end{subfigure}
    \vspace{-1.2em}
    \caption{
    \textbf{(a)} Outputs for MovingMNIST with the first frame compressed at a low bitrate $R_1 = 12$ bits. PLF-SA and PLF-FMD recover from previous errors, while PLF-JD and DCVC-HEM exhibit error permanence.
    \textbf{(b)} Outputs for UVG with the first frame compressed at a low bitrate $R_1 = 0.144$ bpp. PLF-SA and PLF-FMD maintain color tone, whereas PLF-JD propagates color tone errors. DCVC-HEM struggles to reconstruct details like eye pupils, while PLF models perform better.
    \textbf{(c)} Outputs for MovingMNIST with the first frame compressed at a high bitrate $R_1 = \infty$ bits. PLF-FMD produces reconstruction error without maintaining the temporal correlation. PLF-JD propagates the trajectory error while PLF-SA rectifies the error preserves the temporal correlation across different frames.
    }
    \label{fig:mnist_1aa}
    \vspace{-1.5em}
\end{figure*}

In this work, we study causal, low-latency, sequential video compression when the output is subjected to both a mean squared-error (MSE) distortion loss and a new perception loss function, which we refer to as \emph{Self-Adaptive (SA)}. Our proposed metric (PLF-SA) considers the joint distribution between the current source frame and the previous reconstructions.  We establish the rate-distortion-perception function for first-order Markov sources and analyze the Gaussian source in detail for our proposed PLF. We also present experimental results involving moving MNIST and UVG datasets. Our key observation is that our proposed PLF mitigates the disadvantages of previously proposed metrics: 1) when the previous reconstructions are of lower quality our proposed PLF does not suffer from the error permanence phenomenon observed with PLF-JD; 2) when the previous reconstructions are of higher quality our proposed PLF preserves the joint distribution with these frames and yields better temporal consistency than PLF-FMD. We summarize these below: 
\begin{itemize}
\item \emph{Resilience to Error Permanence Phenomenon}: Using theoretical analysis of the rate-distortion-perception function of first-order Gauss-Markov sources and through experiments (see, e.g., Fig.~\ref{fig:mnist_1aa}a and Fig.~~\ref{fig:mnist_1aa}b), we demonstrate that PLF-SA does not suffer from the error permanence phenomenon. In particular, when the first source frame is compressed at a low bitrate, PLF-JD fails to correct mistakes appearing in this frame in subsequent reconstructions. PLF-SA does not suffer from this effect.

\item \emph{Sensitivity to Temporal Correlation across Frames}: 
Through both theoretical analysis and experimental findings (see Fig.~~\ref{fig:mnist_1aa}c), we demonstrate that our proposed PLF-SA metric can better exploit temporal correlation across frames to yield improved reconstruction. In this setting we assume that the first frame is compressed at a higher bitrate while the second frame is compressed at a lower bitrate. We note that while PLF-FMD yields incorrect output in the second frame, PLF-SA is able to exploit the temporal correlation with the first frame to output the correct digit in the second frame. We also note that PLF-JD still suffers from error permanence as observed in the incorrect trajectory in the reconstruction of the third frame in Fig.~\ref{fig:mnist_1aa}c;  PLF-SA also does not suffer from this effect.
\end{itemize}

PLF-SA does not suffer from the error permanence phenomenon at low bitrates and maintains temporal correlation among frames, especially when the first frame undergoes high-rate compression. Consequently, it takes advantage of both metrics (PLF-FMD or PLF-JD), depending on the operating rate regime. This adaptability to varying rates is the rationale behind naming this PLF as Self-Adaptive.

\section{System Model}
\label{system_model}

\begin{figure}[t]
    \centering
    \includegraphics[scale=0.25]{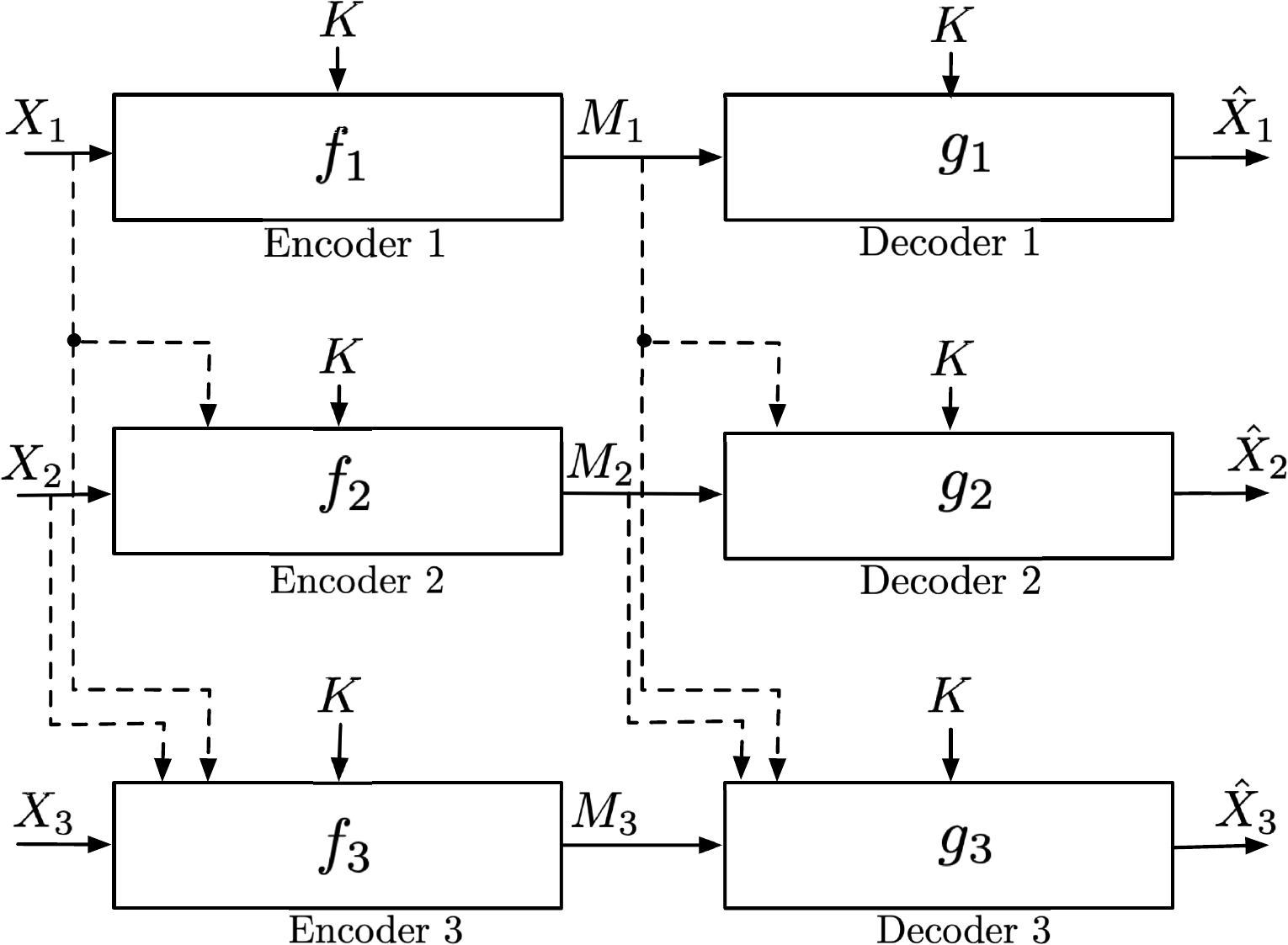}
    \caption{System model for a sequential lossy compression.}
    \label{fig:system}
    \vspace{-0.5cm}
    \label{fig:sys}
\end{figure}

Assume that we have $T$ frames of video denoted by $(X_1,\ldots,X_T)\in\mathcal{X}_1\times\ldots\times \mathcal{X}_T$ (where $\mathcal{X}_i \subseteq {\mathbbm{R}}^d$) distributed according to joint distribution $P_{X_1\ldots X_T}$. The encoders and decoders have access to a shared common randomness $K\in\mathcal{K}$. The (possibly stochastic) $j$th encoding function gets the sources $(X_1,\ldots,X_j)$ and the key $K$ and outputs a variable length message $M_{j} \in \mathcal{M}_j(=\{0,1\}^\star)$, i.e., 
\begin{IEEEeqnarray}{rCl}
f_{j}&\colon& \mathcal{X}_1\times \ldots \times \mathcal{X}_j\times \mathcal{K}\to \mathcal{M}_{j},\qquad j=1,\ldots,T.
\end{IEEEeqnarray}
The $j$th decoding function receives the messages $(M_1,\ldots,M_j)$ and using the key $K$, it outputs a reconstruction $\hat{X}_j \in \hat{\mathcal{X}}_j (\subseteq {\mathbbm{R}}^d)$, i.e.,
\begin{IEEEeqnarray}{rCl}
g_{j} &\colon& \mathcal{M}_{1}\times \mathcal{M}_{2}\times \ldots \times \mathcal{M}_{j}\times \mathcal{K}\to \hat{\mathcal{X}}_j.\;\;\label{decoding-function}
\end{IEEEeqnarray}
The mappings $\{f_{j}\}_{j=1}^T$ and $\{g_{j}\}_{j=1}^T$ induce the conditional distribution $P_{\hat{X}_1\ldots \hat{X}_T|X_1\ldots X_T}$ for the reconstructed video given the original video. The proposed framework illustrated in Fig.~\ref{fig:sys} is a \emph{one-shot} setting i.e., a single sample of the source is compressed at a time. 

The reconstruction of each frame $j$ should satisfy a certain distortion from the source where the metric is assumed to be the mean squared-error (MSE) function i.e. $d(x_j, \hat{x}_j) = ||x_j-\hat{x}_j||^2$, which is widely used in many applications. From the perceptual perspective, for given probability distributions $P_{\hat{X}_{1}\ldots\hat{X}_{j-1} X_{j}}$ and $P_{\hat{X}_{1}\ldots\hat{X}_{j-1}\hat{X}_{j}}$, let $\phi_j(P_{\hat{X}_{1}\ldots\hat{X}_{j-1} X_{j}},P_{\hat{X}_{1}\ldots\hat{X}_{j-1}\hat{X}_{j}})$ denote the perception metric capturing the divergence between them. We call this PLF as \emph{self-adaptive (SA)}. If $\phi_j(P_{\hat{X}_{1}\ldots\hat{X}_{j-1} X_{j}},P_{\hat{X}_{1}\ldots\hat{X}_{j-1}\hat{X}_{j}})=0$, we get 
\begin{IEEEeqnarray}{rCl}
P_{\hat{X}_{1}\ldots\hat{X}_{j-1} X_{j}}=P_{\hat{X}_{1}\ldots\hat{X}_{j-1}\hat{X}_{j}},\qquad j=1,\ldots,T,\label{percep-joint}
\end{IEEEeqnarray}
which is called as \emph{self-adaptive zero-perception loss ($0$-PLF-SA)}. In the following, we define two other perception metrics which are extensively used in many works. For given probability distributions $P_{X_{1}\ldots X_{j}}$ and $P_{\hat{X}_{1}\ldots\hat{X}_{j}}$,  let $\xi_j(P_{X_{1}\ldots X_{j}},P_{\hat{X}_{1}\ldots\hat{X}_{j}})$ be called as \emph{perception loss function based on joint distribution (PLF-JD)}. Alternatively, the \emph{perception loss function based on framewise marginal distribution (PLF-FMD)} is shown by $\psi_j(P_{X_j},P_{\hat{X}_j})$. Notice that $0$-PLF-JD and $0$-PLF-FMD imply that $P_{X_1\ldots X_j}=P_{\hat{X}_1\ldots \hat{X}_j}$ and $P_{X_j}=P_{\hat{X}_j}$ for $j=1,\ldots,T$, respectively.
\begin{definition}[Operational RDP region] An RDP tuple $(\mathsf{R},\mathsf{D},\mathsf{P})$ is said to be achievable for the one-shot setting if there exist encoders and decoders such that:
\begin{IEEEeqnarray}{rCl}
 \mathbbm{E}[\ell(M_{j})] &\leq & R_j, \\
\mathbbm{E}[\|X_j-\hat{X}_j\|^2] &\leq & D_j, \\
\phi_j(P_{\hat{X}_{1}\ldots\hat{X}_{j-1} X_{j}},P_{\hat{X}_{1}\ldots\hat{X}_{j-1}\hat{X}_{j}})&\leq & P_j,\;\; j=1,2,3,\label{perception-condition}
\end{IEEEeqnarray}
where $\ell(M_{j})$ denotes the length of the message $M_{j}$. The operational RDP region, denoted by $\mathcal{RDP}^o$, is the closure of the set of all achievable tuples. Moreover, for a given $(\mathsf{D},\mathsf{P})$, the operational rate region, denoted by $\mathcal{R}^o(\mathsf{D},\mathsf{P})$, is the closure of the set of all tuples $\mathsf{R}$ such that $(\mathsf{R},\mathsf{D},\mathsf{P})\in \mathcal{RDP}^o$.
\end{definition}

We consider Gauss-Markov sources as follows. We assume that $X_1 \sim {\mathcal N}(0, \sigma^2)$ for some $\sigma^2>0$,
\begin{IEEEeqnarray}{rCl}
X_2 = \rho X_1+N_1, \qquad
X_3 = \rho X_2+N_2,\label{Gaus-def2}
\end{IEEEeqnarray}
for some $0\leq \rho\leq 1$, where $N_j$ is independent of $X_j$ with mean zero and variance $(1-\rho^2)\sigma^2$ for $j=1,2$.    The model extends naturally to the case of $T$ time-steps.  We assume that the perception metric is Wasserstein-2 distance, i.e., \begin{IEEEeqnarray}{rCl}&&\hspace{-1cm}\phi_j(P_{\hat{X}_{1}\ldots\hat{X}_{j-1} X_{j}},P_{\hat{X}_{1}\ldots\hat{X}_{j-1}\hat{X}_{j}}):=\nonumber\\&&W_2^2(P_{\hat{X}_{1}\ldots\hat{X}_{j-1} X_{j}},P_{\hat{X}_{1}\ldots\hat{X}_{j-1}\hat{X}_{j}}).
\end{IEEEeqnarray}

\begin{table*}[t]
\renewcommand{\arraystretch}{2.0}
\centering
\caption{Achievable reconstructions and distortions for $R_1=\epsilon$ and an arbitrary nonnegative $R_2$. }
\label{table-ach-recons-low-rate}
\begin{center}
\begin{tiny}
\begin{tabular}{|l|l|l|l|}
\hline
\multirow{2}*{ } & \multicolumn{3}{c|}{Second Frame: $\hat{X}_2=\omega_1\hat{X}_1+\omega_2X_2+Z_2$} \\
\cline{2-4}
~ & Coefficients $\omega_1$, $\omega_2$ & Distortion $D^0_2$ & $Z_2$ \\
\hline 
$0$-PLF-FMD & \makecell[l]{$\omega_1=\frac{\sqrt{2\epsilon\ln 2}}{\sqrt{1-2^{-2R_2}+2\epsilon\ln 2}}$, \\ $\omega_2=\frac{1-2^{-2R_2}}{\sqrt{1-2^{-2R_2}+ (2\epsilon\ln 2)}}$ } & 
\makecell[l]{$2\sigma^2 (1-\sqrt{1-2^{-2R_2}+\rho^22\epsilon\ln 2})+O(\epsilon)$ \\ (Appendix~\ref{FMD-low-rate}) } & 
$\mathcal{N}(0,(1-\omega_1^2-\omega_2^2-2\omega_1\omega_2\sqrt{2\epsilon\ln 2})\sigma^2)$\\
\hline
$0$-PLF-SA & \makecell[l]{$\omega_1=\sqrt{2\epsilon\ln 2}(1-\sqrt{1-2^{-2R_2}})$, \\ $\omega_2=\sqrt{1-2^{-2R_2}}X_2$ } & 
\makecell[l]{$2\sigma^2 (1-\sqrt{1-2^{-2R_2}})+O(\sqrt{\epsilon})$ \\ (Appendix~\ref{CP-low-rate}) } & 
$\mathcal{N}(0,(2^{-2R_2}-(1-\sqrt{1-2^{-2R_2}})^2(2\epsilon\ln 2))\sigma^2)$ \\
\hline
$0$-PLF-JD & $\omega_1=1$,\;\;$\omega_2=0$ & $2\sigma^2 (1-\sqrt{2\epsilon \ln 2})=D_1$ (Appendix~\ref{JD-low-rate}) &  $0$\\
\hline 
\end{tabular}
\end{tiny}
\end{center}
\vspace{-1.5em}
\end{table*}

\section{RDP Regions}

In this section, we provide an approximation for the operational RDP region and then analyze it for Gauss-Markov source model. In general, it is not feasible to compute the region $\mathcal{RDP}^o$ directly since it involves searching over all possible encoding-decoding functions. But, for first-order Markov sources where the Markov chain $X_1\to X_2\to X_3$ holds, the following region can be used as an approximation. 

\begin{definition}[Information  RDP Region]\label{iRDP-region} For first-order Markov sources, let the information RDP region, denoted by $\mathcal{RDP}$, be the set of all tuples $(\mathsf{R},\mathsf{D},\mathsf{P})$ which satisfy the following
\begin{IEEEeqnarray}{rCl}
R_1 &\geq & I(X_1;X_{r,1}),\\
 R_2 &\geq&  I(X_2;X_{r,2}|X_{r,1}),\\
R_3 &\geq&  (X_3;X_{r,3}|X_{r,1},X_{r,2}) \label{rate1-3}\\
D_j &\geq & \mathbbm{E}[\|X_j-\hat{X}_j\|^2], \\
P_j&\geq & \phi_j(P_{\hat{X}_{1}\ldots\hat{X}_{j-1} X_{j}}, P_{\hat{X}_{1}\ldots\hat{X}_{j-1}\hat{X}_{j}}), \qquad j=1,2,3,\nonumber\\\label{perception}
\end{IEEEeqnarray}
for auxiliary random variables $(X_{r,1},X_{r,2},X_{r,3})$ and $(\hat{X}_1,\hat{X}_2,\hat{X}_3)$ satisfying the following
\begin{IEEEeqnarray}{rCl}
&&\hat{X}_1= \eta_1(X_{r,1}), \;\;\hat{X}_2=\eta_2(X_{r,1},X_{r,2}),\;\; \hat{X}_3=X_{r,3},\nonumber\\\label{function-condition}\\
&&X_{r,1}\to X_1\to (X_2,X_3),\\&& X_{r,2}\to (X_2,X_{r,1})\to (X_1,X_3),\\&& X_{r,3}\to (X_3,X_{r,1},X_{r,2})\to (X_1,X_2),\label{Markov-conditions3}
\end{IEEEeqnarray}
for some deterministic functions $\eta_1(.)$ and $\eta_2(.,.)$. Moreover, for a given $(\mathsf{D},\mathsf{P})$, the information rate region, denoted by $\mathcal{R}(\mathsf{D},\mathsf{P})$, is the closure of the set of all tuples $\mathsf{R}$ that $(\mathsf{R},\mathsf{D},\mathsf{P})\in \mathcal{RDP}$. 
\end{definition}
The following theorem provides upper and lower bounds on the operational RDP region.
\begin{theorem}
For first-order Markov sources, a given $(\mathsf{D},\mathsf{P})$ and $\mathsf{R}\in \mathcal{R}(\mathsf{D},\mathsf{P})$, we have
\begin{IEEEeqnarray}{rCl}
\mathsf{R}+\log(\mathsf{R}+1)+5\in \mathcal{R}^{o}(\mathsf{D},\mathsf{P}) \subseteq \mathcal{R}(\mathsf{D},\mathsf{P}).\label{1-shot-thm-inner}
\end{IEEEeqnarray}
\end{theorem}
\begin{IEEEproof} This statement can be proved using similar lines to the proof of Theorem 3 in \cite{Jun-Ashish2023} which was originally proposed for PLF-JD and PLF-FMD. The proof for PLF-SA is provided in Appendix~\ref{operational-RDP-app} for completeness.
\end{IEEEproof}
Thus, for sufficiently large rates, we can approximate $\mathcal{R}^{o}(\mathsf{D},\mathsf{P})$ by $\mathcal{R}(\mathsf{D},\mathsf{P})$.  

In the following, we analyze the RDP region $\mathcal{R}(\mathsf{D},\mathsf{P})$ for Gauss-Markov source model. First, we show that one can restrict to jointly Gaussian distribution over reconstructions and sources, without loss of optimality.

\begin{theorem} 
\label{thm:rdp-gauss}
For the Gauss-Markov source model, any  tuple $(\mathsf{R},\mathsf{D},\mathsf{P})\in \mathcal{RDP}$ can be achieved by a jointly Gaussian distribution over $(X_{r,1},X_{r,2},X_{r,3})$ and identity functions for $\eta_j(\cdot)$. That is, for the Gauss-Markov source model, the $\mathcal{RDP}$ region of Definition~\ref{iRDP-region} simplifies to the set of all $(\mathsf{R},\mathsf{D},\mathsf{P})$ tuples such that 
\begin{subequations}\label{RDP-Gauss-Markov}
\begin{IEEEeqnarray}{rCl}
R_1 &\geq&  I(X_1;\hat{X}_1),\\
 R_2 &\geq&  I(X_2;\hat{X}_{2}|\hat{X}_{1}),\\
R_3 &\geq&  I(X_3;\hat{X}_{3}|\hat{X}_{1},\hat{X}_{2}) \label{R-G}\\
D_j &\geq & \mathbbm{E}[\|X_j-\hat{X}_j\|^2], \\
P_j&\geq & \phi_j(P_{\hat{X}_{1}\ldots\hat{X}_{j-1} X_{j}}, P_{\hat{X}_{1}\ldots\hat{X}_{j-1}\hat{X}_{j}}), \qquad j=1,2,3,\nonumber\\\label{perception}
\end{IEEEeqnarray}
\end{subequations}
for some auxiliary random variables $(\hat{X}_1,\hat{X}_2,\hat{X}_3)$ which satisfy the following Markov chains
\begin{IEEEeqnarray}{rCl}
\hat{X}_{1}&\to& X_1\to (X_2,X_3),\\\ \hat{X}_{2}&\to& (X_2,\hat{X}_{1})\to (X_1,X_3),\\ \hat{X}_{3}&\to& (X_3,\hat{X}_{1},\hat{X}_{2})\to (X_1,X_2).\;\;\;\label{Markov-G}
\end{IEEEeqnarray}
\end{theorem}

\begin{IEEEproof} The proof uses similar lines to the proof of Theorem 4 in \cite{Jun-Ashish2023}. It is provided in Appendix~\ref{Gauss-Markov-app} for completeness.
\end{IEEEproof}

We next discuss various insights from the analysis of the RDP region in Theorem~\ref{thm:rdp-gauss}. We will often consider asymptotic regimes as follows. When we set the compression rate $R_j = \epsilon$, it will indicate a low-rate regime, i.e., we will assume that $\epsilon>0$ is a small constant. When we refer to high-rate compression we will assume that $R_j \rightarrow \infty$. 

\section{Distortion Analysis for Gauss-Markov Sources and Zero-Perception Loss}
\label{sec:distortion_analysis}

In this section, we present practical insights from analyzing the Gauss-Markov source model when we have a zero-perception loss. We study two criteria on different PLFs: resilience to error permanence phenomenon and sensitivity to temporal correlation across frames.

\begin{table*}[t]
\renewcommand{\arraystretch}{1.0}
\centering
\caption{Achievable reconstructions and distortions for $R_1,R_3\to\infty$ and $R_2=\epsilon$. }
\vspace{-0.5em}
\label{table-ach-recons-inf-eps-inf}
\begin{center}
\begin{tiny}
\begin{sc}
\begin{tabular}{|l|l|l|}
\hline
& Second Step  & Third Step  \\
\hline 
$0$-PLF-FMD & $\hat{X}_2=(1-O(\epsilon))\hat{X}_1+O(\epsilon)X_2+Z_{2,\text{FMD}}$ & $\hat{X}_3= X_3$ \hspace{4.2cm}(Appendix~\ref{FMd-high-R1-low-R23})\\
(\!$\sqrt{\epsilon}\!\ll\!\! \rho\!\! <\! 1$\!)& $Z_{2,\text{FMD}}\sim\mathcal{N}(0,O(\epsilon)\sigma^2)$ & \\
& $D_{2,\text{FMD}}^{\infty}=2(1-\rho-O(\epsilon))\sigma^2$\qquad Table 2 in \cite{Jun-Ashish2023} & \\
\hline 
$0$-PLF-FMD & $\hat{X}_2= O(\sqrt{\epsilon})X_2+Z'_{2,\text{FMD}}$ & $\hat{X}_3=X_3$\hspace{4.2cm} (Appendix~\ref{FMd-high-R1-low-R23})\\
($0\!<\!\! \rho\!\! \ll \! \sqrt{\epsilon}$) & $Z'_{2,\text{FMD}}\sim\mathcal{N}(0,(1-O(\epsilon))\sigma^2)$ & \\
& $D_{2,\text{FMD}}^{\infty}=2\sigma^2(1-O(\sqrt{\epsilon}))$\hspace{3.2cm} (Appendix~\ref{FMd-high-R1-low-R23}) &  \qquad\qquad \\
\hline
$0$-PLF-JD & $\hat{X}_2=(\rho-O(\sqrt{\epsilon}))\hat{X}_1+O(\sqrt{\epsilon})X_2+Z_{2,\text{JD}}$ & $\hat{X}_3=(\rho-O(\sqrt{\epsilon})) \hat{X}_2+\frac{1}{\sqrt{1+\rho^2}}(\rho N_1+N_2+O(\sqrt{\epsilon}))+Z_{3,\text{JD}}$\\
&$Z_{2,\text{JD}}\sim \mathcal{N}(0,(1-\rho^2+O(\epsilon))\sigma^2)$ & $Z_{3,\text{JD}}\sim \mathcal{N}(0,O(\sqrt{\epsilon})\sigma^2)$\\
&$D_{2,\text{JD}}^{\infty}= 2\sigma^2(1-\rho^2-O(\sqrt{\epsilon}))$\qquad  Table 2 in \cite{Jun-Ashish2023} & $D_{3,\text{JD}}^{\infty}= 2\sigma^2\left(1-\rho^4\right)\left(1-\frac{1}{\sqrt{1+\rho^2}}\right)+O(\sqrt{\epsilon})$\;\; (Appendix~\ref{JD-high-R1-low-R23})\\
\hline
$0$-PLF-SA  & $\hat{X}_2=(\rho-O(\sqrt{\epsilon}))\hat{X}_1+O(\sqrt{\epsilon})X_2+Z_{2,\text{SA}}$ & $\hat{X}_3=X_3$ \hspace{4.2cm} (Appendix~\ref{CP-high-R1-low-R23})\\
& $Z_{2,\text{SA}}=Z_{2,\text{JD}}$ & \\
& $D_{2,\text{SA}}^{\infty}= D^{\infty}_{2,\text{JD}}$\hspace{4.5cm} (Appendix~\ref{CP-high-R1-low-R23}) &  \\
\hline
\end{tabular}
\end{sc}
\end{tiny}
\end{center}
\vspace{-1.2em}
\end{table*}

\subsection{Resilience to Error Permanence Phenomenon}
\label{low_rate}

In this section, we analyze the Gauss-Markov model to investigate resilience of different PLFs to error permanence phenomenon, initially identified in \cite{Jun-Ashish2023}. In our analysis, we assume that the first frame is compressed at a low rate, i.e., $R_1=\epsilon$ for sufficiently small $\epsilon>0$. The rates of the second and third steps, $R_2$ and $R_3$, can take on any nonnegative values. For simplicity, assume the case of $\rho=1$, i.e., $X_2=X_1$ where the error permanence phenomenon can be clearly demonstrated for PLF-JD (the results for other values of $\rho$ please see Appendix~\ref{low-rate-app}).

We note that reconstructions of the first frame in all cases are identical. Using standard analysis of the rate-distortion-perception function for Gaussian sources, we have that when $R_1 = \epsilon$, the reconstruction is given by: $\hat{X}_1{=}\sqrt{2\epsilon\ln 2}X_1{+}Z_1$ where $Z_1{\sim} \mathcal{N}(0,(1{-}2\epsilon\ln 2)\sigma^2)$ is independent of $X_1$ and the resulting distortion is given by $D_1{=}2(1{-}\sqrt{2\epsilon\ln 2})\sigma^2$.

For the second step, the achievable reconstructions of different $0$-PLFs are shown in Table~\ref{table-ach-recons-low-rate}.  Most strikingly, we note that regardless of the value of $R_2$, the reconstruction of $0$-PLF-JD  is of the form $\hat{X}_2=\hat{X}_1$ when $\rho=1$. Intuitively once $\hat{X}_1$ is generated and $\rho=1$, the PLF-JD metric forces all the future reconstructions to be an identical copy and ignore any new information available to the decoder. This is referred to as the error-permanence phenomenon~\cite{Jun-Ashish2023}. In contrast note that the reconstructions associated with PLF-SA and PLF-FMD are of the form: $\hat{X}_2 = \omega_1 \hat{X_1} + \omega_2 X_2 + Z_2$, where the coefficients $\omega_1$ and $\omega_2$ are stated in Table~\ref{table-ach-recons-low-rate}. In this case the reconstruction $\hat{X}_2$ incorporates new information available to the decoder in the second step as reflected in the coefficient $\omega_2$. Interestingly, as $\epsilon$ approaches $0$, the coefficient $\omega_2$ in both cases becomes identical, indicating that both metrics capture similar information.  

Although our discussion above is limited to the case when the compression rate of the first frame is very small, similar conclusions also appear to hold for moderate compression rates. We illustrate this behavior numerically in Fig.~\ref{fig:RD2}
in Appendix ~\ref{low-rate-app}. In particular for $R_1=0.1$ and $R_2 \ge 0.05$, the distortion of the second frame for $0$-PLF-SA outperforms that of $0$-PLF-JD. We  also discuss the reconstruction associated with the third frame in the same Appendix. While, by design, PLF-FMD achieves a lower distortion than PLF-SA, it does not always output the most deisered reconstructions. As discussed in the next section, PLF-FMD fails to effectively preserve temporal correlation across frames. 

\begin{figure*}[h]
    \centering
    \includegraphics[width=0.253\linewidth]{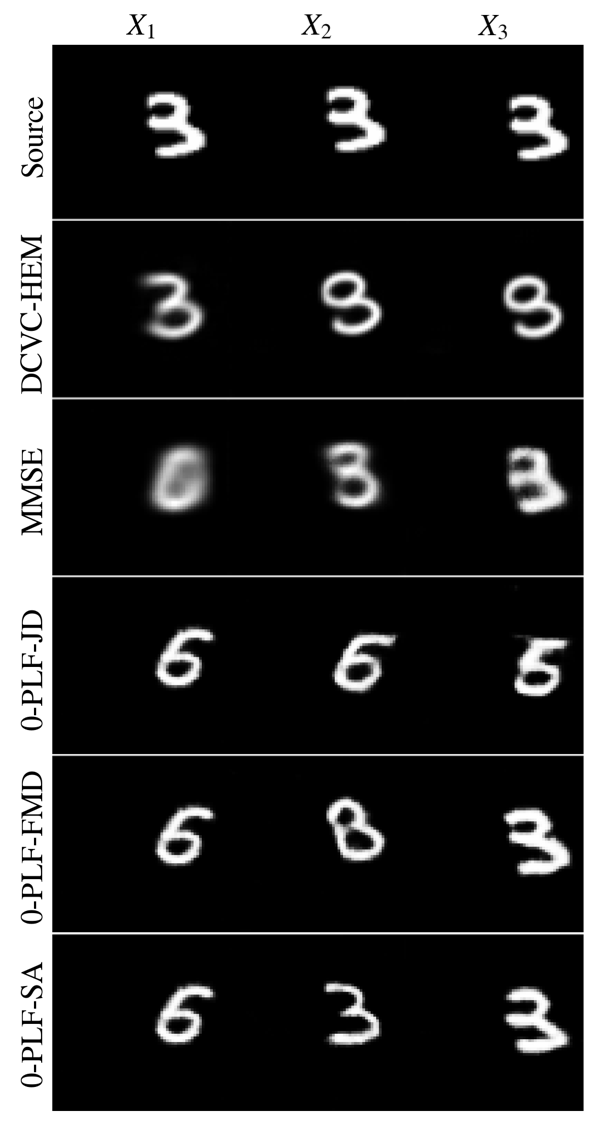}
    \includegraphics[width=0.232\linewidth]{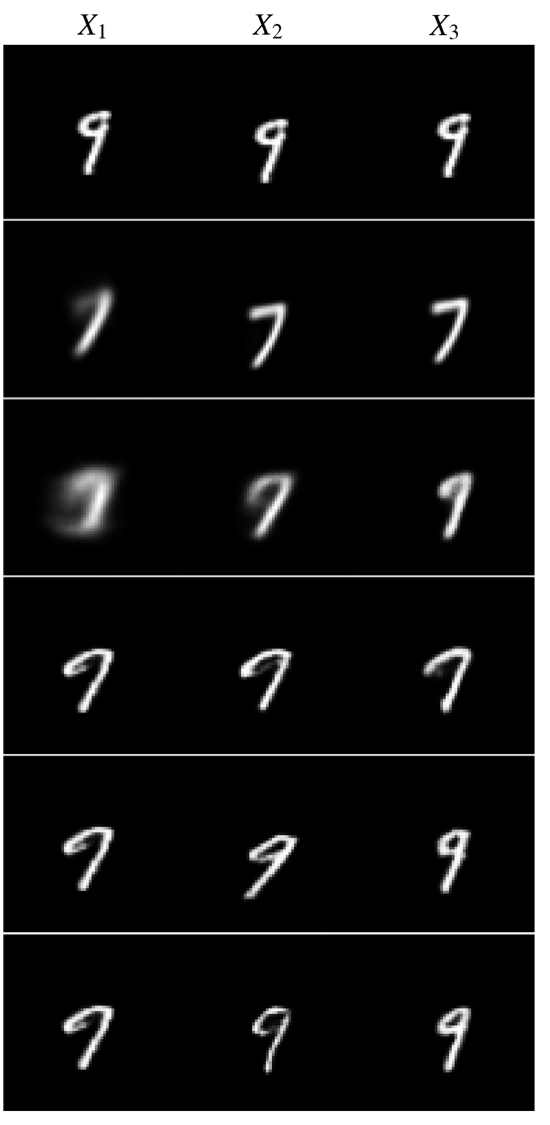}
    \includegraphics[width=0.232\linewidth]{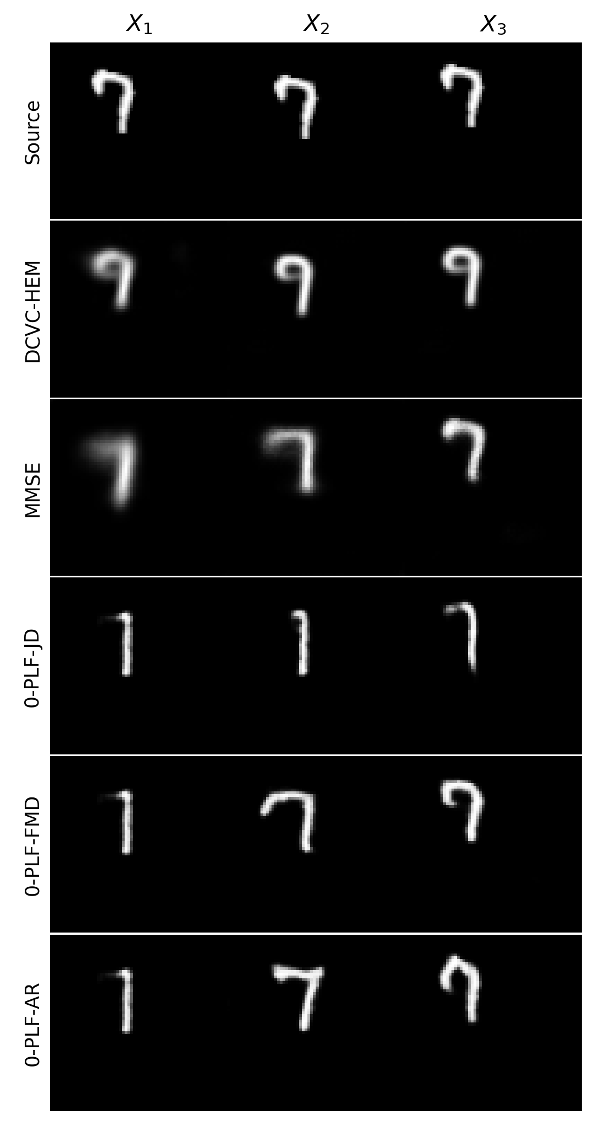}
    \includegraphics[width=0.232\linewidth]{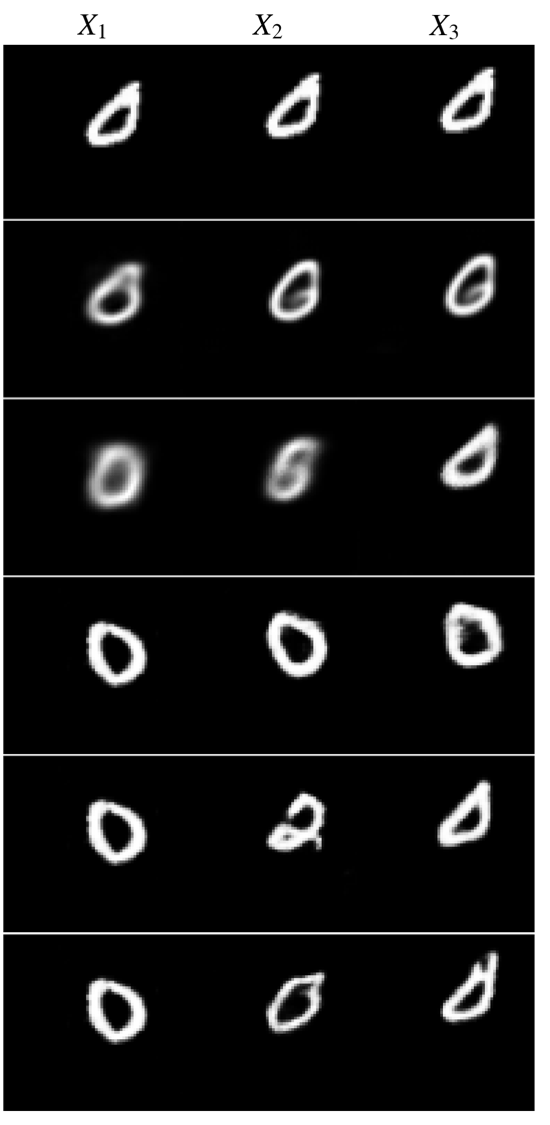}
    \caption{The reconstruction results on the MovingMNIST dataset when the first frame is compressed at a low rate $R_1=12$ bits. 
    Similar to the Guass-Markov case presented in Section~\ref{low_rate}, both PLF-SA and PLF-FMD demonstrate resilience to prior errors (digit contour errors) by incorporating new information from $X_2$ and $X_3$, while PLF-JD suffers from error permanence phenomenon as it tends to ignore new information. DCVC-HEM exhibits a comparable tendency for error permanence.
    }
    \label{fig:appendix_3}
    \vspace{-0.2cm}
\end{figure*}

\subsection{Sensitivity to Temporal Correlation Across Frames}
\label{high_rate}

In this section, we discuss that the choice of PLF affects the temporal correlation across different frames. Specifically, we consider the case where the first and thirds frames are compressed at a high rate, i.e., $R_1,R_3\to\infty$, and the rate of the second frame is small enough, i.e., $R_2=\epsilon$ for a sufficiently small $\epsilon>0$. In order to develop a full qualitative understanding, we also consider the case of case of $R_2=R_3=\epsilon$. This case  is more involved and discussed in Appendix~\ref{high-rate-app}. In the first step, the high rate assumption implies that $\hat{X}_1=X_1$. The achievable reconstructions of all $0$-PLFs for the second and third steps are discussed in the following and summarized in Table~\ref{table-ach-recons-inf-eps-inf}.  

\textit{Achievable Reconstructions of $0$-PLF-FMD}:

\underline{\textit{Large Correlation Coefficient}}: As it can be observed from the first row of Table~\ref{table-ach-recons-inf-eps-inf}, for a sufficiently large correlation coefficient, $\sqrt{\epsilon}\ll \rho<1$, i.e., the movements between frames are smooth, the reconstruction based on $0$-PLF-FMD for the second frame is given by $\hat{X}_2\approx (1-O(\epsilon))\hat{X}_1+O(\epsilon)X_2$, implying that the first frame is \emph{copied} in the future reconstruction.  In the third frame we have that $\hat{X}_3=X_3$  as $R_3 \rightarrow \infty$. 
On the other hand PLF-FMD also exhibits a tendency to copy the first frame when $R_3$ is small, as shown
in Appendix~\ref{FMd-high-R1-low-R23}. In our experiments we observe that the output of $0$-PLF-FMD  looks more \emph{static} when compared to the other PLFs. 

\underline{\textit{Small Correlation Coefficient}}: The case when $0<\rho\ll \sqrt{\epsilon}$, operationally captures the scenario when there are some sharp movements in frames. In this case we have $\hat{X}_2=O(\sqrt{\epsilon})X_2+Z'_{2,\text{FMD}}$ where $Z'_{2,\text{FMD}}\sim\mathcal{N}(0,(1-O(\epsilon))\sigma^2)$ is independent of $X_2$ and  $\hat{X}_3=X_3$. Note that the reconstruction  $\hat{X}_2$, largely ignores any correlation with $X_1$, which is undesirable in practice. We will demonstrate that this property of PLF-FMD leads to temporal inconsistency in the reconstructed frames in our experiments.

\textit{Achievable Reconstructions of $0$-PLF-JD}:

According to the third row of Table~\ref{table-ach-recons-inf-eps-inf}, the reconstruction of $0$-PLF-JD in the second and third frames are given by $\hat{X}_2{=}(\rho-O(\sqrt{\epsilon}))\hat{X}_1+O(\sqrt{\epsilon})X_2+Z_{2,\text{JD}}$ 
and $\hat{X}_3=(\rho-O(\sqrt{\epsilon}) \hat{X}_2+\frac{1}{\sqrt{1+\rho^2}}(\rho N_1+N_2)+Z_{3,\text{JD}}$, which mimic the correlation structure of the source model. 
One weakness of this decoder is that the noise $Z_{2,\text{JD}}$ introduced in the second step continues to propagate in the third step through the term  $\rho Z_{2,\text{JD}}$ in the expression for $\hat{X}_3$. 
We will see in our experiements that this can lead to undesirable errors in the reconstruction, indicating error propagation effect.  

\textit{Achievable Reconstructions of $0$-PLF-SA}:

The $0$-PLF-SA condition in the second frame is expressed as $P_{\hat{X}_1X_2}=P_{\hat{X}_1\hat{X}_2}$. When combined with the high compression rate for the initial frame (i.e., $R_1\to \infty$), it reduces to $P_{X_1X_2}= P_{\hat{X}_1\hat{X}_2}$, which is equivalent to the constraint in the $0$-PLF-JD framework. Thus, the reconstruction of the second frame for $0$-PLF-SA is similar to that of $0$-PLF-JD, i.e., $\hat{X}_2{=}(\rho-O(\sqrt{\epsilon}))\hat{X}_1+O(\sqrt{\epsilon})X_2+Z_{2,\text{SA}}$ where $Z_{2,\text{SA}}=Z_{2,\text{JD}}$. For the third frame, the reconstruction is given by $\hat{X}_3{=}X_3$ due to the high rate.  Thus, the decoder based PLF-SA differs from PLF-JD in that the reconstruction is not strongly dependent on the noise in the second step. In our experiments, we also demonstrate that PLF-SA indeed has an improved reconstruction over PLF-JD.

\begin{figure*}[ht]
    \centering
    \includegraphics[width=0.98\linewidth]{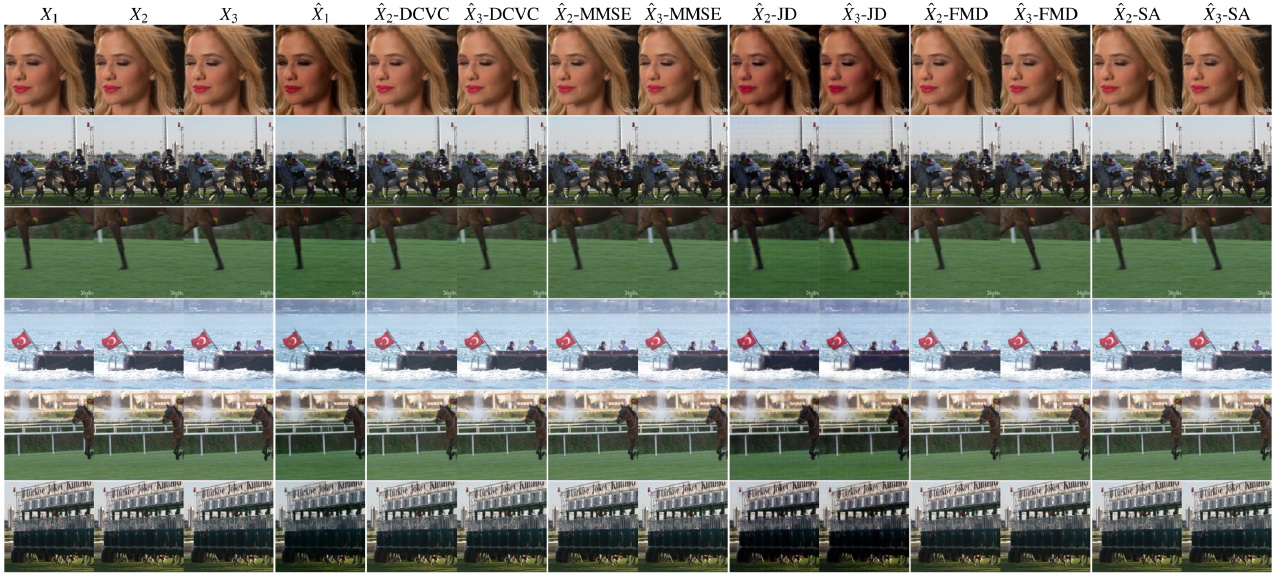}
    \caption{The reconstruction results on the UVG dataset when the first frame is compressed at a low rate $R_1=0.144$ bpp. 
    $\hat{X_1}$ is shared across all models. Similar to the Gauss-Markov case and MovingMNIST results, PLF-SA and PLF-FMD exhibit robustness to first-frame errors (color tone mismatches) while PLF-JD suffers from error permanence.
    }
    \label{fig:appendix_2}
    \vspace{-1.5em}
\end{figure*}

\section{Experimental Results}\label{experimental}

Our theoretical results for PLF-SA show that PLF-SA is a new perceptual metric that inherits advantages in both PLF-JD and PLF-FMD.  In this section we provide experimental results to further demonstrate this effect.

\subsection{Implementation Details}
\label{sec:exp_details}

Expanding upon the experimental framework established in \cite{Jun-Ashish2023}, we merge the scale-space-flow neural video coding architecture introduced by \cite{agustsson2020scale} with Wasserstein GANs for perceptual quality enhancement, as proposed in \cite{gulrajani2017improved}. 
We employ two datasets: the 1-digit MovingMNIST dataset \cite{srivastava2015unsupervised} and UVG dataset \cite{mercat2020UVG}, offering varying levels of video resolution and scene complexity. 
The MovingMNIST dataset consists of low-complexity synthetic sequences with dimensions of $64\times64$, while the UVG dataset comprises high-definition real-life video patches sized at $256\times256$. 
The preference for certain deep learning structures and datasets aims at confirming the suggested theory rather than developing the most advanced neural network architectures.


To evaluate the compression performance of the proposed PLF-SA, we compare it with prior perception loss models, namely PLF-FMD  and PLF-JD~\cite{Jun-Ashish2023}. We also compare with another baseline,
DCVC-HEM~\cite{li2022hybrid}, which makes  use of MS-SSIM loss during training and its manually designed module for capturing strong temporal correlations through multi-scale features from previously decoded frames.
Further experimental details can be found in Appendix~\ref{experiment-app}.

\subsection{Main Results}

We first present the results validating the low-rate regime analysis described in Section~\ref{low_rate}. 
Following that, we provide the complementary results for the high-rate regime analysis discussed in Section~\ref{high_rate}. 

\begin{figure*}[h]
    \centering
    \begin{subfigure}{0.5\linewidth}
        \centering
        \includegraphics[width=0.98\linewidth]{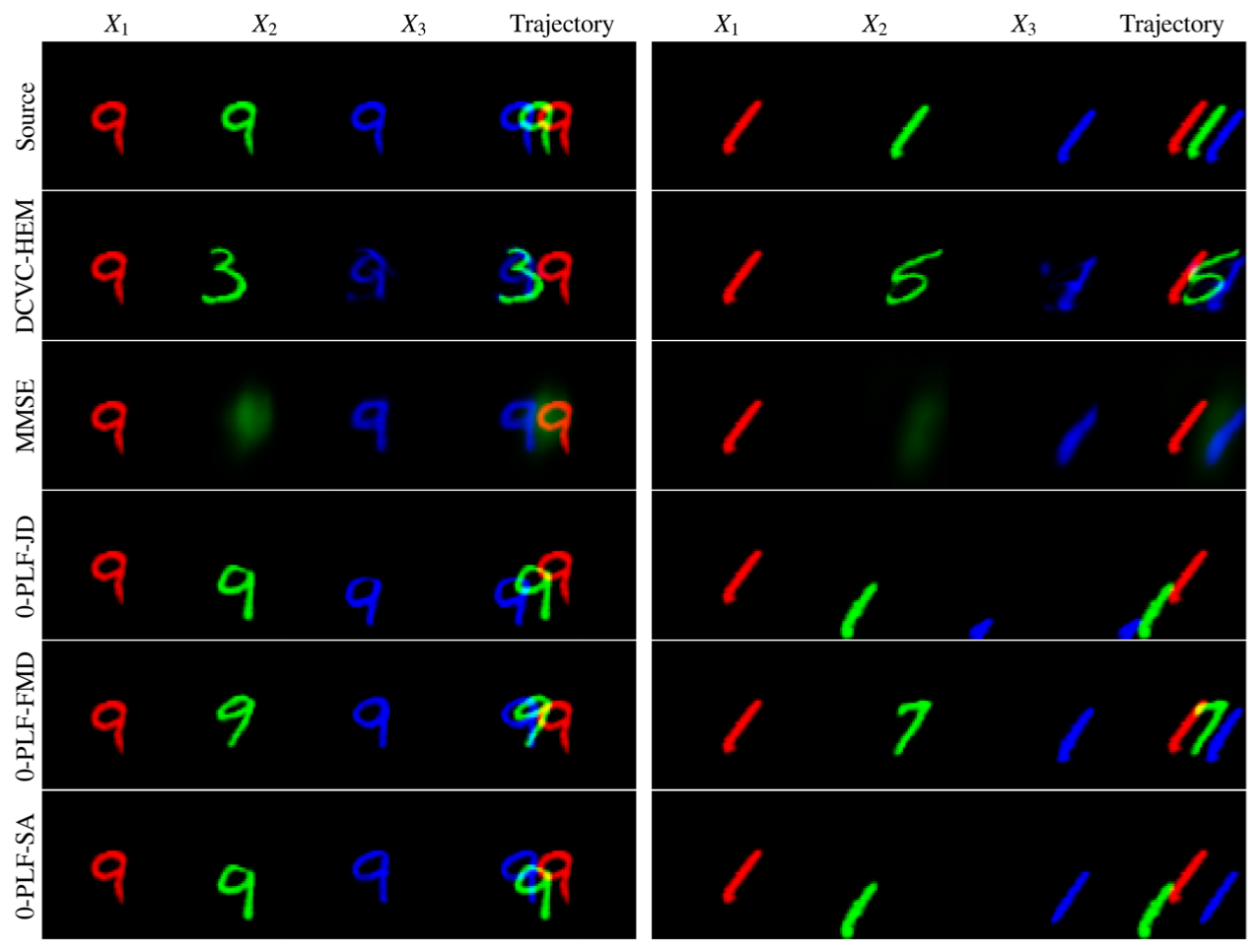}
        \subcaption{Sharp movement scenario.}
        \label{fig:appendix_1_a}
    \end{subfigure}
    \vspace{-.5em}
    \hspace{-10pt}
    \begin{subfigure}{0.5\linewidth}
    \centering
        \includegraphics[width=0.98\linewidth]{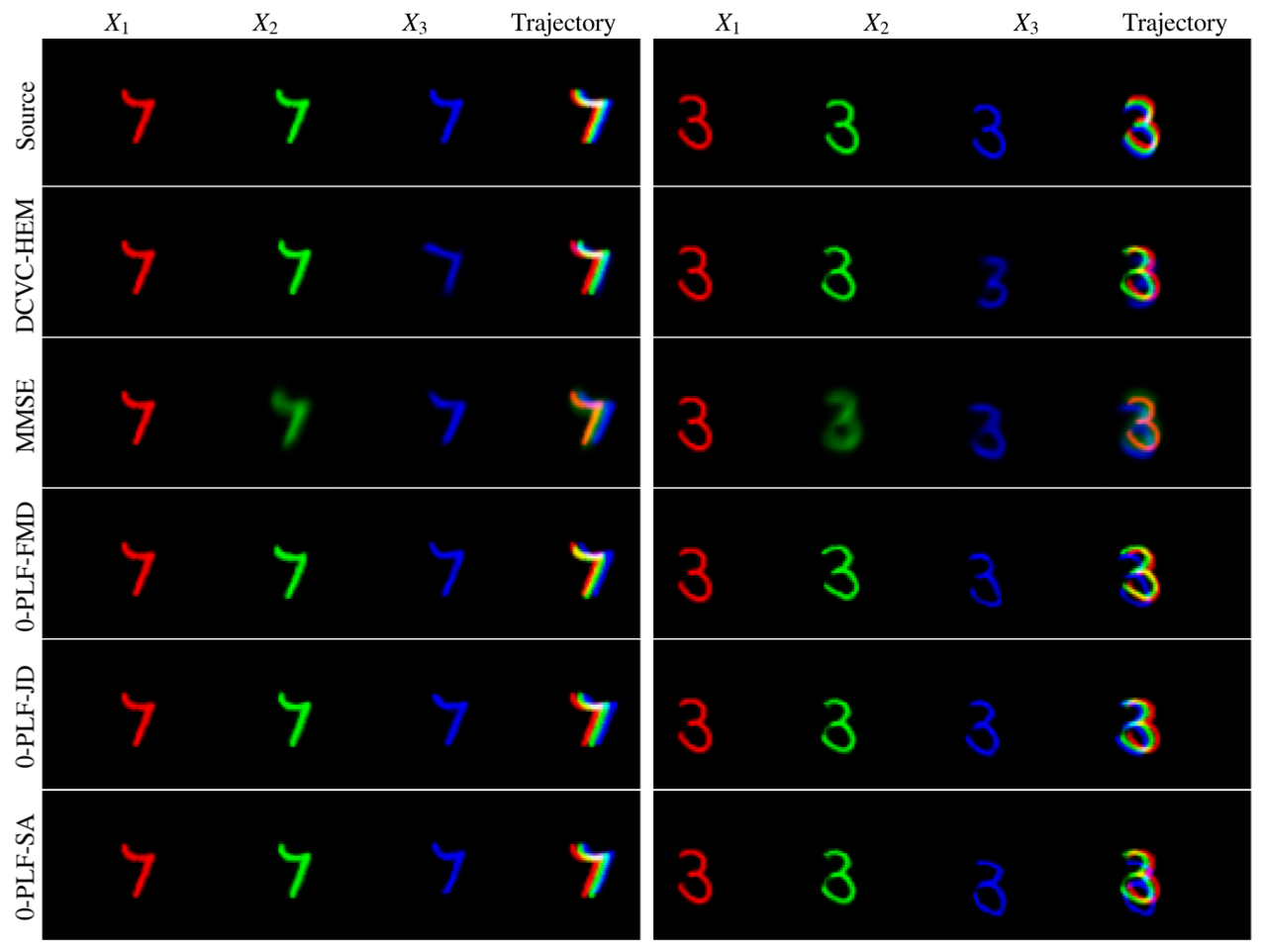}
        \subcaption{Slow movement scenario.}
        \label{fig:appendix_1_b}
    \end{subfigure}
    \vspace{-.5em}
    \captionsetup{width=0.9\linewidth}
    \caption{Reconstruction results on the MovingMNIST dataset for $\infty$-$R_2$-$R_3$ with $R_2 = 2$ bits and $R_3 = 16$ bits. Colored digits highlight trajectory across frames.
    (a) With small correlation coefficient $0 < \rho \ll \sqrt{\epsilon}$, PLF-FMD preserves direction but loses temporal consistency in digits' contour. PLF-JD and PLF-SA fail to identify the direction in the second frame, but PLF-SA rectifies the error in the third frame.
    (b) With large correlation coefficient $\sqrt{\epsilon}\ll \rho < 1$, PLF-FMD tends to replicate the first frame without capturing motion effectively, while PLF-JD and PLF-SA show greater generative diversity.}
    \label{fig:appendix_1}
    \vspace{-1em}
\end{figure*}

\subsubsection{Low-rate Case $R_1 = 12$ bits}

We first validate the achievable reconstructions and distortions for $R_1 = \epsilon$ discussed in Section~\ref{low_rate}. 
Fig.~\ref{fig:mnist_1aa}a and Fig.~\ref{fig:appendix_3} show samples of 3-frame MovingMNIST sequences where the first frame is encoded at a low bitrate $R_1=12\ \text{bits}$. 
As shown in third row of Table~\ref{table-ach-recons-low-rate}, given an incorrect reconstruction in $\hat{X}_1$, the decoder with $0$-PLF-JD exhibit the error permanence phenomenon for future frame reconstructions, as it tends to replicate the reconstructed first frame as discussed in Table~\ref{table-ach-recons-low-rate}.
Furthermore, as in the first and second rows of Table~\ref{table-ach-recons-low-rate}, the decoders with $0$-PLF-FMD and proposed $0$-PLF-SA utilize new information from $X_2$ to recover from wrongly predicted $\hat{X}_1$ with $\hat{X}_2 = \omega_1 \hat{X_1} + \omega_2 X_2 + Z_2$. 
This highlights their capability to rectify previous mistakes. 
Results for DCVC-HEM and MMSE-based are also presented. 
Due to the low bitrate setting, the MMSE reconstructions tend to be blurry. 
DCVC-HEM also suffers from error propagation with digit ``$7$'' wrongly decoded as ``$3$'' in Fig.~\ref{fig:mnist_1aa}a.

\begin{table}[t]
    \centering
    \caption{
    PSNR and LPIPS comparisons under the low-bitrate setting $R_1 = 0.144$ bpp on the UVG dataset. Among these models, PLF-FMD achieves the lowest distortion across two frames $\hat{X}_2$ and $\hat{X}_3$, with PLF-SA closely following. PLF-JD performs the worse due to error permanence.}
    \vspace{+.5em}
    \small
    \begin{tabular}{|p{0.25\linewidth}|p{0.12\linewidth}<{\centering}p{0.12\linewidth}<{\centering}|p{0.12\linewidth}<{\centering}p{0.12\linewidth}<{\centering}|}
        \hline
        & \multicolumn{2}{c|}{PSNR$\uparrow$} & \multicolumn{2}{c|}{LPIPS$\downarrow$} \\
        \cline{2-5}
        \hfill & $\hat{X}_2$ & $\hat{X}_3$ & $\hat{X}_2$ & $\hat{X}_3$ \\
        \hline
        DCVC-HEM & 28.11 & 28.74 & 0.039 & 0.028 \\
        \hline
        PLF-JD & 22.38 & 21.99 & 0.049 & 0.053 \\
        \hline
        PLF-SA & 30.59 & 30.67 & 0.0039 & 0.0043 \\
        \hline
        PLF-FMD & 31.02 & 30.72 & 0.0036 & 0.0041 \\
        \hline
    \end{tabular}
    \label{tab:r1_epsilon}
    \vspace{-1.5em}
\end{table}

Analogous results for UVG dataset are shown in Fig.~\ref{fig:mnist_1aa}b and Fig.~\ref{fig:appendix_2}. 
When the first frame is compressed at a low rate $R_1 = 0.144$ bpp, the reconstructed frame $\hat{X}_1$ exhibits a noticeable degradation in overall color tone. 
For the decoder with $0$-PLF-JD, this error propagates to future reconstructions, $\hat{X}_2$ and $\hat{X}_3$. 
In contrast, $0$-PLF-FMD and $0$-PLF-SA correct the color tone in the reconstructions of $\hat{X}_2$ and $\hat{X}_3$. 
Additionally, DCVC-HEM preserves the correct color tone but struggles to reconstruct fine details, such as eye pupils in Fig.\ref{fig:mnist_1aa}b, where the PLF models demonstrate better performance.
Furthermore, PSNR and Perceptual metric comparisons on UVG dataset are presented in Table~\ref{tab:r1_epsilon}. 
All models are evaluated across 2000 frames for $\hat{X}_2$ and 2000 frames for $\hat{X}_3$ under the same low-bitrate setting ($R_1 = 0.144$ bpp). 
As discussed in Section~\ref{low_rate}, PLF-FMD achieves the lowest distortion, with PLF-SA closely following. 
In contrast, PLF-JD exhibits the worst performance due to the error permanence phenomenon.
For perceptual metric, PLF-SA and PLF-FMD exhibit similar performance.

\subsubsection{High-rate Case $R_1 = \infty$ bits}

\begin{table}[t]
    \centering
    \caption{
    PSNR and LPIPS comparisons under the high-bitrate setting $R_1 = \infty$ on MovingMNIST with small $\rho$. PLF-JD and PLF-SA exhibit higher distortion than PLF-FMD due to trajectory errors. PLF-SA achieves the best perceptual quality on $\hat{X}_2$, $\hat{X}_3$, while PLF-FMD struggles with temporal correlation.}
    \vspace{+.5em}
    \small
    \begin{tabular}{|p{0.25\linewidth}|p{0.12\linewidth}<{\centering}p{0.12\linewidth}<{\centering}|p{0.12\linewidth}<{\centering}p{0.12\linewidth}<{\centering}|}
        \hline
        & \multicolumn{2}{c|}{PSNR$\uparrow$} & \multicolumn{2}{c|}{LPIPS$\downarrow$} \\
        \cline{2-5}
        & $\hat{X}_2$ & $\hat{X}_3$ & $\hat{X}_2$ & $\hat{X}_3$ \\
        \hline
        DCVC-HEM & 14.50 & 20.42 & 0.115 & 0.073 \\
        \hline
        PLF-JD & 13.54 & 14.82 & 0.026 & 0.077 \\
         \hline
        PLF-SA & 13.54 & 20.47 & 0.026 & 0.021 \\
        \hline
        PLF-FMD & 14.74 & 20.73 & 0.114 & 0.024 \\
        \hline
    \end{tabular}
    \label{tab:r1_infty}
    \vspace{-1.5em}
\end{table}

We now validate the achievable reconstructions and distortions for $R_1 = \infty$ ($\hat{X}_1=X_1$) discussed in Section \ref{high_rate}.
and Table~\ref{table-ach-recons}. 
Fig.~\ref{fig:mnist_2_fast} and Fig.~\ref{fig:appendix_1} show experimental results on MovingMNIST where $R_2=2$ bits and $R_3=16$ bits representing low and medium rates. 
The source digit maintains its motion direction across three frames.
We evaluate each model's performance on reconstruction of $\hat{X}_2$ and $\hat{X}_3$ and analyze the digit moving trajectory across three frames.
We consider both small and large correlation coefficient $\rho$ correspond to the scenarios where the video sampling rate is high and low respectively. 

\underline{\textit{Small Correlation Coefficient}}: Fig.\ref{fig:mnist_2_fast} and Fig.\ref{fig:appendix_1_a} show results for small correlation coefficients $0 < \rho \ll \sqrt{\epsilon}$. 
Both $0$-PLF-SA and $0$-PLF-JD fail to identify the correct direction in the second frame, producing identical reconstructions ($\hat{X}_{2,\text{SA}} = \hat{X}_{2,\text{JD}}$) as shown in third and fourth rows of Table~\ref{table-ach-recons-inf-eps-inf}. 
 By the third frame, $0$-PLF-JD exhibits error permanence, propagating second frame noise ($\rho Z_{2,\text{JD}}$) to third frame (see the third row of Table~\ref{table-ach-recons-inf-eps-inf}), while $0$-PLF-SA reduces noise dependence and accurately reconstructs $\hat{X}_3 = X_3$ when $R_3 \to \infty$ (see the fourth row of Table~\ref{table-ach-recons-inf-eps-inf}). For $0$-PLF-FMD, temporal correlation is less effectively preserved. It introduces synthetic noise in the second frame ($\hat{X}_2 = O(\sqrt{\epsilon})X_2 + Z'_{2,\text{FMD}}$) as in the second row of Table~\ref{table-ach-recons-inf-eps-inf}, decoding correct direction but often changing digit contours. 
In contrast, $0$-PLF-SA balances content preservation and error correction, maintaining the digit's identity and direction under low bitrate conditions. 
The MMSE model produces blurry $\hat{X}_2$ at low rates but retains correct direction. 
By the third frame, $\hat{X}_3$ improves with a medium bitrate but lacks fine details. 
DCVC-HEM, using $\hat{X}_{2,\text{SA}}$ from $0$-PLF-SA as input, corrects the direction in $\hat{X}_3$ but struggles with digit contours. 
Numerical results in Table~\ref{tab:r1_infty} show that in the second frame, $0$-PLF-JD and $0$-PLF-SA exhibit higher distortion than $0$-PLF-FMD due to trajectory errors. By the third frame, $0$-PLF-JD propagates these errors to $\hat{X}_3$, whereas $0$-PLF-SA corrects them, approaching $0$-PLF-FMD’s distortion. 
Additionally, $0$-PLF-FMD struggles with temporal correlation in $\hat{X}_2$, resulting in the worst LPIPS score, while $0$-PLF-SA achieves the best perceptual quality for both $\hat{X}_2$ and $\hat{X}_3$. 
These results highlight PLF-SA’s robustness to small $\rho$ under low-bitrate settings ($R_2 = 2$ bits, $R_3 = 16$ bits).

\underline{\textit{Large Correlation Coefficient}}: Fig.~\ref{fig:appendix_1_b} show results for large correlation coefficient $\sqrt{\epsilon}\ll \rho < 1$. 
Corresponding to the first row of Table~\ref{table-ach-recons-inf-eps-inf}, $0$-PLF-FMD tends to \textit{copy} the first-frame $\hat{X}_1$ when reconstructing the second frame with $\hat{X}_2\approx (1-O(\epsilon))\hat{X}_1+O(\epsilon)X_2$, resulting in a lack of generative diversity. 
In contrast, $0$-PLF-JD and $0$-PLF-SA do not exhibit such ``static'' reconstruction behavior for the second frame. 
MMSE and DCVC-HEM perform better compared with small $\rho$ case. 
However, issues such as blurriness and discrepancies in image details still persist. 
Overall, PLF-SA demonstrates a superior ability to balance reconstruction distortion and perceptual quality across various bitrate settings.

\section{Conclusions}

We observe that previously proposed perception loss functions (PLF) in video compression can have disadvantages in different operating regimes. In particular, the PLF-JD metric that preserves the joint distribution of all the frames suffers from the effect of error permanence, where mistakes made in previously reconstructed frames carry over in subsequent frames. On the other hand, the PLF-FMD metric that only preserves  marginal distribution of frames does not effectively exploit the temporal correlation during reconstruction. Motivated by these observations, we propose a new metric PLF-SA that mitigates the disadvantages of each. When the previously reconstructed frames are of lower quality, our proposed metric avoids the error permanence phenomenon in PLF-JD. When the previously reconstructed frames are of higher quality, the decoder based on PLF-SA effectively exploits temporal correlation between frames.  We validate the merits of our proposed metric through experimental results involving moving-MNIST and UVG datasets in a variety of operating regimes. We also provide information theoretic analysis of the first order Gauss-Markov source model to further explain the  qualitative  behavior of each PLF metric.

\section*{Impact Statement}
This paper presents work whose goal is to advance the field of Machine Learning. There are many potential societal consequences of our work, none of which we feel must be specifically highlighted here.

\bibliography{NeurIPS_references}

\bibliographystyle{icml2025}

\newpage


\appendix
\clearpage
\section{Operational RDP Region}\label{operational-RDP-app}
It is not feasible to compute the region $\mathcal{RDP}^o$ directly since it involves searching over all possible encoding-decoding functions. But, for first-order Markov sources where the Markov chain $X_1\to X_2\to X_3$ holds, the following region can be used as an approximation. So, with this motivation, we introduce the information RDP region as follows. 

\begin{definition}[Information  RDP Region]\label{iRDP-region} For first-order Markov sources, let the information RDP region, denoted by $\mathcal{RDP}$, be the set of all tuples $(\mathsf{R},\mathsf{D},\mathsf{P})$ which satisfy the following
$(\mathsf{R},\mathsf{D},\mathsf{P})$ satisfying
\begin{IEEEeqnarray}{rCl}
R_1 &\geq & I(X_1;X_{r,1}),\label{MaIsh-rate1}\\
R_2 &\geq & I(X_2;X_{r,2}|X_{r,1}),\label{MaIsh-rate2}\\
R_3 &\geq & I(X_3;X_{r,3}|X_{r,1},X_{r,2}),\label{MaIsh-rate3}\\
D_j &\geq & \mathbbm{E}[\|X_j-\hat{X}_j\|^2], \qquad\qquad\;\;\; j=1,2,3,\label{MaIsh-distortion}\\
P_j &\geq & \phi_j(P_{\hat{X}_{1}\ldots\hat{X}_{j-1} X_{j}}, P_{\hat{X}_{1}\ldots\hat{X}_{j-1}\hat{X}_{j}}), \;\;\;\;\;\; j=1,2,3,\nonumber\\\label{MaIsh-perception}
\end{IEEEeqnarray}
for auxiliary random variables $(X_{r,1},X_{r,2},X_{r,3})$ and $(\hat{X}_1,\hat{X}_2,\hat{X}_3)$ such that
\begin{IEEEeqnarray}{rCl}
\hat{X}_1&=& \eta_1(X_{r,1}), \;\;\hat{X}_2=\eta_2(X_{r,1},X_{r,2}),\;\;\hat{X}_3=X_{r,3},\nonumber\\\label{Markov1-new}\\
X_{r,1}&\to& X_1\to (X_2,X_3),\label{Markov2-new}\\X_{r,2}&\to& (X_2,X_{r,1})\to (X_1,X_3),\label{Markov3-new}\\X_{r,3}&\to& (X_3,X_{r,1},X_{r,2})\to (X_1,X_2),\label{Markov4-new}
\end{IEEEeqnarray}
for some deterministic functions $\eta_1(.)$ and $\eta_2(.,.)$. Moreover, for a given $(\mathsf{D},\mathsf{P})$, the information rate region, denoted by $\mathcal{R}(\mathsf{D},\mathsf{P})$, is the closure of the set of all tuples $\mathsf{R}$ that $(\mathsf{R},\mathsf{D},\mathsf{P})\in \mathcal{RDP}$. 
\end{definition}

\begin{proposition} For first-order Markov sources, a given $(\mathsf{D},\mathsf{P})$ and $\mathsf{R}\in \mathcal{R}(\mathsf{D},\mathsf{P})$, we have
\begin{IEEEeqnarray}{rCl}
\mathsf{R}+\log(\mathsf{R}+1)+5\in \mathcal{R}^{o}(\mathsf{D},\mathsf{P}).\label{inner-bound-1-shot}
\end{IEEEeqnarray}
Moreover, the following holds:
\begin{IEEEeqnarray}{rCl}
\mathcal{R}^{o}(\mathsf{D},\mathsf{P})\subseteq \mathcal{R}(\mathsf{D},\mathsf{P}).\label{outer-bound-1-shot}
\end{IEEEeqnarray}
\end{proposition}

To prove the above statement, we first discuss the achievable scheme that results in~\eqref{inner-bound-1-shot}. Then, we will provide the proof of outer bound in~\eqref{outer-bound-1-shot}.

Before stating the achievable scheme, we remind the strong functional representation lemma \cite{LiElGamal}. It states that for jointly distributed random variables $X$ and $Y$, there exists a random variable $U$ independent of $X$, and function $\phi$ such that $Y = \phi(X, U)$. Here, $U$ is not necessarily unique. The
strong functional representation lemma states further that there exists a $U$ which has information of $Y$ in the sense that
\begin{IEEEeqnarray}{rCl}
H(Y|U)\leq I(X;Y)+\log(I(X;Y)+1)+4.
\end{IEEEeqnarray}
Notice that the strong functional representation lemma can be applied conditionally. Given $P_{XY|W}$, we can represent $Y$ as a function of $(X,W,U)$ such that $U$ is independent of $(X,W)$ and 
\begin{IEEEeqnarray}{rCl}
H(Y|W,U)\leq I(X;Y|W)+\log(I(X;Y|W)+1)+4.\nonumber\\
\end{IEEEeqnarray}

\underline{\textit{Proof of \eqref{inner-bound-1-shot} (Inner bound)}}:

For a given $(\mathsf{D},\mathsf{P})$ and $\mathsf{R}\in\mathcal{R}(\mathsf{D},\mathsf{P})$, let $\mathsf{X}_r=(X_{r,1},X_{r,2},X_{r,3})$ be jointly distributed with $\mathsf{X}=(X_1,X_2,X_3)$ where the Markov chains \eqref{Markov2-new}--\eqref{Markov4-new} hold and the rate constraints in \eqref{MaIsh-rate1}--\eqref{MaIsh-rate3} are satisfied such that there exist $(\hat{X}_1,\hat{X}_2,\hat{X}_3)$ for which distortion-perception constraints \eqref{MaIsh-distortion}--\eqref{MaIsh-perception} hold. Denote the joint distribution of $(\mathsf{X},\mathsf{X}_r,\hat{\mathsf{X}})$ by $P_{\mathsf{X}\mathsf{X}_r\hat{\mathsf{X}}}$ and notice that according to the Markov chains in \eqref{Markov2-new}--\eqref{Markov4-new}, it factorizes as the following
\begin{IEEEeqnarray}{rCl}
P_{\mathsf{X}\mathsf{X}_r\hat{\mathsf{X}}}&=&P_{X_1X_2X_3}\cdot P_{X_{r,1}|X_1}\cdot P_{X_{r,2}|X_{r,1}X_2}\nonumber\\&&\hspace{0.5cm}\cdot P_{X_{r,3}|X_{r,2}X_{r,1}X_3} \cdot\mathbbm{1}\{\hat{X}_1=g_1(X_{r,1})\}\nonumber\\&&\hspace{0.5cm}\cdot \mathbbm{1}\{\hat{X}_2=g_2(X_{r,1},X_{r,3})\}\cdot \mathbbm{1}\{\hat{X}_3=X_{r,3}\}.\nonumber\\\label{super-P}
\end{IEEEeqnarray}

\begin{figure*}[t]
\centering
  \includegraphics[scale=0.3]{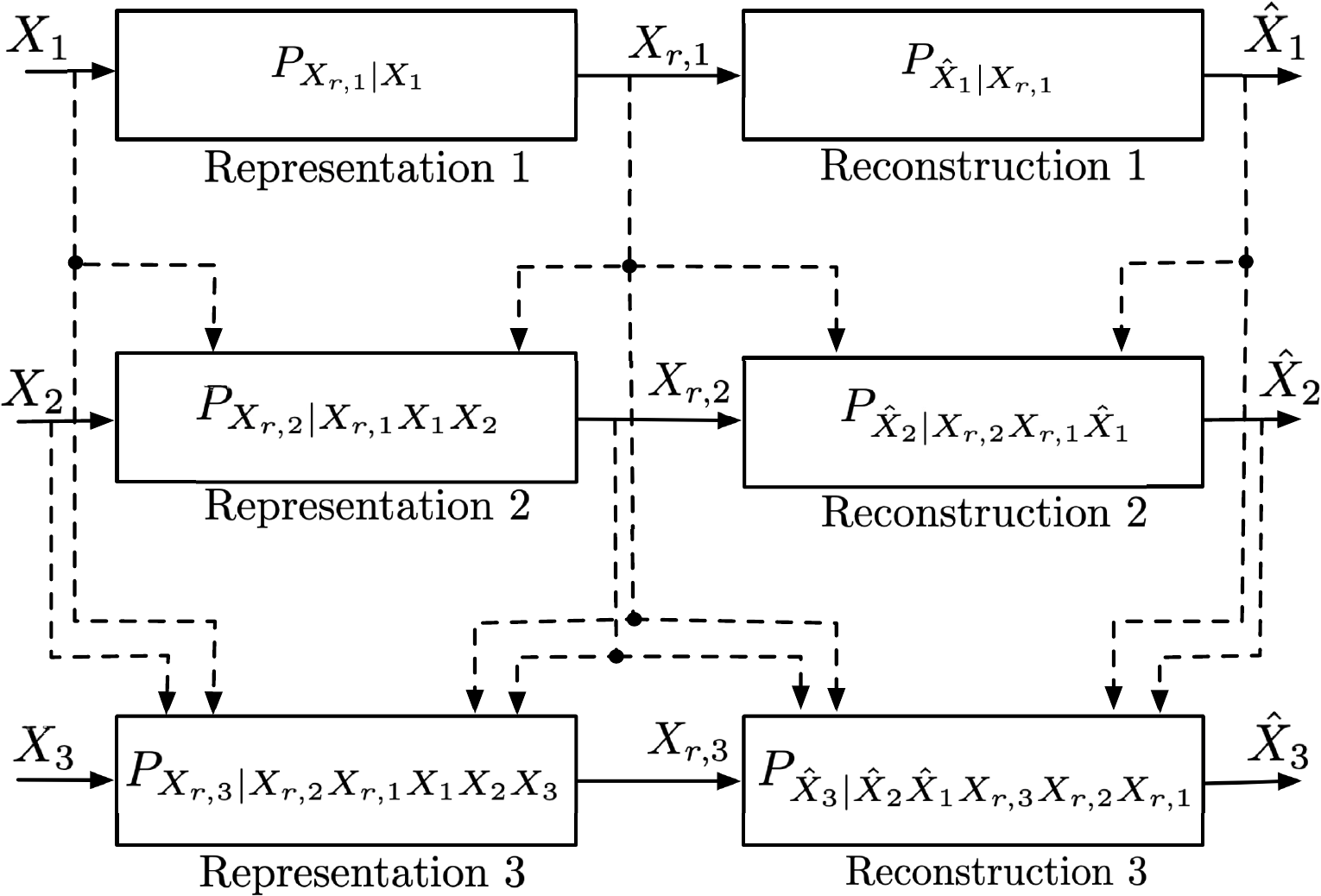}
\caption{Encoded representations and reconstructions of the iRDP region $\mathcal{RDP}$.}
\label{enc-dec2}
\end{figure*}

For an illustration of encoded representations $\mathsf{X}_r$ and reconstructions $\hat{\mathsf{X}}$ in $\mathcal{R}(\mathsf{D},\mathsf{P})$ which are induced by distribution $P_{\mathsf{X}\mathsf{X}_r\hat{\mathsf{X}}}$, see Fig.~\ref{enc-dec2}. 

Now, we show that $\mathsf{R}+\log(\mathsf{R}+1)+5\in\mathcal{R}(\mathsf{D},\mathsf{P})$. The achievable scheme is as follows. Fix the joint distribution $P_{\mathsf{X}_r}$ according to~\eqref{super-P} which constructs the codebook, given by
\begin{IEEEeqnarray}{rCl}
P_{\mathsf{X}_r}=P_{X_{r,1}}P_{X_{r,2}|X_{r,1}}P_{X_{r,3}|X_{r,2}X_{r,1}}.\label{codebook-construction-P}
\end{IEEEeqnarray}

\begin{figure*}[t]
\centering
  \includegraphics[scale=0.32]{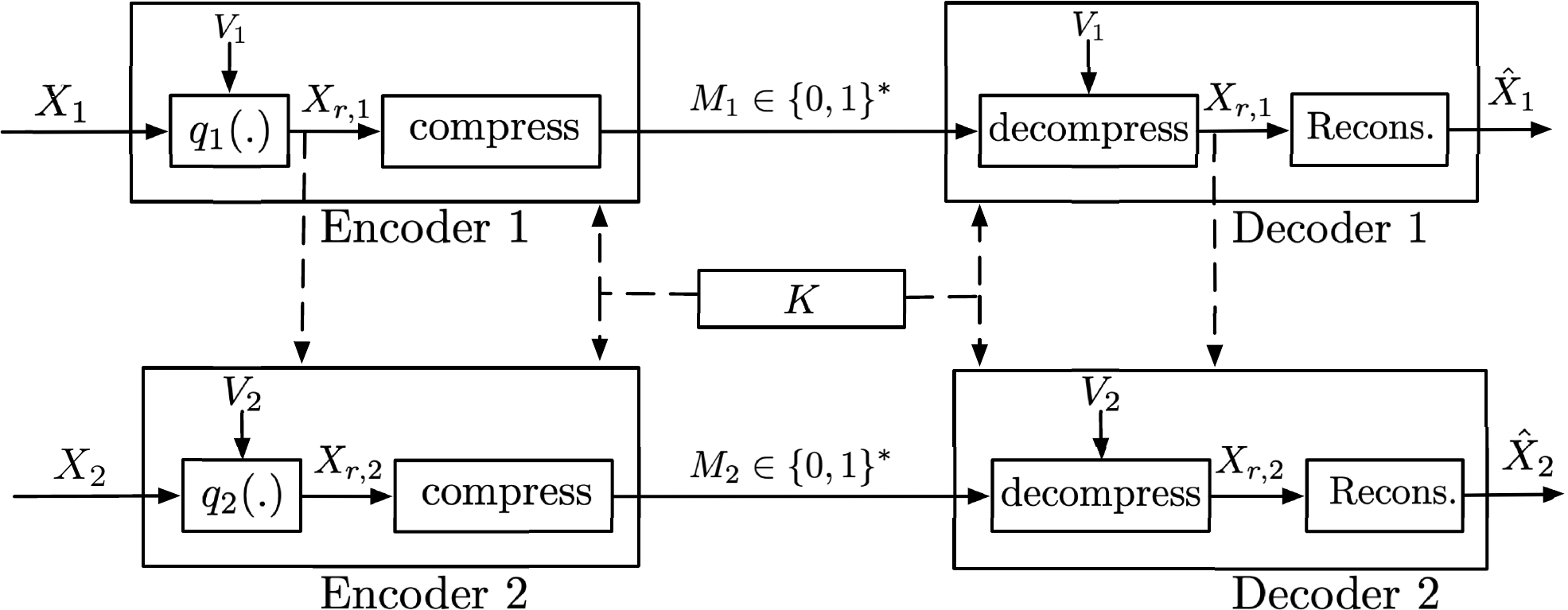}
\caption{Strong functional representation lemma for $T=2$ frames.}
\label{SFRL}
\end{figure*}

From the strong functional representation lemma \cite{LiElGamal}, we know that
\begin{itemize}
\item there exist a random variable $V_1$ independent of $X_1$ and a deterministic function $q_1$ such that $X_{r,1}=q_1(X_1,V_1)$ and
\begin{IEEEeqnarray}{rCl}
H(X_{r,1}|V_1)&\leq& I(X_1;X_{r,1})+\log(I(X_1;X_{r,1})+1)\nonumber\\&&\hspace{3cm}+4,
\end{IEEEeqnarray}
which means that the first encoder observes the source $X_1$ and applies the function $q_1$ to get $X_{r,1}$ whose  distribution needs to be preserved according to~\eqref{codebook-construction-P} (see Fig.~\ref{SFRL});
\item according to the conditional strong functional representation lemma, there exist a random variable $V_2$ independent of $(X_2,X_{r,1})$ and a deterministic function $q_2$ such that $X_{r,2}=q_2(X_{r,1},X_2,V_2)$ and
\begin{IEEEeqnarray}{rCl}
H(X_{r,2}|X_{r,1},V_2)&\leq& I(X_2;X_{r,2}|X_{r,1})\nonumber\\&&\hspace{-1cm}+\log(I(X_2;X_{r,2}|X_{r,1})+1)+4.\nonumber\\
\end{IEEEeqnarray}
At the second step, the representation $X_{r,1}$ is available at the second encoder. So, upon observing the source $X_2$, it applies the function $q_2$ to get $X_{r,2}$ whose conditional distribution given $X_{r,1}$ needs to be preserved according to~\eqref{codebook-construction-P} (see Fig.~\ref{SFRL});
\item according to the conditional strong functional representation lemma, there exist a random variable $V_3$ independent of $(X_3,X_{r,1},X_{r,2})$ and a deterministic function $q_3$ such that $X_{r,3}=q_3(X_{r,1},X_{r,2},X_3,V_3)$ and \begin{IEEEeqnarray}{rCl}H(X_{r,3}|X_{r,1},X_{r,2},V_3)&\leq &I(X_3;X_{r,3}|X_{r,1},X_{r,2})\nonumber\\&&\hspace{-2cm}+\log(I(X_3;X_{r,3}|X_{r,1},X_{r,2})+1)+4.\nonumber\\\end{IEEEeqnarray}
\end{itemize}

Now, the encoding and decoding are as follows
\begin{itemize}
\item With $V_1$ available at all encoders and decoders, we can have a class of prefix-free binary codes indexed by $V_1$ with the expected codeword length not larger than $I(X_1;X_{r,1})+\log(I(X_1;X_{r,1})+1)+5$ to represent $X_{r,1}$, losslessly (see Fig.~\ref{SFRL}).
\item With $V_2$ available at the encoders and decoders, we can design a set of prefix-free binary codes indexed by $(V_2,X_{r,1})$ with expected codeword length not larger than $I(X_2;X_{r,2}|X_{r,1})+\log(I(X_2;X_{r,2}|X_{r,1})+1)+5$ to represent $X_{r,2}$, losslessly (see Fig.~\ref{SFRL}).
\item Similarly, one can represent $X_{r,3}$ losslessly with $V_3$ available at the third encoder and decoder.
\item The decoders can use functions $\hat{X}_1=\eta_1(X_{r,1})$, $\hat{X}_2=\eta_2(X_{r,1},X_{r,2})$ and $\hat{X}_3=X_{r,3}$ to get the reconstruction $\hat{\mathsf{X}}$.
\end{itemize}
This shows that $\mathsf{R}+\log(\mathsf{R}+1)+5\in \mathcal{R}^{o}(\mathsf{D},\mathsf{P})$.

\underline{\textit{Proof of \eqref{outer-bound-1-shot} (Outer Bound)}}:

For any $(\mathsf{D},\mathsf{P})$, $\mathsf{R}\in\mathcal{R}^{o}(\mathsf{D},\mathsf{P})$, shared randomness $K$, encoding functions $f_{j}\colon \mathcal{X}_1\times \ldots \times \mathcal{X}_j\times \mathcal{K}\to \mathcal{M}_{j}$ and decoding functions $g_{j} \colon \mathcal{M}_{1}\times \mathcal{M}_{2}\times \ldots \times \mathcal{M}_{j}\times \mathcal{K}\to \hat{\mathcal{X}}_j$ such that 
\begin{IEEEeqnarray}{rCl}
R_j\geq \mathbbm{E}[\ell(M_j)],\qquad j=1,2,3,
\end{IEEEeqnarray}
and
\begin{IEEEeqnarray}{rCl}
D_j &\geq & \mathbbm{E}[\|X_j-\hat{X}_j\|^2], \qquad\qquad\;\;\; j=1,2,3,\label{Dj-1-shot}\\
P_j &\geq & \phi_j(P_{\hat{X}_{1}\ldots\hat{X}_{j-1} X_{j}},P_{\hat{X}_1\ldots\hat{X}_j}), \;\;\;\;\;\; j=1,2,3,\label{Pj-1-shot}
\end{IEEEeqnarray}
we lower bound the expected length of the messages. Define
\begin{IEEEeqnarray}{rCl}
X_{r,1}&:=&(M_1,K),\label{Xr1-def-1-shot}\\
X_{r,2}&:=& (M_1,M_2,K),\label{Xr2-def-1-shot}
\end{IEEEeqnarray}
and recall that according to the decoding functions, we have
\begin{IEEEeqnarray}{rCl}
\hat{X}_j=g_j(M_1,\ldots,M_j,K),\qquad j=1,2,3.
\end{IEEEeqnarray}
We can write
\begin{IEEEeqnarray}{rCl}
R_1&\geq& \mathbbm{E}[\ell(M_1)]\\&\geq & H(M_1|K)\\
&=&I(X_1;M_1|K)\\
&=&I(X_{1};M_1,K)\\
&=& I(X_{1};X_{r,1}).\label{1-shot-proof-R1}
\end{IEEEeqnarray}
Now, consider the following set of inequalities
\begin{IEEEeqnarray}{rCl}
R_2&\geq& \mathbbm{E}[\ell(M_2)]\\&\geq &H(M_2|M_1,K)\\
&=&I(X_1,X_2;M_2|M_1,K)\\
&=& I(X_1,X_{2};X_{2,r}|X_{r,1}).\label{1-shot-proof-R2}
\end{IEEEeqnarray}
Similarly, we have
\begin{IEEEeqnarray}{rCl}
R_3&\geq& \mathbbm{E}[\ell(M_3)]\\&\geq& H(M_3|M_1,M_2,K)\\
&=& I(X_1,X_2,X_3;M_3|M_1,M_2,K)\\
&\geq &I(X_1,X_2,X_{3};\hat{X}_{3}|X_{r,1},X_{r,2}).\label{1-shot-proof-R3}
\end{IEEEeqnarray}
Notice that the definitions in \eqref{Xr1-def-1-shot}--\eqref{Xr2-def-1-shot} imply the following Markov chains
\begin{IEEEeqnarray}{rCl}
X_{r,1}&\to& X_1\to (X_2,X_3),\label{1-shot-proof-Markov1}\\
X_{r,2} &\to & (X_1,X_2,X_{r,1})\to X_3.\label{1-shot-proof-Markov2}
\end{IEEEeqnarray}
On the other hand, the decoding functions of the first and second steps imply that
\begin{IEEEeqnarray}{rCl}
\hat{X}_1&=&g_1(M_1,K),\label{dec-1-proof-app}\\
\hat{X}_2&=&g_2(M_1,M_2,K),\label{dec-2-proof-app}
\end{IEEEeqnarray}
where together with definitions in~\eqref{Xr1-def-1-shot} and \eqref{Xr2-def-1-shot}, we can write
\begin{IEEEeqnarray}{rCl}
\hat{X}_1&=&g_1(M_1,K):=\eta_1(X_{r,1}),\label{1-shot-proof-g1}\\
\hat{X}_2&=&g_2(M_1,M_2,K):=\eta_2(X_{r,1},X_{r,2}),\label{1-shot-proof-g2}
\end{IEEEeqnarray}
such that $\eta_1(.)$ and $\eta_2(.,.)$ are deterministic functions.

Now, consider the fact that the set of constraints in \eqref{Dj-1-shot}--\eqref{Pj-1-shot}, \eqref{1-shot-proof-R1}, \eqref{1-shot-proof-R2}, \eqref{1-shot-proof-R3} with Markov chains in \eqref{1-shot-proof-Markov1}--\eqref{1-shot-proof-Markov2} and deterministic functions in \eqref{1-shot-proof-g1}--\eqref{1-shot-proof-g2} constitute an iRDP region, denoted by $\overline{\mathcal{RDP}}$, which is the set of all tuples $(\mathsf{R},\mathsf{D},\mathsf{P})$ such that
\begin{IEEEeqnarray}{rCl}
R_1 &\geq & I(X_1;X_{r,1}),\label{rate0-ISh}\\
R_2 &\geq & I(X_1,X_2;X_{r,2}|X_{r,1}),\label{rate1-proof}\\
R_3 &\geq & I(X_1,X_2,X_3;\hat{X}_3|X_{r,1},X_{r,2}),\label{rate2-proof}\\
D_j &\geq & \mathbbm{E}[\|X_j-\hat{X}_j\|^2], \qquad\qquad\;\;\;\; j=1,2,3,\label{distortion-proof}\\
P_j &\geq & \phi_j(P_{\hat{X}_{1}\ldots\hat{X}_{j-1} X_{j}},P_{\hat{X}_1\ldots\hat{X}_j}), \;\;\; j=1,2,3,\label{perception-proof}
\end{IEEEeqnarray}
for auxiliary random variables $(X_{r,1},X_{r,2})$ and $(\hat{X}_1,\hat{X}_2,\hat{X}_3)$ satisfying the following
\begin{IEEEeqnarray}{rCl}
\hat{X}_1&=& \eta_1(X_{r,1}), \;\;\hat{X}_2=\eta_2(X_{r,1},X_{r,2})\label{Markov1-Ish}\\
X_{r,1}&\to& X_1\to (X_2,X_3),\label{Markov2-Ish}\\
X_{r,2} &\to & (X_1,X_2,X_{r,1})\to X_3.\label{Markov3-Ish0}
\end{IEEEeqnarray}
for some deterministic functions $\eta_1(.)$ and $\eta_2(.,.)$.

 Comparing the two regions $\overline{\mathcal{RDP}}$ and $\mathcal{RDP}$, we identify the following differences. The Markov chain in \eqref{Markov2-new} is more restricted comparing to \eqref{Markov3-Ish0}. Moreover, the Markov chain \eqref{Markov3-new} does not exist in $\overline{\mathcal{RDP}}$. The following lemma states that $\overline{\mathcal{RDP}}=\mathcal{RDP}$. Now, for a given $(\mathsf{D},\mathsf{P})$, let $\overline{\mathcal{R}}(\mathsf{D},\mathsf{P})$ denote the set of rate tuples $\mathsf{R}$ such $(\mathsf{R},\mathsf{D},\mathsf{P})\in \overline{\mathcal{RDP}}$, then this lemma implies that $\overline{\mathcal{R}}(\mathsf{D},\mathsf{P})=\mathcal{R}(\mathsf{D},\mathsf{P})$ which completes the proof of the outer bound.
 
 We conclude this section by the following lemma.

\begin{lemma}\label{ISh-thm} For first-order Markov sources, we have
\begin{IEEEeqnarray}{rCl}
\mathcal{RDP}=\overline{\mathcal{RDP}}.
\end{IEEEeqnarray}

\end{lemma}

\begin{IEEEproof} This result for the scenario without perception constraint has been similarly observed in Eq. (12) of \cite{Skoglund}. The proof in this section is provided for completeness.

First, notice that the set of Markov chains in \eqref{Markov2-new}--\eqref{Markov4-new} is more restricted than the ones in \eqref{Markov2-Ish}--\eqref{Markov3-Ish0}, hence $\mathcal{RDP}\subseteq \overline{\mathcal{RDP}}$. Now, it remains to prove that $\overline{\mathcal{RDP}}\subseteq \mathcal{RDP}$. Consider the following facts
\begin{enumerate}
\item The distortion constraints in~\eqref{distortion-proof} depend only on the joint distribution of $(X_j, \hat{X}_j)$, and thus on the joint distribution of $(X_j,X_{r,1},\ldots,X_{r,j})$. So, imposing the Markov chain $X_{r,2}\to (X_2,X_{r,1})\to X_1$ does not affect the expected distortion $\mathbbm{E}[\|X_2-\hat{X}_2\|^2]$ since it does not depend on the joint distribution of $X_1$ with $(X_{r,1},X_{r,2},X_2)$. A similar argument holds for other frames;

\item The perception constraints in~\eqref{perception-proof} depend on the joint distributions  $P_{\hat{X}_{1}\ldots\hat{X}_{j-1} X_{j}}$ and $P_{\hat{X}_1,\ldots,\hat{X}_j}$ (hence on $P_{X_{r,1}\ldots X_{r,j}}$). Thus, imposing $X_{r,2}\to (X_2,X_{r,1})\to X_1$ does not affect $\phi_2(P_{\hat{X}_1X_2},P_{\hat{X}_1\hat{X}_2})$ since it does not depend on the joint distribution of $X_1$ with $(X_{r,1},X_{r,2},X_2)$. A similar argument holds for other frames;
\item Moreover, the rate constraints in \eqref{rate1-proof} and \eqref{rate2-proof} would be further lower bounded by 
\begin{IEEEeqnarray}{rCl}
R_2 &\geq & I(X_1,X_2;X_{r,2}|X_{r,1})\nonumber\\&\geq & I(X_2;X_{r,2}|X_{r,1}),\label{lower-bound-Ish1}\\
R_3 &\geq & I(X_1,X_2,X_3;\hat{X}_3|X_{r,1},X_{r,2})\nonumber\\&\geq & I(X_3;\hat{X}_3|X_{r,1},X_{r,2}).\label{lower-bound-Ish2}
\end{IEEEeqnarray}
Thus, the set of rate constraints is optimized by the set of Markov chains \eqref{Markov2-new}--\eqref{Markov4-new}.
\item The mutual information terms $I(X_1;X_{r,1})$, $I(X_2;X_{r,2}|X_{r,1})$ and $I(X_3;\hat{X}_3|X_{r,1},X_{r,2})$ depend on distributions  $P_{X_1X_{r,1}}$, $P_{X_{r,1}X_{r,2}X_2}$ and $P_{X_3\hat{X}_3X_{r,1}X_{r,2}}$, respectively. So, these distributions should be preserved by the set of Markov chains. The first two distributions are preserved by the choice of~\eqref{Markov1-new}--\eqref{Markov2-new}. 
Now, since we have first-order Markov sources, preserving the joint distributions of $P_{X_{r,1}X_1}$ and $P_{X_{r,1}X_{r,2}X_2}$ is sufficient to preserve the distribution $P_{X_{r,1}X_{r,2}X_3}$. So, preserving the joint distribution of $P_{\hat{X}_3 X_{r,1}X_{r,2}}$ is sufficient to keep $I(X_3;\hat{X}_3|X_{r,1},X_{r,2})$ unchanged.
\end{enumerate}

Considering the above four facts, without loss of optimality, one can impose the following Markov chains
\begin{IEEEeqnarray}{rCl}
X_{r,1}&\to& X_1\to (X_2,X_3),\\
X_{r,2}&\to& (X_2,X_{r,1})\to (X_1,X_3),\\\hat{X}_3&\to& (X_3,X_{r,1},X_{r,2})\to (X_1,X_2).
\end{IEEEeqnarray}
This  concludes the proof of the lemma.

\end{IEEEproof}

\section{Gauss-Markov Source Model}\label{Gauss-Markov-app}

In this section, we prove that for Gaussian sources, jointly Gaussian reconstructions are optimal. 

\begin{proposition}\label{Gaussian-optimality-thm}
 For the Gauss-Markov source model, any  tuple $(\mathsf{R},\mathsf{D},\mathsf{P})\in \mathcal{RDP}$ can be attained by a jointly Gaussian distribution over $(X_{r,1},X_{r,2},X_{r,3})$ and identity mappings for $\eta_j(\cdot)$ in Definition~\ref{iRDP-region}. 
\end{proposition}
 First, notice that a proof for the setting without perception constraint is provided in \cite{Ashishproof}. The following proof is different from \cite{Ashishproof} in some steps and also involves the perception constraint.

For a given tuple $(\mathsf{R},\mathsf{D},\mathsf{P})\in\mathcal{RDP}$, let $X^*_{r,1}$, $X_{r,2}^*$, $\hat{X}^*_1=\eta_1(X_{r,1}^*)$, $\hat{X}^*_2=\eta_2(X_{r,1}^*,X_{r,2}^*)$ and $\hat{X}^*_3$ be random variables satisfying \eqref{Markov1-new}--\eqref{Markov3-new}. Let $P_{\hat{X}^G_1|X_1}$, $P_{\hat{X}^G_2|\hat{X}^G_1X_2}$ and $P_{\hat{X}^G_3|\hat{X}^G_1\hat{X}_2^GX_3}$ be jointly Gaussian distributions such that the following conditions are satisfied.  
\begin{IEEEeqnarray}{rCl}
\text{cov}(\hat{X}_1^G,X_1)&=&\text{cov}(\hat{X}^*_1,X_1),\label{cov-pres2}\\
\text{cov}(\hat{X}_1^G,\hat{X}^G_2,X_2)&=&\text{cov}(\hat{X}^*_1,\hat{X}^*_2,X_2),\label{cov-pres3}\\
\text{cov}(\hat{X}_1^G,\hat{X}^G_2,\hat{X}_3^G,X_3)&=&\text{cov}(\hat{X}^*_1,\hat{X}^*_2,\hat{X}^*_3,X_3),\label{cov-pres4}
\end{IEEEeqnarray}
In general, the Gaussian random variables which satisfy the constraints in \eqref{cov-pres2}--\eqref{cov-pres4} can be written in the following format
\begin{IEEEeqnarray}{rCl}
X_1&=& \nu \hat{X}_1^G+Z_1,\label{rv-relation1}\\
\hat{X}_2^G&=& \omega_1 \hat{X}_1^G+\omega_2X_2+Z_2,\label{rv-relation3}\\
\hat{X}_3^G&=& \tau_1 \hat{X}_1^G+\tau_2\hat{X}_2^G+\tau_3X_3+Z_3,\label{rv-relation-end}
\end{IEEEeqnarray}
for some real $\nu$, $\omega_1$, $\omega_2$, $\tau_1$, $\tau_2$, $\tau_3$ where $\hat{X}^G_1\sim \mathcal{N}(0,\sigma^2_{\hat{X}_1^G})$, $\hat{X}^G_2\sim \mathcal{N}(0,\sigma^2_{\hat{X}_2^G})$, $Z_1$, $Z_2$ and $Z_3$ are Gaussian random variables with zero mean and variances $\alpha_1^2, \alpha_2^2, \alpha_3^2$, independent of $\hat{X}_1^G$,  $(\hat{X}_1^G,X_2)$ and $(\hat{X}_1^G,\hat{X}_2^G,X_3)$, respectively.

We explicitly derive the coefficients $\nu$, $\omega_1$, $\omega_2$, $\tau_1$, $\tau_2$ and  $\tau_3$ in the following. Multiplying both sides of~\eqref{rv-relation1} by $\hat{X}_1^G$ and taking an expectation, we get
\begin{IEEEeqnarray}{rCl}
\mathbbm{E}[X_1\hat{X}_1^G]=\nu \sigma^2_{\hat{X}_1^G}.
\end{IEEEeqnarray}
According to~\eqref{cov-pres2}, the above equation can be written as follows
\begin{IEEEeqnarray}{rCl}
\mathbbm{E}[X_1\hat{X}_1^*]=\nu \mathbbm{E}[\hat{X}^{*2}_1].\label{equation-nu-1}
\end{IEEEeqnarray}
Multiplying both sides of~\eqref{rv-relation3} by the vector $[\hat{X}_1^G\;\; X_2]$ and taking an expectation, we have
\begin{IEEEeqnarray}{rCl}
&&\hspace{-1cm}[\mathbbm{E}[\hat{X}_1^G\hat{X}_2^G]\;\; \mathbbm{E}[X_2\hat{X}_2^G]]=\nonumber\\&&[\omega_1\;\;\omega_2]\begin{pmatrix}\sigma^2_{\hat{X}_1^G} & \mathbbm{E}[X_2\hat{X}_1^G]\\ \mathbbm{E}[X_2\hat{X}_1^G] & \sigma_2^2\end{pmatrix}.
\end{IEEEeqnarray}
Considering the fact that $\mathbbm{E}[X_2\hat{X}_1^G]=\rho_1\mathbbm{E}[X_1\hat{X}_1^G]$ and according to~\eqref{cov-pres3}, the above equation can be written as follows
\begin{IEEEeqnarray}{rCl}
&&\hspace{-1cm}[\mathbbm{E}[\hat{X}_1^*\hat{X}_2^*]\;\; \mathbbm{E}[X_2\hat{X}_2^*]]=\nonumber\\&&[\omega_1\;\;\omega_2]\begin{pmatrix}\mathbbm{E}[\hat{X}_1^{*2}] & \rho_1\mathbbm{E}[X_1\hat{X}_1^*]\\ \rho_1\mathbbm{E}[X_1\hat{X}_1^*] & \sigma_2^2\end{pmatrix}.\label{equation-omega-2}
\end{IEEEeqnarray}
Similarly, multiplying both sides of~\eqref{rv-relation-end} by the vector $[\hat{X}_1^G\;\;\hat{X}_2^G\;\;X_3]$, taking an expectation and considering~\eqref{cov-pres4}, we get 
\begin{IEEEeqnarray}{rCl}
&&[\mathbbm{E}[\hat{X}_1^*\hat{X}_3^*]\;\;\mathbbm{E}[\hat{X}_2^*\hat{X}_3^*]\;\;\mathbbm{E}[X_3\hat{X}_3^*]]=[\tau_1\;\;\tau_2\;\;\tau_3]\nonumber\\&&\hspace{0.5cm}\begin{pmatrix}\mathbbm{E}[\hat{X}^{*2}_1] & \mathbbm{E}[\hat{X}^*_1\hat{X}^*_2] & \rho_1\rho_2\mathbbm{E}[X_1\hat{X}^*_1]\\\mathbbm{E}[\hat{X}_1^*\hat{X}^*_2] & \mathbbm{E}[\hat{X}^{*2}_2] & \rho_2\mathbbm{E}[X_2\hat{X}_2^*]\\\rho_1\rho_2\mathbbm{E}[X_1\hat{X}_1^*] & \rho_2\mathbbm{E}[X_2\hat{X}_2^*] & \mathbbm{E}[\hat{X}^{*2}_3]\end{pmatrix}.\nonumber\\\label{equation-tau-3}
\end{IEEEeqnarray}
Solving equations~\eqref{equation-nu-1},~\eqref{equation-omega-2} and~\eqref{equation-tau-3}, we get
\begin{IEEEeqnarray}{rCl}
\sigma^2_{\hat{X}_1^G}&=&\mathbb{E}[\hat{X}^{*2}_1],\\
\nu &=& \frac{\mathbb{E}[X_1\hat{X}^*_1]}{\mathbb{E}[\hat{X}^{*2}_1]},\\
\alpha_1^2&=& \sigma_1^2-\frac{\mathbb{E}[X_1\hat{X}^*_1]}{\mathbb{E}[\hat{X}^{*2}_1]},\\
\omega_1&=& \frac{\nu\rho_1\mathbbm{E}[\hat{X}_1^*\hat{X}_2^*]-\mathbbm{E}[X_2\hat{X}^*_2]}{\nu^2\rho_1^2\sigma^2_{\hat{X}_1^G}-\sigma_2^2},\\
\omega_2 &=& \frac{\nu\rho_1\sigma_{\hat{X}_1^G}^2\mathbbm{E}[X_2\hat{X}^*_2]-\sigma_2^2\mathbbm{E}[\hat{X}_1^*\hat{X}^*_2]}{\nu^2\rho_1^2\sigma^4_{\hat{X}_1^G}-\sigma_2^2\sigma^2_{\hat{X}_1^G}},\\
\alpha_2^2 &=& \mathbbm{E}[\hat{X}_2^{*2}]-\alpha_2^2\sigma^2_{\hat{X}_1^G}-\omega_2^2\sigma_2^2-2\omega_1\omega_2\rho_1\nu \sigma_{\hat{X}_1^G}^2.\nonumber\\
\end{IEEEeqnarray}
For the third step, the coefficients and noise variance of \eqref{rv-relation-end} are given as follows
\begin{IEEEeqnarray}{rCl}
&&\hspace{-0.7cm}[\tau_1\;\;\tau_2\;\;\tau_3] =[\mathbbm{E}[\hat{X}_1^*\hat{X}_3^*]\;\;\mathbbm{E}[\hat{X}_2^*\hat{X}_3^*]\;\;\mathbbm{E}[X_3\hat{X}_3^*]]\cdot\nonumber\\&&\hspace{-0.5cm}\begin{pmatrix}\mathbbm{E}[\hat{X}^{*2}_1] & \mathbbm{E}[\hat{X}^*_1\hat{X}^*_2] & \rho_1\rho_2\mathbbm{E}[X_1\hat{X}^*_1]\\\mathbbm{E}[\hat{X}_1^*\hat{X}^*_2] & \mathbbm{E}[\hat{X}^{*2}_2] & \rho_2\mathbbm{E}[X_2\hat{X}_2^*]\\\rho_1\rho_2\mathbbm{E}[X_1\hat{X}_1^*] & \rho_2\mathbbm{E}[X_2\hat{X}_2^*] & \mathbbm{E}[\hat{X}^{*2}_3]\end{pmatrix}^{-1},\nonumber\\\\
\alpha_3^2 &= & \mathbbm{E}[\hat{X}^{*2}_3]-\tau_1^2\mathbbm{E}[\hat{X}^{*2}_1]-\tau_2^2\mathbbm{E}[\hat{X}^{*2}_2]-\tau_3^2\mathbbm{E}[X_3^2]\nonumber\\&&\hspace{1cm}-2\tau_1\tau_2\mathbbm{E}[\hat{X}_1^*\hat{X}_2^*]-2\tau_1\tau_3\rho_1\rho_2\mathbbm{E}[X_1\hat{X}_1^*]\nonumber\\&&\hspace{1cm}-2\tau_2\tau_3\rho_2\mathbbm{E}[X_2\hat{X}_2^*],
\end{IEEEeqnarray}
where $(.)^{-1}$ denotes the inverse of a matrix.

Now, we look at the rate constraints.

\underline{\textit{Rate Constraints}}:

Consider the rate constraint of the first step as follows
\begin{IEEEeqnarray}{rCl}
R_1&\geq &I(X_1;X^*_{r,1})\\
&=& H(X_1)-H(X_1|X^*_{r,1})\label{rate1-step1}\\
&\geq & H(X_1)-H(X_1|\hat{X}^*_1)\label{rate1-step3}\\
&=& H(X_1)-H(X_1-\mathbbm{E}[X_1|\hat{X}^*_1]|\hat{X}^*_1)\\
&\geq& H(X_1)-H(X_1-\mathbbm{E}[X_1|\hat{X}^*_1])\\
&\geq & H(X_1)-H(X_1-\mathbbm{E}[X_1|\hat{X}^G_1])\label{lemma2-step2}\\
&= & H(X_1)-H(X_1-\mathbbm{E}[X_1|\hat{X}^G_1]|\hat{X}^G_1)\label{rate1-step4}\\
&=& I(X_1;\hat{X}^G_1)
\end{IEEEeqnarray}
where 
\begin{itemize}
\item \eqref{rate1-step3} follows because $\hat{X}^*_1$ is a function of $X^*_{r,1}$;
    \item \eqref{lemma2-step2} follows because for a given covariance matrix in \eqref{cov-pres2}, the Gaussian distribution maximizes the differential entropy;
    \item \eqref{rate1-step4} follows because the MMSE is uncorrelated from the data and since the random variables are Gaussian, the MMSE would be independent of the data.
\end{itemize}
Next, consider the rate constraint of the second step as the following
\begin{IEEEeqnarray}{rCl}
R_2 &\geq & I(X_2;X^*_{r,2}|X^*_{r,1})\\
&=& H(X_2|X^*_{r,1})-H(X_2|X^*_{r,1},X^*_{r,2})\\
&\geq & H(X_2|X^*_{r,1})-H(X_2|\hat{X}^*_1,\hat{X}^*_2)\label{rate2-step1}\\
&\geq & H(X_2|X^*_{r,1})-H(X_2|\hat{X}^G_1,\hat{X}^G_2)\label{rate2-step4}\\
&=& H(\rho_1X_1+N_1|X^*_{r,1})-H(X_2|\hat{X}^G_1,\hat{X}^G_2)\\
&\geq & \frac{1}{2}\log \left(\rho_1^2 2^{2H(X_1|X^*_{r,1})}+2^{2H(N_1)}\right)\nonumber\\&&\hspace{0.5cm}-H(X_2|\hat{X}^G_1,\hat{X}^G_2)\label{rate2-step2}\\
&\geq  & \frac{1}{2}\log \left(\rho_1^2 2^{-2R_1}2^{2H(X_1)}+2^{2H(N_1)}\right)\nonumber\\&&\hspace{0.5cm}-H(X_2|\hat{X}^G_1,\hat{X}^G_2),\label{rate2-step3}
\label{2nd-step-lower-bound}
\end{IEEEeqnarray}
where
\begin{itemize}
    \item \eqref{rate2-step1} follows because $\hat{X}^*_1$ and $\hat{X}^*_2$ are deterministic functions of $X^*_{r,1}$ and $(X_{r,1}^*,X_{r,2}^*)$, respectively;
    \item \eqref{rate2-step4} follows because for a given covariance matrix in \eqref{cov-pres3}, the Gaussian distribution maximizes the differential entropy;
    \item \eqref{rate2-step2} follows from entropy power inequality (EPI) (see pp. 22 of \cite{KimElGamal});
    \item \eqref{rate2-step3} follows from \eqref{rate1-step1}.
\end{itemize}
Similarly, consider the rate constraint of the third frame as the following,
\begin{IEEEeqnarray}{rCl}
    R_3 &\geq & I(X_3;\hat{X}^*_3|X^*_{r,1},X^*_{r,2})\\
    &=& H(X_3|X^*_{r,1},X^*_{r,2})-H(X_3|X^*_{r,1},X^*_{r,2},\hat{X}^*_3)\\
    &\geq & H(X_3|X^*_{r,1},X^*_{r,2})-H(X_3|\hat{X}^*_1,\hat{X}^*_2,\hat{X}^*_3)\\
    &\geq & H(X_3|X^*_{r,1},X^*_{r,2})-H(X_3|\hat{X}^G_1,\hat{X}^G_2,\hat{X}^G_3)\\
    &=& H(\rho_2X_2+N_2|X^*_{r,1},X^*_{r,2})-H(X_3|\hat{X}^G_1,\hat{X}^G_2,\hat{X}^G_3)\nonumber\\\\
    &\geq & \frac{1}{2}\log \left(\rho_2^2 2^{2H(X_2|X^*_{r,1},X^*_{r,2})}+2^{2H(N_2)}\right)\nonumber\\&&\hspace{0.5cm}-H(X_3|\hat{X}^G_1,\hat{X}^G_2,\hat{X}^G_3)\\
    &\geq  & \frac{1}{2}\log \left(\rho_2^2 2^{-2R_2}2^{2H(X_2|X^*_{r,1})}+2^{2H(N_2)}\right)\nonumber\\&&\hspace{0.5cm}-H(X_3|\hat{X}^G_1,\hat{X}^G_2,\hat{X}^G_3)\\
    &\geq& \frac{1}{2}\log \Big(\rho_1^2\rho_2^2 2^{-2R_1-2R_2}2^{2H(X_1)}+\rho_2^2 2^{-2R_2}2^{2H(N_1)}+\nonumber\\&&\hspace{2cm}2^{2H(N_2)}\Big)-H(X_3|\hat{X}_1^G,\hat{X}_2^G,\hat{X}^G_3)\nonumber\\\label{rate3-Gaussian-proof-opt}
\end{IEEEeqnarray}

Next, we look at the distortion constraint.

\underline{\textit{Distortion Constraint}}: The choices in \eqref{cov-pres2}--\eqref{cov-pres4} imply that
\begin{IEEEeqnarray}{rCl}
D_j\geq \mathbbm{E}[\|X_j-\hat{X}^*_{j}\|^2]=\mathbbm{E}[\|X_j-\hat{X}^G_{j}\|^2],\qquad j=1,2,3.\nonumber\\\label{distortion-Gaussian-proof-opt}
\end{IEEEeqnarray}

Finally, we look at the perception constraint

\underline{\textit{Perception Constraint}}:

Define the following distribution
\begin{IEEEeqnarray}{rCl}
P_{U^*V^*}:= \arg\inf_{\substack{\tilde{P}_{UV}:\\\tilde{P}_U=P_{X_1}\\\tilde{P}_{V}=P_{\hat{X}^*_{1}}}}\mathbbm{E}_{\tilde{P}}[\|U-V\|^2].\label{step4}
\end{IEEEeqnarray}
Now, define $P_{U^GV^G}$ to be a Gaussian joint distribution with the following covariance matrix \begin{IEEEeqnarray}{rCl}\text{cov}(U^G,V^G)=\text{cov}(U^*,V^*).\label{covariance-G-match}\end{IEEEeqnarray}

Then, we have the following set of inequalities: 

\begin{IEEEeqnarray}{rCl}
P_1&\geq& W_2^2(P_{X_1},P_{\hat{X}^*_1})\\
&=& \inf_{\substack{\tilde{P}_{UV}:\\\tilde{P}_U=P_{X_1}\\\tilde{P}_{V}=P_{\hat{X}^*_{1}}}}\mathbbm{E}_{\tilde{P}}[\|U-V\|^2]\label{step5}\\
&=& \mathbbm{E}[\|U^*-V^*\|^2]\label{P-step}\\
&=& \mathbbm{E}[\|U^G-V^G\|^2]\label{step6}\\
&\geq &  W_2^2(P_{U^G},P_{V^G})\\
&=& \inf_{\substack{\hat{P}_{UV}:\\\hat{P}_U=P_{U^G}\\\hat{P}_V=P_{V^G}}}\mathbbm{E}_{\hat{P}}[\|U-V\|^2]\\
&=& \inf_{\substack{\hat{P}_{UV}:\\\hat{P}_U=P_{X_1}\\\hat{P}_V=P_{\hat{X}^G_1}}}\mathbbm{E}_{\hat{P}}[\|U-V\|^2]\label{step7}\\
&=& W_2^2(P_{X_1},P_{\hat{X}^G_1}),\label{step8}
\end{IEEEeqnarray}
where
\begin{itemize}
\item \eqref{P-step} follows from the definition in \eqref{step4};
\item \eqref{step6} follows from~\eqref{covariance-G-match} which implies that $(U^*,V^*)$ and $(U^G,V^G)$ have the same second-order statistics;
\item \eqref{step7} follows because $P_{V^G}=P_{\hat{X}^G_1}$ which is justified in the following. First, notice that both $P_{V^G}$ and $P_{\hat{X}^G_1}$ are Gaussian distributions. Denote the variance of $V^G$ by $\sigma_{V^G}^2$ and recall that the variance of $\hat{X}^G_1$ is denoted by $\sigma^2_{\hat{X}_1^G}$. According to~\eqref{covariance-G-match}, $\sigma_{V^G}^2$ is equal to the variance of $V^*$. Also, from~\eqref{step4}, we know that $P_{V^*}=P_{\hat{X}^*_1}$, hence the variances of $V^*$ and $\hat{X}^*_1$ are the same. On the other side, according to \eqref{cov-pres2}, we know that the variance of $\hat{X}^*_1$ is equal to $\sigma^2_{\hat{X}_1^G}$. Thus, we conclude that $\sigma^2_{\hat{X}_1^G}=\sigma_{V^G}^2$, which yields $P_{V^G}=P_{\hat{X}^G_1}$. A similar argument shows that $P_{U^G}=P_{X_1}$.
\end{itemize}
A similar argument holds for the perception constraint of the second and third steps for both PLFs.

 Thus, we have proved the set of Gaussian auxiliary random variables $(\hat{X}_1^G,\hat{X}_2^G,\hat{X}_3^G)$ given in \eqref{rv-relation1}--\eqref{rv-relation-end} where the coefficients are chosen according to distortion-perception constraints, provides an outer bound to $\mathcal{RDP}$ which is the set of all tuples $(\mathsf{R},\mathsf{D},\mathsf{P})$ such that
\begin{IEEEeqnarray}{rCl}
R_1&\geq & I(X_1;\hat{X}_1^G),\label{R1-Gaussian-outer}\\
R_2&\geq & \frac{1}{2}\log \left(\rho_1^2 2^{-2R_1}2^{2H(X_1)}+2^{2H(N_1)}\right)\nonumber\\&&\hspace{0.5cm}-H(X_2|\hat{X}^G_1,\hat{X}^G_2),\label{R2-Gaussian-outer}\\
R_3&\geq & \frac{1}{2}\log \Big(\rho_1^2\rho_2^2 2^{-2R_1-2R_2}2^{2H(X_1)}+\rho_2^2 2^{-2R_2}2^{2H(N_1)}\nonumber\\&&\hspace{1cm}+2^{2H(N_2)}\Big)-H(X_3|\hat{X}_1^G,\hat{X}_2^G,\hat{X}^G_3),\\
D_j&\geq &\mathbbm{E}[\|X_j-\hat{X}^G_{j}\|^2],\qquad\qquad  j=1,2,3\label{distortion-Gaussian-outer1}\\
P_j&\geq & W_2^2(P_{X_1\ldots X_j},P_{\hat{X}_1^G\ldots\hat{X}_j^G}).\label{perception-Gaussian-outer2}
\end{IEEEeqnarray}

Now, we need to show that the above RDP region is also an inner bound to $\mathcal{RDP}$. This is simply verified by the following choice.  In iRDP region of \eqref{MaIsh-rate1}--\eqref{Markov4-new}, choose the following:
\begin{IEEEeqnarray}{rCl}
X_{r,j}=\hat{X}_j=\hat{X}_j^G,\qquad j=1,2,3,
\end{IEEEeqnarray}
where $(\hat{X}_1^G,\hat{X}_2^G,\hat{X}_3^G)$ satisfy \eqref{rv-relation1}--\eqref{rv-relation-end} with coefficients chosen according to distortion-perception constraints.
The lower bounds on distortion and perception constraints in \eqref{distortion-Gaussian-outer1} and \eqref{perception-Gaussian-outer2} are immediately achieved by this choice. Now, we will look at the rate constraints. The achievable rate constraint of the first step can be written as follows
\begin{IEEEeqnarray}{rCl}
R_1 &\geq  & I(X_1;\hat{X}_1^G),\label{R1-ach-Gaussian}
\end{IEEEeqnarray}
which immediately coincides with~\eqref{R1-Gaussian-outer}.
The achievable rate of the second step can be written as follows
\begin{IEEEeqnarray}{rCl}
R_2&\geq &I(X_2;\hat{X}^G_2|\hat{X}_1^G)\\
&=& H(X_2|\hat{X}_1^G)-H(X_2|\hat{X}_1^G,\hat{X}_2^G)\\
&=& H(\rho_1X_1+N_1|\hat{X}_1^G)-H(X_2|\hat{X}_1^G,\hat{X}_2^G)\\
&=&  \frac{1}{2}\log (\rho_1^22^{2H(X_1|\hat{X}_1^G)}+2^{2H(N_1)})\nonumber\\&&\hspace{0.5cm}-H(X_2|\hat{X}_1^G,\hat{X}_2^G)\label{ach-step3}\\
&\geq &\frac{1}{2}\log \left(\rho_1^2 2^{-2R_1}2^{2H(X_1)}+2^{2H(N_1)}\right)\nonumber\\&&\hspace{0.5cm}-H(X_2|\hat{X}_1^G,\hat{X}_2^G),\label{opt-ach-Gaussian-step5}
\end{IEEEeqnarray}
where 
\begin{itemize}
    \item  \eqref{ach-step3} follows because EPI holds with ``equality'' for jointly Gaussian distributions (see pp. 22 of \cite{KimElGamal});
    \item \eqref{opt-ach-Gaussian-step5} follows from~\eqref{R2-Gaussian-outer}.
\end{itemize}
Thus, the bound in \eqref{opt-ach-Gaussian-step5} coincides with~\eqref{rate2-step3}. A similar argument holds for the achievable rate of the third frame.

Notice that the above proof (both converse and achievability) can be extended to $T$ frames using the sequential analysis that was presented. Thus,  without loss of optimality, one can restrict to the jointly Gaussian distributions and identity functions $\eta_1(.)$ and $\eta_2(.,.)$ in iRDP region $\mathcal{RDP}$.

\section{Low-rate Regime for the First Frame}\label{low-rate-app}

\medskip

In this section, we prove the following theorem when the first frame is compressed at a low rate. The rate of the second frame is an arbitrary nonnegative value.

\begin{theorem} Let $R_1=\epsilon$ for a sufficiently small $\epsilon>0$ and $R_2$ be an arbitrary nonnegative rate. The achievabale distortions for the second frame, $D_{2,\text{AR}}^0$ (for $0$-PLF-SA), $D_{2,\text{FMD}}^0$ (for $0$-PLF-FMD) and $D_{2,\text{JD}}^0$ (for $0$-PLF-JD) are given by
\begin{IEEEeqnarray}{rCl}
D_{2,\text{SA}}^0 &=& 2\sigma^2 (1-\sqrt{1-2^{-2R_2}}),\\  D_{2,\text{FMD}}^0&=&  2\sigma^2 (1-\sqrt{1-2^{-2R_2}+\rho^22\epsilon\ln 2}),\\D_{2,\text{JD}}^0 &=& 2\sigma^2 (1-\sqrt{1-\rho^2}\sqrt{1-2^{-2R_2}}-\rho^2\sqrt{2\epsilon \ln 2}).\nonumber\\
\end{IEEEeqnarray}
\end{theorem}

To prove the above theorem, we first remind the RDP region of the Gauss-Markov source model. Then, we will look at each PLF separately; $0$-PLF-SA,  $0$-PLF-FMD, and $0$-PLF-JD. For each of these PLFs, we discuss the second step and provide the analysis of the third step for completeness.

Recall the RDP region of the Gauss-Markov model which is the set of all tuples $(\mathsf{R},\mathsf{D},\mathsf{P})$ such that 
\begin{subequations}\label{RDP-Gauss-Markov-app}
\begin{IEEEeqnarray}{rCl}
R_1 &\geq&  I(X_1;\hat{X}_1),\\
 R_2 &\geq&  I(X_2;\hat{X}_{2}|\hat{X}_{1}),\\
R_3 &\geq&  I(X_3;\hat{X}_{3}|\hat{X}_{1},\hat{X}_{2}), \\
D_j &\geq & \mathbbm{E}[\|X_j-\hat{X}_j\|^2],\\
P_j&\geq&  \phi_j(P_{\hat{X}_{1}\ldots\hat{X}_{j-1} X_{j}}, P_{\hat{X}_{1}\ldots\hat{X}_{j-1}\hat{X}_{j}}), \qquad j=1,2,3,\nonumber\\
\end{IEEEeqnarray}
\end{subequations}
for some auxiliary random variables $(\hat{X}_1,\hat{X}_2,\hat{X}_3)$ which satisfy the following Markov chains
\begin{IEEEeqnarray}{rCl}
\hat{X}_{1}&\to& X_1\to (X_2,X_3),\; \hat{X}_{2}\to (X_2,\hat{X}_{1})\to (X_1,X_3),\nonumber\\&& \hat{X}_{3}\to (X_3,\hat{X}_{1},\hat{X}_{2})\to (X_1,X_2).\;\label{Markov-G-app}
\end{IEEEeqnarray}

For the Gauss-Markov source model, the reconstructions that satisfy the Markov chains in~\eqref{Markov-G-app} can be generally written as follows
\begin{IEEEeqnarray}{rCl}
\hat{X}_1&=& \nu X_1+Z_1,\label{X1hat-Gaussian-ach}\\
\hat{X}_2&=& \omega_1 \hat{X}_1+\omega_2X_2+Z_2,\label{X2hat-Gaussian-ach}\\
\hat{X}_3&=& \tau_1\hat{X}_1+\tau_2\hat{X}_2+\tau_3X_3+Z_3,\label{X3hat-Gaussian-ach}
\end{IEEEeqnarray}
where $\hat{X}_j\sim\mathcal{N}(0,\hat{\sigma}^2_j)$ for $j=1,2$, $Z_1$, $Z_2$ and $Z_3$ are independent of $X_1$, $(\hat{X}_1,X_2)$ and $(\hat{X}_1,\hat{X}_2, X_3)$, respectively.

According to~\eqref{RDP-Gauss-Markov-app}, the optimization program of the first step is as follows
\begin{IEEEeqnarray}{rCl}
&&\hspace{0cm}\min_{P_{\hat{X}_1|X_1}} \mathbbm{E}[\|X_1-\hat{X}_1\|^2]\nonumber\\
&&\text{s.t.}\qquad I(X_1;\hat{X}_1)\leq R_1,\nonumber\\
&&\qquad \phi_1(P_{X_1},P_{\hat{X}_1})\leq P_1. 
\end{IEEEeqnarray}

Using the choice in~\eqref{X1hat-Gaussian-ach}, the optimization program of the first step for $P_1=0$ simplifies as follows
\begin{subequations}
\begin{IEEEeqnarray}{rCl}
&&\hspace{0cm}\min_{\nu}\; 2\sigma^2(1-\nu),\label{obj-simplified}\\
&&\text{s.t.}\qquad \nu^2\leq  (1-2^{-2R_1}),\label{1st-step-opt}
\end{IEEEeqnarray}
\end{subequations}
When $R_1=\epsilon$ for a sufficiently small $\epsilon>0$, the solution of the above program is as follows 
\begin{IEEEeqnarray}{rCl}
D_{1}^0=2\sigma^2(1-\sqrt{2\epsilon\ln 2})+O(\epsilon),\label{1st-step-sol}
\end{IEEEeqnarray}
where the optimal choice of $\nu$ is given by
\begin{IEEEeqnarray}{rCl}
\nu=\sqrt{1-2^{-2R_1}}=\sqrt{2\epsilon\ln 2}+O(\epsilon).
\end{IEEEeqnarray}
Next, consider the optimization programs for different steps and PLFs as follows.

\subsection{$0$-PLF-SA}\label{CP-low-rate}

\begin{figure}[t]
\centering
\begin{subfigure}[b]{0.35\textwidth}  \includegraphics[width=\textwidth]{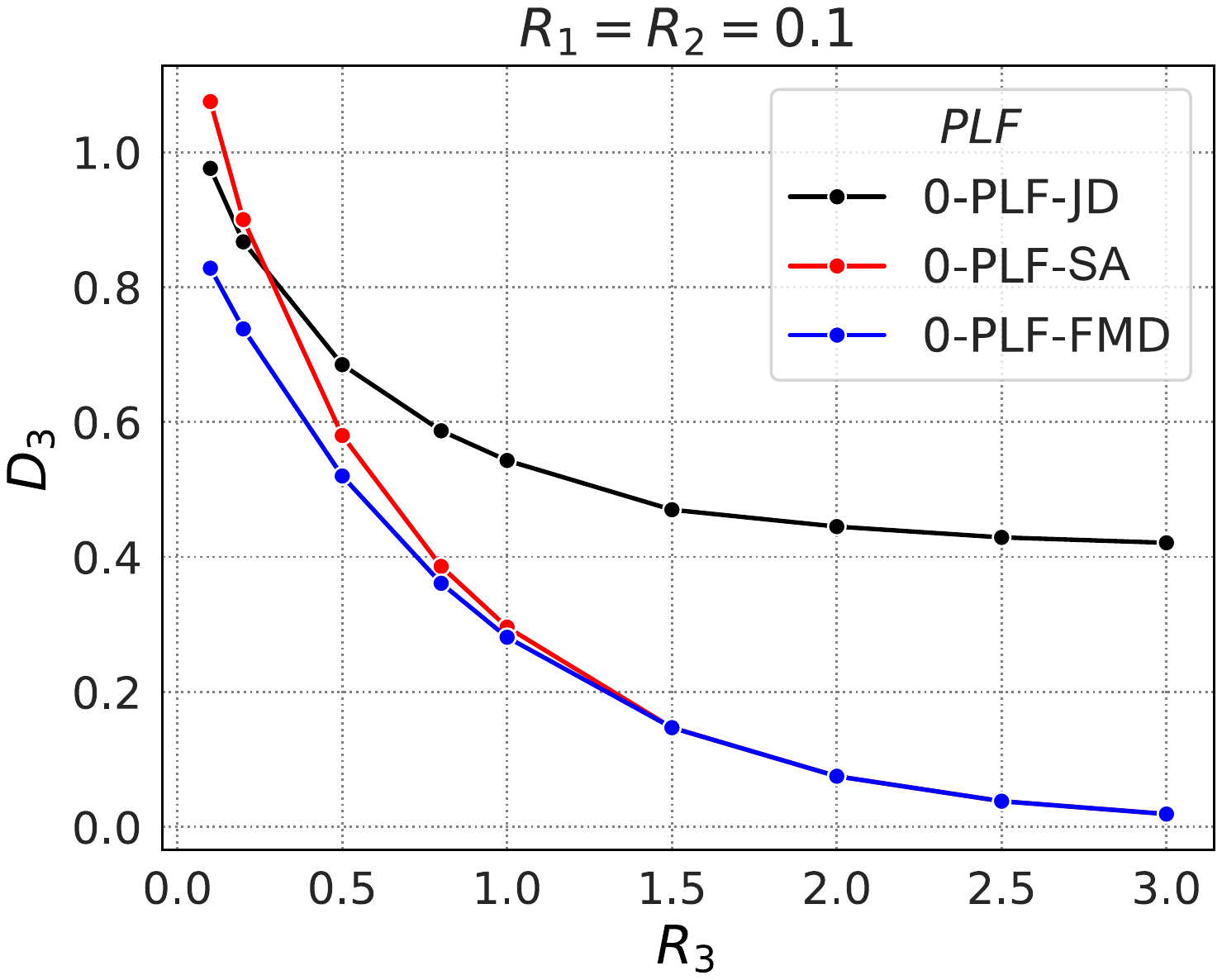}
\end{subfigure}
\centering
\hspace{1cm}\begin{subfigure}[b]{0.35\textwidth}
   \includegraphics[width=\textwidth]{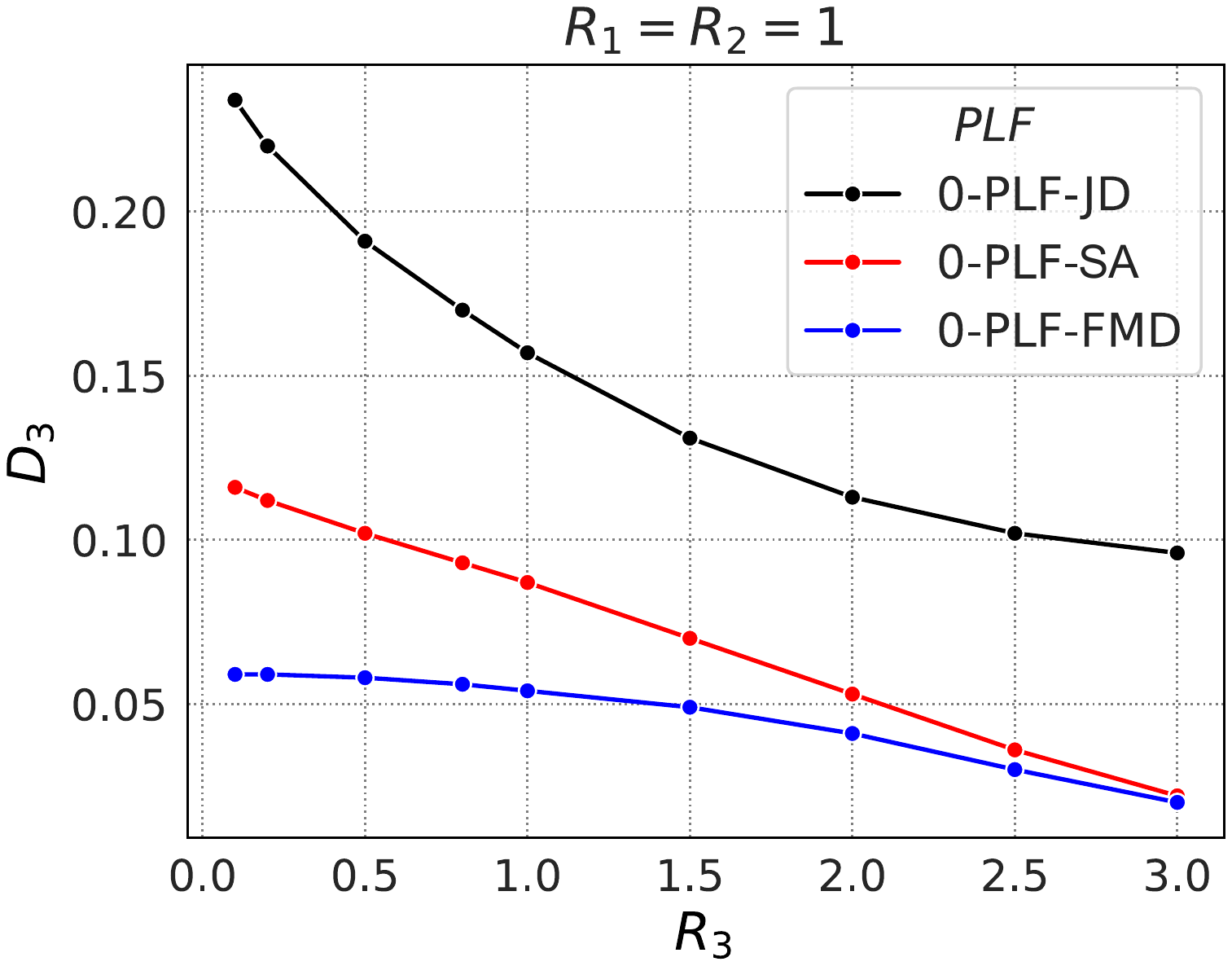}  
\end{subfigure}
\caption{Distortion of the third frame versus its rate for the low-rate regime and $\rho=1$.}
\label{fig:RD3}
\end{figure}

In this section, we provide the optimization programs for different steps of $0$-PLF-SA. For the second step, we are able to provide an approximate solution for the low compression rate, i.e., $R_1=\epsilon$. For the third step, we plot the tradeoff in Fig.~\ref{fig:RD3}.

\underline{\textit{Second Step:}}

The optimization program of the second step is given as follows. 

\begin{proposition}\label{0-PLF-AR-2nd-step} The optimization program of $0$-PLF-SA for the second step can be written as
\begin{subequations}\label{opt-program-JD-simplified-concl}
\begin{IEEEeqnarray}{rCl}
&&\hspace{0cm}\min_{\substack{\omega_1,\omega_2}}\;\; 2\sigma^2-2 \omega_1\rho\nu\sigma^2-2\omega_2\sigma^2,\label{D-G-simplification}\\
&&\text{s.t.}\qquad \omega_2^2(1-\rho^2\nu^22^{-2R_2})\nonumber\\&&\hspace{1.3cm}\leq (1-\omega_1^2-2\omega_1\omega_2\rho\nu)(1-2^{-2R_2}),\;\;\label{R-G-simplification}\\
&&\qquad\;\;\;\; \omega_1+\nu\omega_2\rho=\rho\nu,\label{perception-derivation}\\
&&\qquad\;\;\;\; \nu=\sqrt{1-2^{-2R_1}}.
\end{IEEEeqnarray}
\end{subequations}
\end{proposition}
\begin{IEEEproof} 
According to~\eqref{RDP-Gauss-Markov-app}, the optimization problem of the second step is as follows,
\begin{IEEEeqnarray}{rCl}
&&\hspace{0cm}\min_{P_{\hat{X}_2|X_2\hat{X}_1}} \mathbbm{E}[\|X_2-\hat{X}_2\|^2]\nonumber\\
&&\;\;\;\text{s.t.}\qquad I(X_2;\hat{X}_2|\hat{X}_1)\leq R_2,\nonumber\\
&&\qquad\qquad\;\; P_{\hat{X}_1X_2}=P_{\hat{X}_1\hat{X}_2}. 
\end{IEEEeqnarray}
We proceed with simplifying the rate constraint as follows,
\begin{IEEEeqnarray}{rCl}
R_2&\geq &I(X_2;\hat{X}_2|\hat{X}_1)\\&=&h(\hat{X}_2|\hat{X}_1)-h(Z_2)\label{AR-2nd-R-step1}\\
&=& h(\omega_2X_2+Z_2|\hat{X}_1)-h(Z_2)\label{AR-2nd-R-step2}\\
&=& \frac{1}{2}\log 2^{-2h(Z_2)}\left(\omega_2^22^{2h(X_2|\hat{X}_1)}+2^{2h(Z_2)}\right)\label{AR-2nd-R-step3}\\
&=& \frac{1}{2}\log 2^{-2h(Z_2)}\left(\omega_2^22^{2h(\rho X_1+N_1|\hat{X}_1)}+2^{2h(Z_2)}\right)\nonumber\\\label{AR-2nd-R-step4}\\
&=& \frac{1}{2}\log 2^{-2h(Z_2)} \Bigg(\omega_2^2(\rho^22^{2h( X_1|\hat{X}_1)}+2^{2h(N_1)})\nonumber\\&&\hspace{3cm}+2^{2h(Z_2)}\Bigg)\label{AR-2nd-R-step5}\\
&=& \frac{1}{2}\log 2^{-2h(Z_2)}\Bigg(\omega_2^2(\rho^22^{2h( X_1|\hat{X}_1)}+(1-\rho^2)\sigma^2)\nonumber\\&&\hspace{3cm}+2^{2h(Z_2)}\Bigg)\label{AR-2nd-R-step6}\\
&\geq & \frac{1}{2}\log 2^{-2h(Z_2)}\Bigg(\omega_2^2(\rho^2\sigma^2 2^{-2R_1}+(1-\rho^2)\sigma^2)\nonumber\\&&\hspace{3cm}+2^{2h(Z_2)}\Bigg),\label{AR-2nd-R-step7}
\end{IEEEeqnarray}
where
\begin{itemize}
\item~\eqref{AR-2nd-R-step1} and~\eqref{AR-2nd-R-step2} follow from~\eqref{X2hat-Gaussian-ach};
\item~\eqref{AR-2nd-R-step3} and~\eqref{AR-2nd-R-step5} follow because Entropy Power Inequality (EPI)  (see pp. 22 of \cite{KimElGamal}) holds with equality for Gaussian sources;
\item~\eqref{AR-2nd-R-step4} follows from~\eqref{Gaus-def2} where $X_2=\rho X_1+N_1$;
\item~\eqref{AR-2nd-R-step7} follows from the rate constraint of the first step, i.e., $R_1\geq I(X_1;\hat{X}_1)$.
\end{itemize}
Inequality~\eqref{AR-2nd-R-step7} can be further simplified as follows,
\begin{IEEEeqnarray}{rCl}
&&\hspace{-1cm}(\omega_2^2(\rho^2\sigma^22^{-2R_1}+(1-\rho^2)\sigma^2))2^{-2R_2}\nonumber\\&\geq &(1-2^{-2R_2})2^{2h(Z_2)}\\
&=&(1-2^{-2R_2})\cdot (1-\omega_1^2-\omega_2^2-2\omega_1\omega_2\nu\rho)\sigma^2.
\end{IEEEeqnarray}
Considering that $\nu=\sqrt{1-2^{-2R_1}}$ and re-arranging the terms in the above inequality, we get to constraint in~\eqref{R-G-simplification}. 

The objective function in~\eqref{D-G-simplification} can be obtained as follows,
\begin{IEEEeqnarray}{rCl}
\mathbbm{E}[\|X_2-\hat{X}_2\|^2]&=&2\sigma^2-2\mathbbm{E}[X_2\hat{X}_2]\\
&=& 2\sigma^2-2(\rho\nu\omega_1+\omega_2)\sigma^2,
\end{IEEEeqnarray}
where the last equality follows from~\eqref{X1hat-Gaussian-ach} and~\eqref{X2hat-Gaussian-ach}.

The derivation of the constraint in~\eqref{perception-derivation} is as follows. We multiply both sides of~\eqref{X1hat-Gaussian-ach} and~\eqref{X2hat-Gaussian-ach} by $X_2$ and $\hat{X}_1$, respectively, and take an expectation from both sides. Thus, we have 
\begin{IEEEeqnarray}{rCl}
\mathbbm{E}[X_2\hat{X}_1]&=&\nu\mathbbm{E}[X_1X_2]=\nu\rho\sigma^2,\label{P-simp1}\\
\mathbbm{E}[\hat{X}_1\hat{X}_2]&=& \omega_1\sigma^2+\omega_2\mathbbm{E}[X_2\hat{X}_1].\label{P-simp2}
\end{IEEEeqnarray}
Notice that the perception constraint $P_{X_2\hat{X}_1}=P_{\hat{X}_2\hat{X}_1}$ implies that $\mathbbm{E}[\hat{X}_1\hat{X}_2]=\mathbbm{E}[X_2\hat{X}_1]$ which together with~\eqref{P-simp1} and~\eqref{P-simp2} yields the constraint in~\eqref{perception-derivation}.
\end{IEEEproof}

Now, we provide an approximate solution for the optimization program when the first frame is compressed at a low rate, i.e., $R_1=\epsilon$ where $\epsilon$ is sufficiently small. In this case, we have
\begin{IEEEeqnarray}{rCl}
1-2^{-2R_1}&=&2\epsilon\ln 2+O(\epsilon^2),\\
\nu&=& \sqrt{2\epsilon\ln 2}+O(\epsilon),\label{nu-1st}
\end{IEEEeqnarray}
so the optimization program of the second step in~\eqref{opt-program-JD-simplified-concl} simplifies as follows
\begin{subequations}\label{program-mediumR}
\begin{IEEEeqnarray}{rCl}
&&\hspace{0cm}\min_{\substack{\omega_1,\omega_2}}\;\; 2\sigma^2-2 \omega_1\rho\sigma^2\sqrt{2\epsilon\ln 2+O(\epsilon^2)}-2\omega_2\sigma^2,\label{D-mediumR}\\
&&\text{s.t.}\;\; \omega_2^2(1-\rho^22^{-2R_2}(2\epsilon\ln 2+O(\epsilon^2)))\leq \nonumber\\&&\hspace{0.5cm} (1-\omega_1^2-2\omega_1\omega_2\rho(2\epsilon\ln 2+O(\epsilon^2)))(1-2^{-2R_2}),\nonumber\\\label{R-mediumR}\\
&&\qquad\;\;\;\; \omega_1+\nu\omega_2\rho=\rho\nu.\label{P-mediumR}
\end{IEEEeqnarray}
\end{subequations}
Notice that~\eqref{P-mediumR} and~\eqref{nu-1st} imply that $\omega_1=\Theta(\sqrt{\epsilon})$ which together with~\eqref{R-mediumR} yields the following
\begin{IEEEeqnarray}{rCl}
\omega_2\leq \sqrt{1-2^{-2R_2}}+O(\sqrt{\epsilon}).
\end{IEEEeqnarray}
On the other side, plugging~\eqref{P-mediumR} into~\eqref{D-mediumR}, the program in~\eqref{program-mediumR} is upper bounded by the following
\begin{IEEEeqnarray}{rCl}
&&\min_{\omega_2}\;\; 2\sigma^2-2\omega_2\sigma^2+O(\sqrt{\epsilon})\\
&&\text{s.t.}\qquad \omega_2\leq \sqrt{1-2^{-2R_2}}+O(\sqrt{\epsilon}).
\end{IEEEeqnarray}
The solution of the above program is given by
\begin{IEEEeqnarray}{rCl}
\omega_2=\sqrt{1-2^{-2R_2}}+O(\sqrt{\epsilon}).
\end{IEEEeqnarray}
Plugging the above into~\eqref{P-mediumR}, we get
\begin{IEEEeqnarray}{rCl}
\omega_1=\rho\sqrt{2\epsilon\ln 2}(1-\sqrt{1-2^{-2R_2}})+O(\epsilon).
\end{IEEEeqnarray}
Thus, we have
\begin{IEEEeqnarray}{rCl}
\hat{X}_2&=&\rho\sqrt{2\epsilon\ln 2}(1-\sqrt{1-2^{-2R_2}})\hat{X}_1+\sqrt{1-2^{-2R_2}}X_2\nonumber\\&&+Z_2,
\end{IEEEeqnarray}
where $Z_2\sim\mathcal{N}(0,(2^{-2R_2}-\rho^2(1-\sqrt{1-2^{-2R_2}})^2(2\epsilon\ln 2))\sigma^2)$ and the solution of optimization program is as follows
\begin{IEEEeqnarray}{rCl}
D_{2,\text{SA}}^0 &:=& 2\sigma^2 (1-\sqrt{1-2^{-2R_2}})+O(\sqrt{\epsilon}).
\end{IEEEeqnarray}

\underline{\textit{Third Step:}}

For the third step, we have the following optimization program.
\begin{proposition}\label{3rd-step-prop-AR} The optimization program of $0$-PLF-SA for the third step can be written as follows
\begin{subequations}\label{opt-3rd-new-metric}
\begin{IEEEeqnarray}{rCl}
&&\min_{\tau_1,\tau_2,\tau_3} 2\sigma^2-2\tau_3\sigma^2-2\tau_2\omega_2\rho\sigma^2-2\tau_2\omega_1\nu\rho^2\sigma^2\nonumber\\&&\hspace{1cm}-2\tau_1\nu\rho^2\sigma^2\label{distortion_3rd_frame}\\
&&\text{s.t.}:\\
&&\;\;\tau_3^2\sigma^2 (1-2^{-2R_3}(\rho^42^{-2R_1-2R_2}+\rho^2(1-\rho^2)2^{-2R_2}\nonumber\\&&\hspace{1cm}-\rho^2))\leq (1-2^{-2R_3})(1-\tau_1^2-\tau_2^2-2\tau_1\tau_2\omega_1\nu\nonumber\\&&-2\tau_1\tau_2\omega_2\nu\rho-2\tau_2\tau_3\omega_1\nu\rho^2-2\tau_2\tau_3\omega_2\rho-2\tau_1\tau_3\nu\rho^2)\sigma^2,\nonumber\\\label{rate_3rd_frame}\\
&&\hspace{0.1cm}\rho^2\nu=\tau_1+\tau_2\rho\nu+\tau_3\rho^2\nu,\label{perception_3rd_frame}\\
&&\hspace{0.1cm}\omega_1\rho^2\nu+\rho\omega_2=\tau_1\rho\nu+\tau_2+\tau_3(\omega_1\rho^2\nu+\rho\omega_2),\label{perception_3rd_frameb}\\
&&\hspace{0.1cm}\nu=\sqrt{1-2^{-2R_1}}.
\end{IEEEeqnarray}
\end{subequations}
\end{proposition}
\begin{IEEEproof} According to~\eqref{RDP-Gauss-Markov-app}, the optimization program of the third step is given as follows
\begin{IEEEeqnarray}{rCl}
&&\hspace{0cm}\min_{P_{\hat{X}_3|X_3\hat{X}_1\hat{X}_2}} \mathbbm{E}[\|X_3-\hat{X}_3\|^2]\nonumber\\
&&\text{s.t.}\qquad I(X_3;\hat{X}_3|\hat{X}_1,\hat{X}_2)\leq R_3,\nonumber\\
&&\qquad\qquad\;\; P_{\hat{X}_1\hat{X}_2X_3}=P_{\hat{X}_1\hat{X}_2\hat{X}_3}. 
\end{IEEEeqnarray}

Using the above program, we first derive the rate expression in~\eqref{rate_3rd_frame}. Consider the following set of inequalities

\begin{IEEEeqnarray}{rCl}
R_3&\geq &I(X_3;\hat{X}_3|\hat{X}_1,\hat{X}_2)\\&=&h(\hat{X}_3|\hat{X}_1,\hat{X}_2)-h(Z_3)\\
&=& h(\tau_3X_3+Z_3|\hat{X}_1,\hat{X}_2)-h(Z_3)\\
&=& \frac{1}{2}\log 2^{-2h(Z_3)}\left(\tau_3^22^{2h(X_3|\hat{X}_1,\hat{X_2})}+2^{2h(Z_3)}\right)\\
&=& \frac{1}{2}\log 2^{-2h(Z_3)}\left(\tau_3^22^{2h(\rho X_2+N_2|\hat{X}_1,\hat{X_2})}+2^{2h(Z_3)}\right)\nonumber\\\label{R-3rd-just1}\\
&=& \frac{1}{2}\log 2^{-2h(Z_3)} \Bigg(\tau_3^2(\rho^22^{2h( X_2|\hat{X}_1,\hat{X_2})}+2^{2h(N_2)})\nonumber\\&&\hspace{3cm}+2^{2h(Z_3)}\Bigg)\label{R-3rd-just2}\\
&=& \frac{1}{2}\log 2^{-2h(Z_3)}\Bigg(\tau_3^2(\rho^22^{2h( X_2|\hat{X}_1,\hat{X_2})}+(1-\rho^2)\sigma^2)\nonumber\\&&\hspace{3cm}+2^{2h(Z_3)}\Bigg)\\
&\geq & \frac{1}{2}\log 2^{-2h(Z_3)}\Bigg(\tau_3^2(\rho^22^{2h( X_2|\hat{X}_1)}2^{-2R_2}\nonumber\\&&\hspace{3cm}+(1-\rho^2)\sigma^2)+2^{2h(Z_3)}\Bigg)\label{R-3rd-just3}\\
&= & \frac{1}{2}\log 2^{-2h(Z_3)}\Bigg(\tau_3^2(\rho^22^{2h( \rho X_1+N_1|\hat{X}_1)}2^{-2R_2}\nonumber\\&&\hspace{3cm}+(1-\rho^2)\sigma^2)+2^{2h(Z_3)}\Bigg)\label{R-3rd-just4}\\
&= & \frac{1}{2}\log 2^{-2h(Z_3)}\Bigg(\tau_3^2(\rho^42^{-2R_2}2^{2h(X_1|\hat{X}_1)}\nonumber\\&&\hspace{0.5cm}+\rho^2(1-\rho^2)2^{-2R_2}\sigma^2+(1-\rho^2)\sigma^2)+2^{2h(Z_3)}\Bigg)\nonumber\\\label{R-3rd-just5}\\
&\geq & \frac{1}{2}\log 2^{-2h(Z_3)}\Bigg(\tau_3^2(\rho^4\sigma^22^{-2R_1-2R_2}\nonumber\\&&\hspace{0.5cm}+\rho^2(1-\rho^2)2^{-2R_2}\sigma^2+(1-\rho^2)\sigma^2)+2^{2h(Z_3)}\Bigg),\nonumber\\\label{R-final-just}
\end{IEEEeqnarray}
where 
\begin{itemize}
\item~\eqref{R-3rd-just1} follows from~\eqref{Gaus-def2} where $X_3=\rho X_2+N_2$;
\item~\eqref{R-3rd-just2} and~\eqref{R-3rd-just5} follow from Entropy Power Inequality (EPI) (see pp. 22 in \cite{KimElGamal}) which holds which equality for Gaussian sources;
\item~\eqref{R-3rd-just3} follows from the rate constraint $I(X_2;\hat{X}_2|\hat{X}_1)\leq R_2$ which yields $h(X_2|\hat{X}_2,\hat{X}_1)\geq h(X_2|\hat{X}_1)-R_2$;
\item~\eqref{R-final-just} follows from the rate constraint $I(X_1;\hat{X}_1)\leq R_1$ which yields $h(X_1)\geq h(X_1|\hat{X}_1)-R_1$.
\end{itemize}

Thus, re-arranging the terms in~\eqref{R-final-just}, we have
\begin{IEEEeqnarray}{rCl}
&&\hspace{-0.5cm}(\tau_3^2(\rho^2(1-\rho^2)\sigma^22^{-2R_2}+(1-\rho^2)\sigma^2))2^{-2R_3}\nonumber\\&\geq &(1-2^{-2R_3})2^{2h(Z_3)}\\
&=&(1-2^{-2R_3})\cdot\nonumber\\&&\hspace{0.5cm}(1-\tau_1^2-\tau_2^2-\tau_3^2-2\tau_1\tau_2\omega_1-2\tau_1\tau_2\omega_2\rho\nonumber\\&&\hspace{1cm}-2\tau_2\tau_3\omega_1\rho^2-2\tau_2\tau_3\omega_2\rho-2\tau_1\tau_3\rho^2)\sigma^2.\nonumber\\
\end{IEEEeqnarray}
The above constraint can be simplified as follows
\begin{IEEEeqnarray}{rCl}
&&\hspace{-0.5cm}\tau_3^2\sigma^2 (1-\rho^22^{-2R_3}+\rho^2(1-\rho^2)2^{-2R_2}2^{-2R_3})\nonumber\\&\geq& (1-2^{-2R_3})(1-\tau_1^2-\tau_2^2-2\tau_1\tau_2\omega_1-2\tau_1\tau_2\omega_2\rho\nonumber\\&&\hspace{1cm}-2\tau_2\tau_3\omega_1\rho^2-2\tau_2\tau_3\omega_2\rho-2\tau_1\tau_3\rho^2)\sigma^2,\nonumber\\
\end{IEEEeqnarray}
which is the rate expression in~\eqref{rate_3rd_frame}. 

The derivation of the perception constraint in~\eqref{perception_3rd_frame} is given in the following. 
\begin{IEEEeqnarray}{rCl}
\rho^2\nu\sigma^2&=&\mathbbm{E}[X_3\hat{X}_1]\\&=&\mathbbm{E}[\hat{X}_3\hat{X}_1]\label{P-3rd-just1}\\&=&\tau_1\sigma^2+\tau_2\mathbbm{E}[\hat{X}_2\hat{X}_1]+\tau_3\mathbbm{E}[X_3\hat{X}_1]\label{P-3rd-just2}\\
&=&\tau_1\sigma^2+\tau_2\mathbbm{E}[X_2\hat{X}_1]+\tau_3\rho^2\mathbbm{E}[X_1\hat{X}_1]\label{P-3rd-just4}\\
&=&\tau_1\sigma^2+\tau_2\rho\mathbbm{E}[X_1\hat{X}_1]+\tau_3\rho^2\mathbbm{E}[X_1\hat{X}_1]\label{P-3rd-just3}\\
&=&\tau_1\sigma^2+\tau_2\rho\nu\sigma^2+\tau_3\rho^2\nu\sigma^2,
\end{IEEEeqnarray}
where
\begin{itemize}
\item~\eqref{P-3rd-just1} follows from $0$-PLF-SA condition, i.e., $P_{\hat{X}_3\hat{X}_2\hat{X_1}}=P_{X_3\hat{X}_2\hat{X_1}}$ which implies that $\mathbbm{E}[X_3\hat{X}_1]=\mathbbm{E}[\hat{X}_3\hat{X}_1]$ for the Gauss-Markov source model;
\item~\eqref{P-3rd-just2} follows from~\eqref{X3hat-Gaussian-ach} where we multiply both sides with $\hat{X}_1$ and take an expectation over the distribution;
\item~\eqref{P-3rd-just4} follows from the $0$-PLF-SA condition which implies that $\mathbbm{E}[\hat{X}_2\hat{X}_1]=\mathbbm{E}[X_2\hat{X}_1]$ and also from~\eqref{Gaus-def2}, we have $X_3=\rho^2X_1+\rho N_1+N_2$ where $(N_1,N_2)$ are independent of $\hat{X}_1$;
\item~\eqref{P-3rd-just3} follows from~\eqref{Gaus-def2} where $X_2=\rho X_1+N_1$ and $N_1$ is independent of $\hat{X}_1$.
\end{itemize}
Similarly, for derivation of~\eqref{perception_3rd_frameb}, we have 
\begin{IEEEeqnarray}{rCl}
&&\hspace{-1cm}\omega_1\rho^2\nu\sigma^2+\rho\omega_2\sigma^2\nonumber\\&=&\mathbbm{E}[\hat{X}_2X_3]\\&=&\mathbbm{E}[\hat{X}_2\hat{X}_3]\\&=& \tau_1\mathbbm{E}[\hat{X}_1\hat{X}_2]+\tau_2\sigma^2+\tau_3\mathbbm{E}[X_3\hat{X}_2]\\
&=&\tau_1\mathbbm{E}[\hat{X}_1X_2]+\tau_2\sigma^2+\tau_3\mathbbm{E}[X_3\hat{X}_2]\\
&=&\tau_1\rho\nu\sigma^2+\tau_2\sigma^2+\tau_3(\omega_1\rho^2\nu\sigma^2+\rho\omega_2\sigma^2).
\end{IEEEeqnarray}
The distortion term in~\eqref{distortion_3rd_frame} can be derived as follows
\begin{IEEEeqnarray}{rCl}
&&\hspace{-0.5cm}\mathbbm{E}[\|X_3-\hat{X}_3\|^2]\nonumber\\&=& \mathbbm{E}[X_3^2]+\mathbbm{E}[\hat{X}_3^2]-2\mathbbm{E}[X_3\hat{X}_3]\\
&=& 2\sigma^2-2\mathbbm{E}[X_3\hat{X}_3]\label{D-3rd-just1}\\
&=& 2\sigma^2-2(\tau_1\mathbbm{E}[\hat{X}_1X_3]+\tau_2\mathbbm{E}[\hat{X}_2X_3]+\tau_3\sigma^2)\label{D-3rd-just2}\\
&=& 2\sigma^2-2(\tau_1\rho^2\mathbbm{E}[\hat{X}_1X_1]+\tau_2\rho\mathbbm{E}[\hat{X}_2X_2]+\tau_3\sigma^2)\nonumber\\\label{D-3rd-just3}\\
&=& 2\sigma^2-2(\tau_1\rho^2\nu\sigma^2+\tau_2\rho(\rho\nu\omega_1+\omega_2)\sigma^2+\tau_3\sigma^2),\nonumber\\\label{D-3rd-just4}
\end{IEEEeqnarray}
where 
\begin{itemize}
\item~\eqref{D-3rd-just1} follows because $0$-PLF-SA condition implies that $P_{X_3}=P_{\hat{X}_3}$;
\item~\eqref{D-3rd-just2} follows from~\eqref{X3hat-Gaussian-ach} where $X_3=\tau_1\hat{X}_1+\tau_2\hat{X}_2+\tau_3X_3+Z_3$;
\item~\eqref{D-3rd-just3} follows from~\eqref{Gaus-def2};
\item~\eqref{D-3rd-just4} follows from~\eqref{X1hat-Gaussian-ach} and~\eqref{X2hat-Gaussian-ach}.
\end{itemize}
This concludes the proof. 
\end{IEEEproof}
The solution of the optimization program in Proposition~\ref{3rd-step-prop-AR} is plotted in Fig.~\ref{fig:RD3} for some values of the parameters.

\begin{figure}[ht]
\centering
\begin{subfigure}[b]{0.32\textwidth}
  \includegraphics[width=\textwidth]{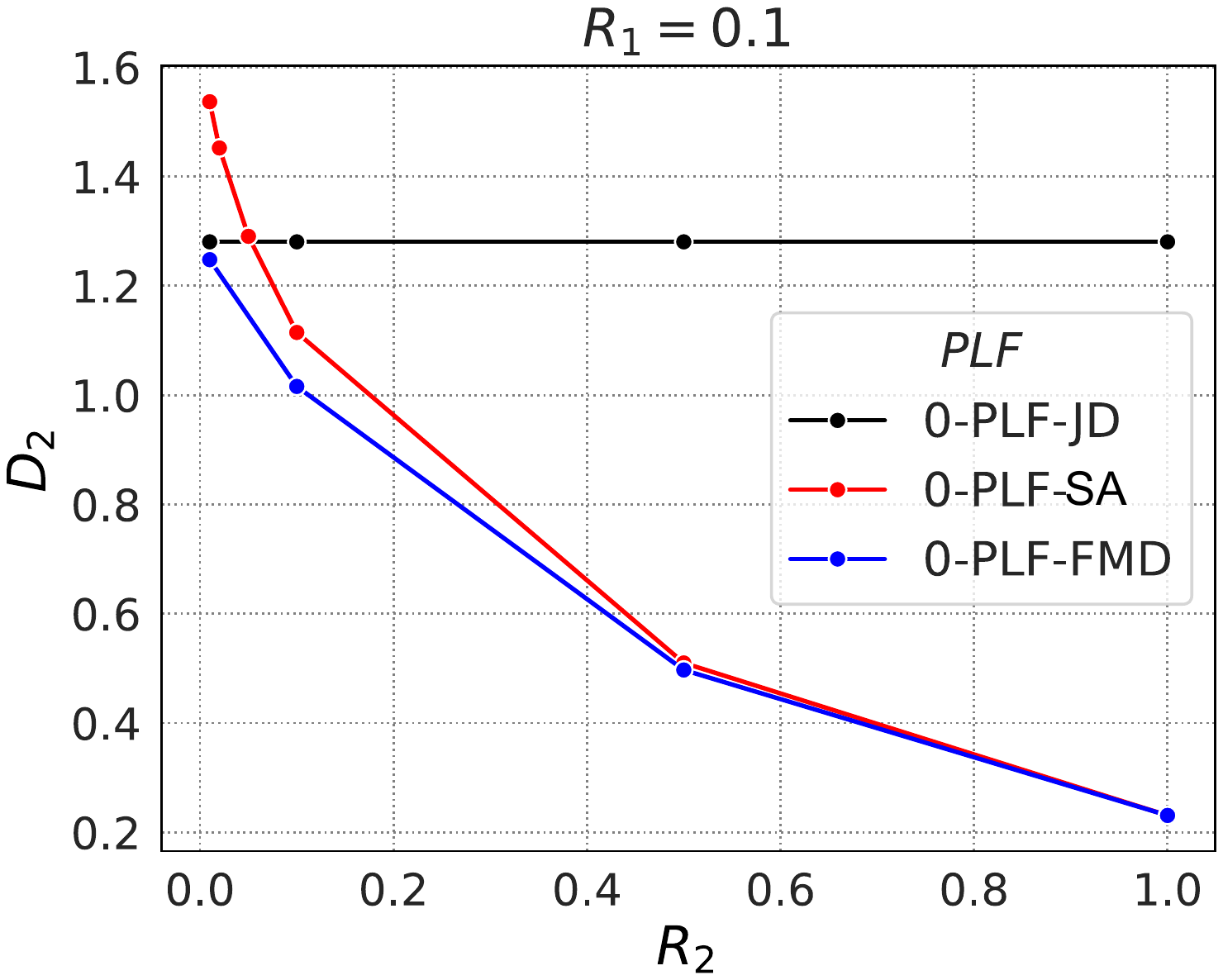}
\end{subfigure}
\centering
\hspace{1cm}\begin{subfigure}[b]{0.32\textwidth}
   \includegraphics[width=\textwidth]{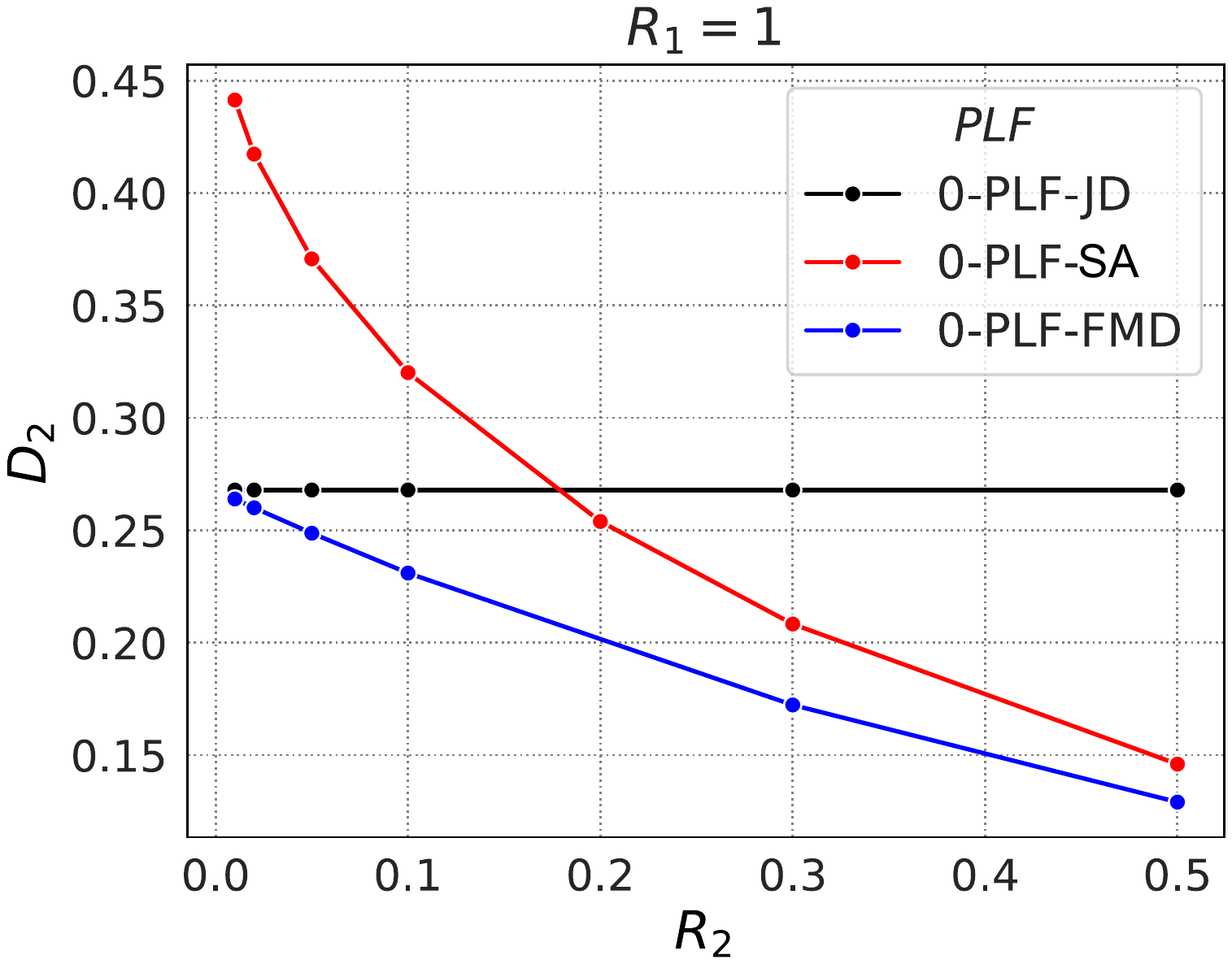}  
\end{subfigure}
\caption{Distortion of the second frame versus its rate for the low-rate regime and $\rho=1$.}
\label{fig:RD2}
\end{figure}

\subsection{$0$-PLF-FMD}\label{FMD-low-rate}

In this section, we propose the optimization program of $0$-PLF-FMD for the second and third steps. We analytically solve the optimization problem of the second step and provide some numerical evaluations for the program of the third step.  

\underline{\textit{Second Step:}}

The optimization program of the second step is similar to that of Proposition~\ref{3rd-step-prop-AR} but with a difference that the condition~\eqref{perception-derivation} which preserves the joint distribution of $(\hat{X}_1,\hat{X}_2)$ is not needed for $0$-PLF-FMD where only marginal distributions are fixed. We also use the following approximation for the rate of the first frame
\begin{IEEEeqnarray}{rCl}
1-2^{-2R_1}=2\epsilon\ln 2 +O(\epsilon^2).
\end{IEEEeqnarray}
Thus, the optimization problem of the second step for $0$-PLF-FMD is as follows
\begin{subequations}\label{FMD-epsilon-program}
\begin{IEEEeqnarray}{rCl}
&&\hspace{0cm}\min_{\substack{\omega_1,\omega_2}}\;\; 2\sigma^2-2 \omega_1\rho\sigma^2\sqrt{2\epsilon\ln 2+O(\epsilon^2)}-2\omega_2\sigma^2,\label{D-FMD}\\
&&\text{s.t.}\qquad \omega_2^2(1-\rho^22^{-2R_2}(2\epsilon\ln 2+O(\epsilon^2)))\leq\nonumber\\&&\hspace{0.5cm} (1-\omega_1^2-2\omega_1\omega_2\rho(2\epsilon\ln 2+O(\epsilon^2)))(1-2^{-2R_2}).\nonumber\\\label{R-FMD}
\end{IEEEeqnarray}
\end{subequations}
Now, we proceed with solving the above optimization program analytically. 
Ignoring the small terms of~\eqref{R-FMD}, this condition reduces to the following 
\begin{IEEEeqnarray}{rCl}
\omega_2^2\leq (1-\omega_1^2)(1-2^{-2R_2}).
\end{IEEEeqnarray} 
Thus, the optimization program of~\eqref{FMD-epsilon-program} with considering the dominant terms reduces to the following 
\begin{subequations}
\begin{IEEEeqnarray}{rCl}
&&\hspace{0cm}\min_{\substack{\omega_1,\omega_2}}\;\; 2\sigma^2-2 \omega_1\rho\sigma^2\sqrt{2\epsilon\ln 2}-2\omega_2\sigma^2,\\
&&\text{s.t.}\qquad \omega_2^2\leq (1-\omega_1^2)(1-2^{-2R_2}).\label{convex-sol}
\end{IEEEeqnarray}
\end{subequations}
The above program is convex and the solution is obtained on the boundary, i.e.,
\begin{IEEEeqnarray}{rCl}
\omega_2^2= (1-\omega_1^2)(1-2^{-2R_2}).
\end{IEEEeqnarray}
Plugging the above  into~\eqref{D-FMD}, we get
\begin{IEEEeqnarray}{rCl}
\min_{\omega_1} 2\sigma^2(1-\rho\omega_1\sqrt{2\epsilon\ln 2}-\sqrt{1-\omega_1^2}\sqrt{1-2^{-2R_2}}).\nonumber\\
\end{IEEEeqnarray}
Taking the derivative of the above expression with respect to $\omega_1$, we have
\begin{IEEEeqnarray}{rCl}
\frac{\omega_1}{\sqrt{1-\omega_1^2}}\sqrt{1-2^{-2R_2}}=\rho\sqrt{2\epsilon\ln 2},
\end{IEEEeqnarray}
which yields
\begin{IEEEeqnarray}{rCl}
\omega_1 = \frac{\rho\sqrt{2\epsilon\ln 2}}{\sqrt{1-2^{-2R_2}+\rho^22\epsilon\ln 2}},
\end{IEEEeqnarray}
and 
\begin{IEEEeqnarray}{rCl}
\omega_2=\frac{1-2^{-2R_2}}{\sqrt{1-2^{-2R_2}+\rho^2 (2\epsilon\ln 2)}}.
\end{IEEEeqnarray}
Thus, we get
\begin{IEEEeqnarray}{rCl}
\hat{X}_2&=&\frac{\rho\sqrt{2\epsilon\ln 2}}{\sqrt{1-2^{-2R_2}+\rho^22\epsilon\ln 2}}\hat{X}_1\nonumber\\&&+\frac{1-2^{-2R_2}}{\sqrt{1-2^{-2R_2}+\rho^2 (2\epsilon\ln 2)}}X_2+Z_{2},
\end{IEEEeqnarray}
where $Z_{2}\sim\mathcal{N}(0,(1-\omega_1^2-\omega_2^2-2\rho\nu\omega_1\omega_2)\sigma^2)$ is a Gaussian random variable independent of $(\hat{X}_1,X_2)$,
and the optimal distortion is given by
\begin{IEEEeqnarray}{rCl}
D_{2,\text{FMD}}^0&:=&  2\sigma^2 (1-\sqrt{1-2^{-2R_2}+\rho^22\epsilon\ln 2})+O(\epsilon).\nonumber\\
\end{IEEEeqnarray}

\underline{\textit{Third Step:}}

The optimization program of the third step for $0$-PLF-FMD is similar to that of~\eqref{opt-3rd-new-metric} with a difference that the conditions~\eqref{perception_3rd_frame} and~\eqref{perception_3rd_frameb} that preserve the joint distributions of $(\hat{X}_1,\hat{X}_2,\hat{X}_3)$ are not needed since for $0$-PLF-FMD, only the marginal distributions are fixed. Thus, we have the following optimization program for the third step
\begin{subequations}
\begin{IEEEeqnarray}{rCl}
&&\min_{\tau_1,\tau_2,\tau_3} 2\sigma^2-2\tau_3\sigma^2-2\tau_2\omega_2\rho\sigma^2-2\tau_2\omega_1\nu\rho^2\sigma^2\nonumber\\&&\hspace{1cm}-2\tau_1\nu\rho^2\sigma^2\\
&&\text{s.t.}:\;\;\tau_3^2\sigma^2 (1-2^{-2R_3}(\rho^42^{-2R_1-2R_2}+\rho^2(1-\rho^2)2^{-2R_2}\nonumber\\&&\hspace{1.7cm}-\rho^2))\leq (1-2^{-2R_3})(1-\tau_1^2-\tau_2^2\nonumber\\&&\hspace{2cm} -2\tau_1\tau_2\omega_1\nu-2\tau_1\tau_2\omega_2\nu\rho-2\tau_2\tau_3\omega_1\nu\rho^2\nonumber\\&&\hspace{2cm}-2\tau_2\tau_3\omega_2\rho-2\tau_1\tau_3\nu\rho^2)\sigma^2.
\end{IEEEeqnarray}
\end{subequations}
The solution of the above optimization program is plotted for some values of parameters in Fig.~\ref{fig:RD3}.

\subsection{$0$-PLF-JD}\label{JD-low-rate} 

In this section, we propose the optimization programs of $0$-PLF-JD for the second and third steps. We analytically solve the optimization problem of the second frame and provide some numerical evaluations for the third step. 

\underline{\textit{Second Step:}}

The optimization program of the second step is similar to that of Proposition~\ref{0-PLF-AR-2nd-step} with a difference that the condition in~\eqref{perception-derivation} is replaced by the corresponding condition of $0$-PLF-JD which is $P_{X_1X_2}=P_{\hat{X}_1\hat{X}_2}$. This constraint implies that $\mathbbm{E}[X_1X_2]=\mathbbm{E}[\hat{X}_1\hat{X}_2]$ which together with~\eqref{X1hat-Gaussian-ach} and~\eqref{X2hat-Gaussian-ach} yields 
\begin{IEEEeqnarray}{rCl}
\omega_1+\nu\omega_2\rho=\rho.
\end{IEEEeqnarray}
Thus, the optimization problem of the second step for $0$-PLF-JD when $R_1=\epsilon$ is as follows
\begin{subequations}
\begin{IEEEeqnarray}{rCl}
&&\hspace{0cm}\min_{\substack{\omega_1,\omega_2}}\;\; 2\sigma^2-2 \omega_1\rho\sigma^2\sqrt{2\epsilon\ln 2+O(\epsilon^2)}-2\omega_2\sigma^2,\label{objective-JD}\\
&&\text{s.t.}\qquad \omega_2^2(1-\rho^22^{-2R_2}(2\epsilon\ln 2+O(\epsilon^2)))\leq\nonumber\\&&\hspace{0.5cm} (1-\omega_1^2-2\omega_1\omega_2\rho\sqrt{2\epsilon\ln 2+O(\epsilon^2)})(1-2^{-2R_2}),\nonumber\\\label{R-JD}\\
&&\qquad\;\;\;\; \omega_1+\nu\omega_2\rho=\rho.\label{P-JD}
\end{IEEEeqnarray}
\end{subequations}
The constraint~\eqref{P-JD} implies that 
\begin{IEEEeqnarray}{rCl}
\omega_1=\rho-\rho\omega_2\sqrt{2\epsilon\ln 2}+O(\epsilon).
\end{IEEEeqnarray}
Plugging the above into~\eqref{objective-JD} and~\eqref{R-JD}, we get 
\begin{subequations}
\begin{IEEEeqnarray}{rCl}
&&\hspace{0cm}\min_{\omega_2}\;\; 2\sigma^2(1-\rho^2\sqrt{2\epsilon\ln 2}-\omega_2)+O(\epsilon) \\
&&\text{s.t.}: \omega_2\leq \sqrt{1-\rho^2}\sqrt{1-2^{-2R_2}}+O(\sqrt{\epsilon}).
\end{IEEEeqnarray}
\end{subequations}
The solution of the above program is given by
\begin{IEEEeqnarray}{rCl}
\omega_2= \sqrt{1-\rho^2}\sqrt{1-2^{-2R_2}}+O(\sqrt{\epsilon}).
\end{IEEEeqnarray}
Thus, we have
\begin{IEEEeqnarray}{rCl}
\hat{X}_2&=&(\rho-\rho\omega_2\sqrt{2\epsilon\ln 2})\hat{X}_1+\sqrt{1-\rho^2}\sqrt{1-2^{-2R_2}}X_2\nonumber\\&&+Z_2,
\end{IEEEeqnarray}
where $Z_2\sim\mathcal{N}(0,((1-\rho^2)2^{-2R_2}-\rho^2\sqrt{1-\rho^2}\sqrt{1-2^{-2R_2}}\sqrt{2\epsilon\ln 2})\sigma^2)$ is a Gaussian random variable independent of $(\hat{X}_1,X_2)$ and the optimal distortion is given by
\begin{IEEEeqnarray}{rCl}
D_{2,\text{JD}}^0 &:=& 2\sigma^2 (1-\sqrt{1-\rho^2}\sqrt{1-2^{-2R_2}}-\rho^2\sqrt{2\epsilon \ln 2})\nonumber\\&&\hspace{0.5cm}+O(\epsilon).
\end{IEEEeqnarray}

\underline{\textit{Third Step:}}

The optimization program of the third step for $0$-PLF-JD is similar to~\eqref{opt-3rd-new-metric} but with a difference that the conditions in~\eqref{perception_3rd_frame} and~\eqref{perception_3rd_frameb} are replaced by the corresponding conditions of $0$-PLF-JD which is $P_{X_1X_2X_3}=P_{\hat{X}_1\hat{X}_2\hat{X}_3}$. This constraint implies that 
\begin{IEEEeqnarray}{rCl}
\mathbbm{E}[X_1X_3]&=&\mathbbm{E}[\hat{X}_1\hat{X}_3],\\
\mathbbm{E}[X_2X_3]&=&\mathbbm{E}[\hat{X}_2\hat{X}_3].
\end{IEEEeqnarray}
Considering~\eqref{X1hat-Gaussian-ach}--\eqref{X3hat-Gaussian-ach} together with the above conditions, we get
\begin{IEEEeqnarray}{rCl}
\rho^2&=&\tau_1+\tau_2\rho+\tau_3\rho^2\nu,\\
\rho&=&\tau_1\rho+\tau_2+\tau_3(\omega_1\rho^2\nu+\rho\omega_2).
\end{IEEEeqnarray}
Thus, we have the following optimization program for the third step
\begin{subequations}
\begin{IEEEeqnarray}{rCl}
&&\min_{\tau_1,\tau_2,\tau_3} 2\sigma^2-2\tau_3\sigma^2-2\tau_2\omega_2\rho\sigma^2-2\tau_2\omega_1\nu\rho^2\sigma^2\nonumber\\&&\hspace{1cm}-2\tau_1\nu\rho^2\sigma^2\\
&&\text{s.t.}:\;\;\tau_3^2\sigma^2 (1-2^{-2R_3}(\rho^42^{-2R_1-2R_2}+\rho^2(1-\rho^2)2^{-2R_2}\nonumber\\&&\hspace{1.5cm}-\rho^2))\leq (1-2^{-2R_3})(1-\tau_1^2-\tau_2^2-2\tau_1\tau_2\omega_1\nu\nonumber\\&&\hspace{1.5cm} -2\tau_1\tau_2\omega_2\nu\rho-2\tau_2\tau_3\omega_1\nu\rho^2-2\tau_2\tau_3\omega_2\rho\nonumber\\&&\hspace{1.5cm}-2\tau_1\tau_3\nu\rho^2)\sigma^2,\\
&&\hspace{1cm}\rho^2=\tau_1+\tau_2\rho+\tau_3\rho^2\nu,\\
&&\hspace{1cm}\rho=\tau_1\rho+\tau_2+\tau_3(\omega_1\rho^2\nu+\rho\omega_2).
\end{IEEEeqnarray}
\end{subequations}
The solution of the above program is plotted in Fig.~\ref{fig:RD3} for some values of parameters. For the case $R_1=R_2=0.1$ (low compression rates) and a large range of rates $R_3$, the performances of $0$-PLF-SA and $0$-PLF-FMD are almost the same. For $R_1=R_2=1$ (low compression rates), the distortion of $0$-PLF-SA is significantly smaller than that of $0$-PLF-JD for all values of $R_3$, and for a large enough $R_3$, it performs similar to $0$-PLF-FMD.

\section{High-Rate Regime for the First Frame}\label{high-rate-app}

In this section, we first prove the  following theorem where the first frame is compressed at a high rate, i.e., $R_1\to \infty$. The rates of all subsequent frames are assumed to be small, i.e., $R_j=\epsilon$ for sufficiently small $\epsilon>0$ and $j\in\{2,\ldots,T\}$. Then, we provide proofs for the achievable reconstructions of $0$-PLF-FMD as outlined in Table~\ref{table-ach-recons}.

\begin{theorem} Let $R_1\to \infty$ and $R_j=\epsilon$ for sufficiently small $\epsilon>0$ and $j\in\{2,\ldots,T\}$. An achievable reconstruction of $0$-PLF-SA in $j$th frame ($j\in\{1,\ldots,T\}$) is given by
\begin{IEEEeqnarray}{rCl}
\hat{X}_j&=&\rho^{j-1}\hat{X}_1+\sum_{i=1}^{j-1}O(\sqrt{\epsilon})N_i+\sum_{i=2}^{j-2}O(\sqrt{\epsilon})Z_{i,\text{SA}}\nonumber\\&&\hspace{0.5cm}+O(\sqrt{\epsilon})Z_{j-1,\text{SA}}+Z_{j,\text{SA}},
\end{IEEEeqnarray}
where $Z_{j,\text{SA}}$ is a Gaussian random noise independent of $(\{N_i\}_{i=1}^{j-1},\{Z_{i,\text{SA}}\}_{i=2}^{j-1})$, with mean zero and variance $(1-\rho^{2(j-1)}+O(\epsilon))\sigma^2$, and the distortion is as follows
\begin{IEEEeqnarray}{rCl}
D_{j,\text{SA}}^{\infty}=2(1-\rho^{2(j-1)}-O(\sqrt{\epsilon}))\sigma^2+O(\epsilon),
\end{IEEEeqnarray}
and an achievable reconstruction of $0$-PLF-JD in $j$th frame is given by
\begin{IEEEeqnarray}{rCl}
\hat{X}_j&=& \rho^{j-1}\hat{X}_1+\sum_{i=1}^{j-1}O(\sqrt{\epsilon})N_i+\sum_{i=2}^{j-2}O(\sqrt{\epsilon})Z_{i,\text{JD}}\nonumber\\&&\hspace{0.5cm}+\rho Z_{j-1,\text{JD}}+Z_{j,\text{JD}},
\end{IEEEeqnarray}
where $Z_{j,\text{JD}}$ is a Gaussian random noise independent of $(\{N_i\}_{i=1}^{j-1},\{Z_{i,\text{JD}}\}_{i=2}^{j-1})$ with mean zero and variance given in Section~\ref{JD-high-R1-low-R23},
and the distortion is as follows
\begin{IEEEeqnarray}{rCl}
D_{j,\text{JD}}^{\infty}=2\left(1-\rho^{2(j-1)}-O(\sqrt{\epsilon})\right)\sigma^2+O(\epsilon).
\end{IEEEeqnarray}
\end{theorem}
To prove the above theorem, we consider each PLF separately. We provide the analysis for the second, third and fourth frames. We then use an induction to derive the achievable reconstruction for $j$th frame. Notice that the solutions for the second and third frames are also presented in Table~\ref{table-ach-recons}.

\subsection{$0$-PLF-SA}\label{CP-high-R1-low-R23}

In this section, we introduce the optimization programs of the second, third and fourth steps for $0$-PLF-SA and provide the solutions for them. The results are further extended to $T$ frames. Similar to~\eqref{X2hat-Gaussian-ach}--\eqref{X3hat-Gaussian-ach}, we write the achievable reconstructions of the second and third steps as follows 
\begin{IEEEeqnarray}{rCl}
\hat{X}_2&=&\omega_1\hat{X}_1+\omega_2X_2+Z_{2,\text{SA}},\label{X2hat-Gaussian-AR}\\
\hat{X}_3&=&\tau_1\hat{X}_1+\tau_2\hat{X}_2+\tau_3X_3+Z_{3,\text{SA}},\label{X3hat-Gaussian-AR}
\end{IEEEeqnarray}
where $Z_{2,\text{SA}}$ and $Z_{3,\text{SA}}$ are Gaussian random variables independent of $(\hat{X}_1,X_2)$ and $(\hat{X}_1,\hat{X}_2,X_3)$, respectively.

\underline{\textit{Second Step:}}

The optimization program of the second step for $0$-PLF-SA is similar to that of Proposition~\ref{0-PLF-AR-2nd-step} but with a difference that $\nu=1$ since we have a high compression rate for the first frame. Thus, the optimization program of the second step is as follows
\begin{subequations}\label{CP-2nd-opt}
\begin{IEEEeqnarray}{rCl}
&&\hspace{0cm}\min_{\substack{\omega_1,\omega_2}}\;\; 2\sigma^2-2 \omega_1\rho\sigma^2-2\omega_2\sigma^2,\\
&&\text{s.t.}\;\; \omega_2^2(1-\rho^22^{-2R_2})\leq (1-\omega_1^2-2\omega_1\omega_2\rho)(1-2^{-2R_2}),\nonumber\\\label{D-cons-high-rate}\\
&&\qquad \omega_1+\omega_2\rho=\rho.\label{P-cons-high-rate}
\end{IEEEeqnarray}
\end{subequations}
For the second frame, the achievable reconstruction is given as follows (see Table 2 in \cite{Jun-Ashish2023})
\begin{IEEEeqnarray}{rCl}
\hat{X}_2=(\rho-\rho\sqrt{2\epsilon\ln 2})\hat{X}_1+\sqrt{2\epsilon\ln 2}X_2+Z_{2,\text{SA}},\nonumber\\\label{2nd-step-reconstruction}
\end{IEEEeqnarray}
where $Z_{2,\text{SA}}\sim\mathcal{N}(0,(1-\rho^2+O(\epsilon))\sigma^2)$ is independent of $(\hat{X}_1,X_2)$ and $\hat{X}_1=X_1$ and the distortion is given as follows
\begin{IEEEeqnarray}{rCl}
D_{2,\text{SA}}^{\infty} := 2(1-\rho^2-(1-\rho^2)\sqrt{2\epsilon\ln 2})\sigma^2.
\end{IEEEeqnarray}

\underline{\textit{Third Step:}}

The optimization program of the third step is similar to that of Proposition~\ref{3rd-step-prop-AR} but when $\nu=1$. Thus, we have the following program
\begin{subequations}\label{CP-3rd-opt}
\begin{IEEEeqnarray}{rCl}
&&\min_{\tau_1,\tau_2,\tau_3} 2\sigma^2-2\tau_3\sigma^2-2\tau_2\omega_2\rho\sigma^2-2\tau_2\omega_1\rho^2\sigma^2-2\tau_1\rho^2\sigma^2\nonumber\\\\
&&\text{s.t.}:\;\;\tau_3^2 (1-2^{-2R_3}(\rho^42^{-2R_1-2R_2}+\rho^2(1-\rho^2)2^{-2R_2}\nonumber\\&&\hspace{1.5cm}-\rho^2))\leq (1-2^{-2R_3})(1-\tau_1^2-\tau_2^2-2\tau_1\tau_2\omega_1\nonumber\\&&\hspace{2.5cm} -2\tau_1\tau_2\omega_2\rho-2\tau_2\tau_3\omega_1\rho^2-2\tau_2\tau_3\omega_2\rho\nonumber\\&&\hspace{2.5cm}-2\tau_1\tau_3\rho^2),\label{rate-R1inf}\\
&&\hspace{1cm}\rho^2=\tau_1+\tau_2\rho+\tau_3\rho^2,\label{perception1-R1inf}\\
&&\hspace{1cm}\omega_1\rho^2+\rho\omega_2=\tau_1\rho+\tau_2+\tau_3(\omega_1\rho^2+\rho\omega_2).\label{perception2-R1inf}
\end{IEEEeqnarray}
\end{subequations}

\textbf{Case of $R_3\to \infty$}: In this case, the solution of the optimization problem is trivially given by $\hat{X}_3=X_3$ since it satisfies the $0$-PLF-SA condition in the third frame which is $P_{\hat{X}_3\hat{X}_2\hat{X}_1}=P_{X_3\hat{X}_2\hat{X}_1}$.

\textbf{Case of $R_3=R_2=\epsilon$}: We will simplify the program~\eqref{CP-3rd-opt} and derive the solution. We consider the following approximation
\begin{IEEEeqnarray}{rCl}
1-2^{-2R_j}=2\epsilon\ln 2 +O(\epsilon^2),\qquad j\in\{2,3\}.
\end{IEEEeqnarray}
Considering the dominant terms of~\eqref{rate-R1inf}, this constraint can be written as follows
\begin{IEEEeqnarray}{rCl}
(1-\rho^4)\tau_3^2\leq (1-\tau_1^2-\tau_2^2)(2\epsilon\ln 2).\label{R3-R1inf}
\end{IEEEeqnarray}
So, the optimization program in~\eqref{CP-3rd-opt} simplifies as follows
\begin{subequations}
\begin{IEEEeqnarray}{rCl}
&&\min_{\tau_1,\tau_2,\tau_3} 2\sigma^2-2\tau_3\sigma^2-2\tau_2\omega_2\rho\sigma^2-2\tau_2\omega_1\rho^2\sigma^2-2\tau_1\rho^2\sigma^2\nonumber\\\\
&&\text{s.t.}:\;\;(1-\rho^4)\tau_3^2\leq (1-\tau_1^2-\tau_2^2)(2\epsilon\ln 2),\\
&&\hspace{1cm}\rho^2=\tau_1+\tau_2\rho+\tau_3\rho^2,\\
&&\hspace{1cm}\omega_1\rho^2+\rho\omega_2=\tau_1\rho+\tau_2+\tau_3(\omega_1\rho^2+\rho\omega_2).
\end{IEEEeqnarray}
\end{subequations}
We write $\tau_1$, $\tau_2$ and $\tau_3$ as $\tau_1=K_1+\delta_1\sqrt{2\epsilon\ln 2}$, $\tau_2=K_2+\delta_2\sqrt{2\epsilon\ln 2}$ and $\tau_3=\delta_3\sqrt{2\epsilon\ln 2}$, and plug them into~\eqref{perception1-R1inf}--\eqref{perception2-R1inf} to get the following equations
\begin{subequations}
\begin{IEEEeqnarray}{rCl}
\rho^2&=& K_1+\rho K_2,\label{equation1}\\
\rho^3 &=& K_1\rho + K_2,\label{equation2}\\
0&=& \delta_1+\rho\delta_2+\rho^2\delta_3,\label{equation3}\\
-\rho^3+\rho&=&\rho\delta_1+\delta_2+\rho^3\delta_3.\label{equation4}
\end{IEEEeqnarray}
\end{subequations}
Notice that~\eqref{equation1}--\eqref{equation2} yields $K_2=0$ and $K_1=\rho^2$. The constant terms of $\tau_1$ and $\tau_2$ which are $K_1$ and $K_2$, contribute to the dominant terms of~\eqref{R3-R1inf}. Plugging the values of $K_1$ and $K_2$ into~\eqref{R3-R1inf}, we have the following inequality
\begin{IEEEeqnarray}{rCl}
\delta_3 \leq 1.
\end{IEEEeqnarray}
So, considering the dominant terms, the optimization program in~\eqref{CP-3rd-opt} is upper bounded by the following 
\begin{subequations}
\begin{IEEEeqnarray}{rCl}
&&\min_{\delta_1,\delta_2,\delta_3} 2\sigma^2(1-\rho^4-(\rho^2\delta_1+\rho^3\delta_2+\delta_3)\sqrt{2\epsilon\ln 2})\nonumber\\\\
&&\text{s.t.}:\;\;\;\delta_3\leq 1,\\
&&\hspace{1cm}\delta_1+\rho\delta_2+\rho^2\delta_3=0,\\
&&\hspace{1cm}\rho\delta_1+\delta_2+\rho^3\delta_3=-\rho^3+\rho.
\end{IEEEeqnarray}
\end{subequations}
The above optimization program is convex, so the solution is obtained at the boundary of the feasible region where we get
\begin{IEEEeqnarray}{rCl}
\delta_1&=& -2\rho^2,\\
\delta_2&=& \rho,\\
\delta_3&=&1.
\end{IEEEeqnarray}
Thus, we get the following achievable reconstruction
\begin{IEEEeqnarray}{rCl}
\hat{X}_3&=& (\rho^2-2\rho^2\sqrt{2\epsilon\ln 2})\hat{X}_1+\rho\sqrt{2\epsilon\ln 2}\hat{X}_2\nonumber\\&&\hspace{0.5cm}+\sqrt{2\epsilon\ln 2}X_3+Z_{3,\text{SA}},\label{CP-3rd-step-reconstruction}
\end{IEEEeqnarray}
where $Z_{3,\text{SA}}\sim\mathcal{N}(0,(1-\rho^4+O(\epsilon))\sigma^2)$ and the distortion is given by
\begin{IEEEeqnarray}{rCl}
D^{\infty}_{3,\text{SA}}:=2(1-\rho^4-(1-\rho^4)\sqrt{2\epsilon\ln 2})\sigma^2.
\end{IEEEeqnarray}
Plugging~\eqref{2nd-step-reconstruction} into~\eqref{CP-3rd-step-reconstruction} yields the following
\begin{IEEEeqnarray}{rCl}
\hat{X}_3 &=& (\rho^2-\rho^2\sqrt{2\epsilon\ln 2})\hat{X}_1+\sqrt{2\epsilon\ln 2}X_3\nonumber\\&&\hspace{0.5cm}+\rho\sqrt{2\epsilon\ln 2}Z_{2,\text{SA}}+Z_{3,\text{SA}},\label{CP-3rd-recons}
\end{IEEEeqnarray}

Using~\eqref{Gaus-def2}, the expression in~\eqref{CP-3rd-recons} can be written as the following
\begin{IEEEeqnarray}{rCl}
\hat{X}_3&=&\rho^2\hat{X}_1+\rho\sqrt{2\epsilon\ln 2}N_1+\sqrt{2\epsilon\ln 2}N_2\nonumber\\&&\hspace{0.5cm}+\rho\sqrt{2\epsilon\ln 2}Z_{2,\text{SA}}+Z_{3,\text{SA}}.
\end{IEEEeqnarray}

\underline{\textit{Fourth Step:}}
We derive the optimization program of the fourth frame and solve it. For the fourth frame, we write the achievable reconstruction as follows
\begin{IEEEeqnarray}{rCl}
\hat{X}_4=\lambda_1\hat{X}_1+\lambda_2\hat{X}_2+\lambda_3\hat{X}_3+\lambda_4X_4+Z_{4,\text{SA}},\label{X4hat-Gaussian-ach}
\end{IEEEeqnarray}
where $Z_{4,\text{SA}}$ is a Gaussian random variable independent of $(\hat{X}_1,\hat{X}_2,\hat{X}_3,X_4)$ with mean zero and its variance will be determined later.

\begin{proposition}\label{4th-step-AR-propostion} The optimization program of the fourth step for $0$-PLF-SA when the first frame has a high compression rate, is given as follows
\begin{subequations}\label{4th-step-program-AR}
\begin{IEEEeqnarray}{rCl}
&&\min_{\lambda_1,\lambda_2,\lambda_3,\lambda_4}2\sigma^2-2\lambda_4\sigma^2-2\lambda_3\rho\tau_3\sigma^2-2\lambda_3\rho^2\tau_2\omega_2\sigma^2\nonumber\\&&\hspace{1.5cm}-2\lambda_3\rho^3\tau_2\omega_1\sigma^2-2\lambda_3\rho^3\tau_1\sigma^2\nonumber\\[-0.5ex]&&\hspace{1.5cm}-2\lambda_2\rho^3\omega_1\sigma^2-2\lambda_2\rho^2\omega_2\sigma^2-2\lambda_1\rho^3\sigma^2\label{objective-4th}\\[2ex]
&&\text{s.t.}: 2^{-2R_4}(\lambda_4^2\rho^62^{-2R_3-2R_2-2R_1}\sigma^2\nonumber\\&&\hspace{0.5cm}+\lambda_4^2\rho^42^{-2R_3-2R_2}(1-\rho^2)\sigma^2+\lambda_4^2\rho^22^{-2R_3}(1-\rho^2)\sigma^2\nonumber\\&&\hspace{0.5cm}+\lambda_4^2(1-\rho^2)\sigma^2)\leq 2^{2h(Z_{4,\text{SA}})}(1-2^{-2R_4}),\label{rate-4th-step}\\[1ex]
&&\hspace{0.6cm} \rho^3 = \lambda_1+\rho \lambda_2+\rho^2 \lambda_3+\rho^3\lambda_4,\label{P1-4tha}\\
&&\hspace{0.6cm} \rho^2(\rho\omega_1+\omega_2)= \rho\lambda_1+\lambda_2+\rho(\rho\omega_1+\omega_2)\lambda_3\nonumber\\&&\hspace{3cm}+\rho^2(\rho\omega_1+\omega_2)\lambda_4,\label{P2-4tha}\\
&&\hspace{0.6cm}\rho (\rho^2\tau_1+\rho(\rho\omega_1+\omega_2)\tau_2+\tau_3)=\nonumber\\&&\hspace{2cm}\rho^2\lambda_1+\rho(\rho\omega_1+\omega_2)\lambda_2+\lambda_3+\rho(\rho^2\tau_1\nonumber\\&&\hspace{2cm}+\rho(\rho\omega_1+\omega_2)\tau_2+\tau_3)\lambda_4.\label{P3-4tha}
\end{IEEEeqnarray}
\end{subequations}
\end{proposition}
\begin{IEEEproof} An extension of~\eqref{RDP-Gauss-Markov-app} to the fourth step yields the following optimization program
\begin{IEEEeqnarray}{rCl}
&&\hspace{0cm}\min_{P_{\hat{X}_4|X_4\hat{X}_1\hat{X}_2\hat{X}_3}} \mathbbm{E}[\|X_4-\hat{X}_4\|^2]\nonumber\\
&&\text{s.t.}\qquad I(X_4;\hat{X}_4|\hat{X}_1,\hat{X}_2,\hat{X}_3)\leq R_4,\nonumber\\
&&\qquad\qquad\;\; P_{\hat{X}_1\hat{X}_2\hat{X}_3X_4}=P_{\hat{X}_1\hat{X}_2\hat{X}_3\hat{X}_4}. 
\end{IEEEeqnarray}

The perception constraints in~\eqref{P1-4tha}--\eqref{P3-4tha} are derived based on $0$-PLF-SA condition which is $P_{\hat{X}_4\hat{X}_3\hat{X}_2\hat{X}_1}=P_{X_4\hat{X}_3\hat{X}_2\hat{X}_1}$. This implies that $\mathbbm{E}[\hat{X}_4\hat{X}_1]=\mathbbm{E}[X_4\hat{X}_1]$, $\mathbbm{E}[\hat{X}_4\hat{X}_2]=\mathbbm{E}[X_4\hat{X}_2]$ and $\mathbbm{E}[\hat{X}_4\hat{X}_3]=\mathbbm{E}[X_4\hat{X}_3]$. These constraints combined with~\eqref{X1hat-Gaussian-ach}--\eqref{X3hat-Gaussian-ach},~\eqref{X4hat-Gaussian-ach} yield ~\eqref{P1-4tha}--\eqref{P3-4tha}. For the rate constraint, consider the following set of inequalities 
\begin{IEEEeqnarray}{rCl}
&& \hspace{-0.5cm}I(X_4;\hat{X}_4|\hat{X}_1,\hat{X}_2,\hat{X}_3)\\&=&h(\hat{X}_4|\hat{X}_1,\hat{X}_2,\hat{X}_3)-h(Z_{4,\text{SA}})\\
&=& h(\lambda_4X_4+Z_{4,\text{SA}}|\hat{X}_1,\hat{X}_2,\hat{X}_3)-h(Z_{4,\text{SA}})\\
&=& \frac{1}{2}\log 2^{-2h(Z_{4,\text{SA}})}\left(\lambda_4^22^{2h(X_4|\hat{X}_1,\hat{X}_2,\hat{X}_3)}+2^{2h(Z_{4,\text{SA}})}\right)\nonumber\\\label{4th-frame-step1}\\
&=&  \frac{1}{2}\log 2^{-2h(Z_{4,\text{SA}})}\Bigg(\lambda_4^2\rho^22^{2h(X_3|\hat{X}_1,\hat{X}_2,\hat{X}_3)}+\lambda_4^22^{2h(N_3)}\nonumber\\&&\hspace{3cm}+2^{2h(Z_{4,\text{SA}})}\Bigg)\\
&\geq & \frac{1}{2}\log 2^{-2h(Z_{4,\text{SA}})}\Bigg(\lambda_4^2\rho^22^{-2R_3}2^{2h(X_3|\hat{X}_1,\hat{X}_2)}\nonumber\\&&\hspace{3cm}+\lambda_4^22^{2h(N_3)}+2^{2h(Z_{4,\text{SA}})}\Bigg)\label{4th-frame-step2}\\
&= & \frac{1}{2}\log 2^{-2h(Z_{4,\text{SA}})}\Bigg(\lambda_4^2\rho^42^{-2R_3}2^{2h(X_2|\hat{X}_1,\hat{X}_2)}\nonumber\\&&\hspace{3cm}+\lambda_4^2\rho^22^{-2R_3}2^{2h(N_2)}+\lambda_4^22^{2h(N_3)}\nonumber\\&&\hspace{3cm}+2^{2h(Z_{4,\text{SA}})}\Bigg)\label{4th-frame-step3}\\
&\geq & \frac{1}{2}\log 2^{-2h(Z_{4,\text{SA}})}\Bigg(\lambda_4^2\rho^42^{-2R_3-2R_2}2^{2h(X_2|\hat{X}_1)}\nonumber\\&&\hspace{3cm}+\lambda_4^2\rho^22^{-2R_3}2^{2h(N_2)}+\lambda_4^22^{2h(N_3)}\nonumber\\&&\hspace{3cm}+2^{2h(Z_{4,\text{SA}})}\Bigg)\label{4th-frame-step4}\\
&=&\frac{1}{2} \log 2^{-2h(Z_{4,\text{SA}})}\Big(\lambda_4^2\rho^62^{-2R_3-2R_2}2^{2h(X_1|\hat{X}_1)}\nonumber\\&&\hspace{3cm}+\lambda_4^2\rho^42^{-2R_3-2R_2}2^{2h(N_1)}\nonumber\\&&\hspace{3cm}+\lambda_4^2\rho^22^{-2R_3}2^{2h(N_2)}\nonumber\\[-1ex]&&\hspace{3cm}+\lambda_4^22^{2h(N_3)}+2^{2h(Z_{4,\text{SA}})}\Big)\label{4th-frame-step5}\\
&\geq &\frac{1}{2}\log 2^{-2h(Z_{4,\text{SA}})}\Big(\lambda_4^2\rho^62^{-2R_3-2R_2-2R_1}\sigma^2\nonumber\\&&\hspace{3cm}+\lambda_4^2\rho^42^{-2R_3-2R_2}2^{2h(N_1)}\nonumber\\&&\hspace{3cm}+\lambda_4^2\rho^22^{-2R_3}2^{2h(N_2)}\nonumber\\[-1ex]&&\hspace{3cm}+\lambda_4^22^{2h(N_3)}+2^{2h(Z_{4,\text{SA}})}\Big),\label{4th-frame-step6}
\end{IEEEeqnarray}

where 
\begin{itemize}
\item~\eqref{4th-frame-step1} follows from EPI (see pp. 22 in \cite{KimElGamal}) which holds with equality for Gaussian sources;
\item~\eqref{4th-frame-step2},~\eqref{4th-frame-step4} and ~\eqref{4th-frame-step6} follows from the rate constraints $R_3\geq I(X_3;\hat{X}_3|\hat{X}_1,\hat{X}_2)$, $R_2\geq I(X_2;\hat{X}_2|\hat{X}_1)$ and $R_1\geq I(X_1;\hat{X}_1)$, respectively;
\item~\eqref{4th-frame-step3} and~\eqref{4th-frame-step5}  follow from~\eqref{Gaus-def2} where $X_3=\rho X_2+N_2$ and $X_2=\rho X_1+N_1$, respectively, and the fact that EPI holds with equality for Gaussian sources.
\end{itemize}
Re-arranging the terms in~\eqref{4th-frame-step6}, we get to the constraint in~\eqref{rate-4th-step}. The objective function in~\eqref{objective-4th} is obtained by the expansion of $\mathbbm{E}[\|X_4-\hat{X}_4\|^2]$ using~\eqref{X2hat-Gaussian-AR},~\eqref{X3hat-Gaussian-AR} and~\eqref{X4hat-Gaussian-ach}.
\end{IEEEproof}
Now, we provide the solution of the optimization program in~\eqref{4th-step-program-AR} when $R_2=R_3=R_4=\epsilon$ for sufficiently small $\epsilon>0$. Using the following approximation
\begin{IEEEeqnarray}{rCl}
1-2^{-2R_j}=2\epsilon\ln 2+O(\epsilon^2),
\end{IEEEeqnarray}
and considering the dominant terms of~\eqref{rate-4th-step}, the solution of the optimization program is upper bounded by

\begin{subequations}\label{opt-AR-4th-program}
\begin{IEEEeqnarray}{rCl}
&&\min_{\lambda_1,\lambda_2,\lambda_3,\lambda_4}2\sigma^2-2\lambda_4\sigma^2-2\lambda_3\rho\tau_3\sigma^2-2\lambda_3\rho^2\tau_2\omega_2\sigma^2\nonumber\\[-1ex]&&\hspace{1.5cm}-2\lambda_3\rho^3\tau_2\omega_1\sigma^2-2\lambda_3\rho^3\tau_1\sigma^2-2\lambda_2\rho^3\omega_1\sigma^2\nonumber\\&&\hspace{1.5cm}-2\lambda_2\rho^2\omega_2\sigma^2-2\lambda_1\rho^3\sigma^2\\[1.5ex]
&&\text{s.t.}: \lambda_4^2(1-\rho^6)\leq (1-\lambda_1^2-\lambda_2^2-\lambda_3^2)(2\epsilon\ln 2),\label{rate-4th-step}\\
&&\hspace{0.7cm} \rho^3 = \lambda_1+\rho \lambda_2+\rho^2 \lambda_3+\rho^3\lambda_4,\label{P1-4th}\\
&&\hspace{0.7cm}  \rho^2(\rho\omega_1+\omega_2)= \rho\lambda_1+\lambda_2+\rho(\rho\omega_1+\omega_2)\lambda_3\nonumber\\&&\hspace{4cm}+\rho^2(\rho\omega_1+\omega_2)\lambda_4,\label{P2-4th}\\
&&\hspace{0.7cm} \rho (\rho^2\tau_1+\rho(\rho\omega_1+\omega_2)\tau_2+\tau_3)=\nonumber\\&&\hspace{2cm}\rho^2\lambda_1+\rho(\rho\omega_1+\omega_2)\lambda_2+\lambda_3+\rho(\rho^2\tau_1\nonumber\\&&\hspace{2cm}+\rho(\rho\omega_1+\omega_2)\tau_2+\tau_3)\lambda_4.\label{P3-4th}
\end{IEEEeqnarray}
\end{subequations}
We proceed with solving the above program. We write $\lambda_j=K_j+\delta_j\sqrt{2\epsilon\ln 2}$ for $j\in\{1,2,3\}$ and $\lambda_4=\delta_4\sqrt{2\epsilon\ln 2}$ and plug them into~\eqref{P1-4th}--\eqref{P3-4th} to get the following
\begin{IEEEeqnarray}{rCl}
\rho^3 &=& K_1+\rho K_2+\rho^2 K_3,\\
\rho^4 &=& \rho K_1+K_2+\rho^3 K_3,\\
\rho^5 &=& \rho^2K_1+\rho^3K_2+K_3.
\end{IEEEeqnarray}
Solving the above equations, we get $K_1=\rho^3$, $K_2=K_3=0$. Notice that the constant factors of $\{\lambda_{j}\}_{j=1}^3$ (i.e., $\{K_j\}_{j=1}^3$) contribute to the dominant terms of~\eqref{rate-4th-step} which simplifies to the following
\begin{IEEEeqnarray}{rCl}
\delta_4\leq 1.
\end{IEEEeqnarray}
So, the optimization problem in~\eqref{opt-AR-4th-program} with dominant terms simplifies to the following
\begin{subequations}
\begin{IEEEeqnarray}{rCl}
&&\min_{\delta_j, j=1:4} 2(1-\rho^6-(\delta_4+\rho^5\delta_3+\rho^4\delta_2+\rho^3\delta_1)\sqrt{2\epsilon\ln 2})\sigma^2\nonumber\\\\[1.5ex]
&&\text{s.t.}: \delta_4\leq 1,\\
&&\hspace{0.7cm} 0=\delta_1+\rho\delta_2+\rho^2\delta_3+\rho^3\delta_4,\\
&&\hspace{0.7cm}  \rho^2(1-\rho^2)= \rho\delta_1+\delta_2+\rho^3\delta_3+\rho^4\delta_4,\\
&&\hspace{0.7cm}\rho(1-\rho^4)= \rho^2\delta_1+\rho^3\delta_2+\delta_3+\rho^5\delta_4.
\end{IEEEeqnarray}
\end{subequations}
Solving the above optimization problem, we get 
\begin{IEEEeqnarray}{rCl}
\delta_2=\rho^2,\qquad \delta_3=\rho, \qquad \delta_1=-3\rho^3,\qquad \delta_4=1.\nonumber\\
\end{IEEEeqnarray}
In summary, we get the following reconstruction
\begin{IEEEeqnarray}{rCl}
\hat{X}_4&=&(\rho^3-3\rho^3\sqrt{2\epsilon\ln 2})\hat{X}_1+\rho^2\sqrt{2\epsilon\ln 2}\hat{X}_2\nonumber\\&&\hspace{0.5cm}+\rho\sqrt{2\epsilon\ln 2}\hat{X}_3+\sqrt{2\epsilon\ln 2}X_4+Z_{4,\text{SA}}.\nonumber\\
\end{IEEEeqnarray}
Plugging~\eqref{2nd-step-reconstruction} and~\eqref{CP-3rd-step-reconstruction} into the above expression, we get
\begin{IEEEeqnarray}{rCl}
\hat{X}_4&=&\rho^3\hat{X}_1+\rho^2\sqrt{2\epsilon\ln 2}N_1+\rho\sqrt{2\epsilon\ln 2}N_2+N_3\nonumber\\&&+\rho^2\sqrt{2\epsilon\ln 2}Z_{2,\text{SA}}+\rho\sqrt{2\epsilon\ln 2}Z_{3,\text{SA}}+Z_{4,\text{SA}},\nonumber\\
\end{IEEEeqnarray}
where $Z_{4,\text{SA}}$ has variance $(1-\rho^6+O(\epsilon))\sigma^2$ and the distortion is given by
\begin{IEEEeqnarray}{rCl}
D_{4,\text{SA}}^{\infty}=2(1-\rho^{6}-\sqrt{2\epsilon\ln 2}(1-\rho^6))\sigma^2+O(\epsilon).\nonumber\\
\end{IEEEeqnarray}
Now, we use an induction to derive the achievable reconstruction of $j$th frame. \\\\

\underline{\textit{$j$th Step}}:

Using induction and extension of the above analysis to $j$ frames, we get the following achievable reconstruction for $j$th frame
\begin{IEEEeqnarray}{rCl}
\hat{X}_j&=&\rho^{j-1}\hat{X}_1+\sqrt{2\epsilon\ln 2}\sum_{i=1}^{j-1}\rho^{j-1-i}N_i\nonumber\\&&\hspace{0.5cm}+\sqrt{2\epsilon\ln 2}\sum_{i=2}^{j-1}\rho^{j-i}Z_{i,\text{SA}}+Z_{j,\text{SA}},
\end{IEEEeqnarray}
where $Z_{j,\text{SA}}\sim \mathcal{N}(0,(1-\rho^{2(j-1)}+O(\epsilon))\sigma^2)$ is a Gaussian random variable independent of $(\hat{X}_1,\{N_i\}_{i=1}^{j-1},\{Z_{i,\text{SA}}\}_{i=2}^{j-1})$ and the distortion is given by
\begin{IEEEeqnarray}{rCl}
&&\hspace{-0.5cm}D_{j,\text{SA}}^{\infty}=\nonumber\\&&2\sigma^2(1-\rho^{2(j-1)}-\sqrt{2\epsilon\ln 2}(1-\rho^2)\sum_{i=1}^{j-1}\rho^{2(j-1-i)})\nonumber\\&&+O(\epsilon).
\end{IEEEeqnarray}

\subsection{$0$-PLF-JD}\label{JD-high-R1-low-R23} 

\underline{\textit{Second Step:}} When the first frame is compressed at a high rate, the optimization program of the second step for $0$-PLF-JD is similar to that in~\eqref{CP-2nd-opt} for $0$-PLF-SA and the solution is given in~\eqref{2nd-step-reconstruction}.

\underline{\textit{Third Step:}}

The optimization program of the third step for $0$-PLF-JD is similar to~\eqref{CP-3rd-opt} but when the perception constraints in~\eqref{perception1-R1inf}--\eqref{perception2-R1inf} are replaced by 
\begin{IEEEeqnarray}{rCl}
\rho^2&=&\tau_1+\tau_2\rho+\tau_3\rho^2,\\
\rho&=&\tau_1\rho+\tau_2+\tau_3(\omega_1\rho^2+\rho\omega_2).
\end{IEEEeqnarray}
The above equations come from the fact that $P_{X_1X_2X_3}=P_{\hat{X}_1\hat{X}_2\hat{X}_3}$ which implies that $\mathbbm{E}[\hat{X}_1\hat{X}_3]=\mathbbm{E}[X_1X_3]=\rho^2\sigma^2$ and $\mathbbm{E}[\hat{X}_2\hat{X}_3]=\mathbbm{E}[X_2X_3]=\rho\sigma^2$. Thus, we have the following optimization program for the third step of $0$-PLF-JD when the first frame is compressed at a high rate,
\begin{subequations}\label{JD-optimization-3rd-step-program}
\begin{IEEEeqnarray}{rCl}
&&\min_{\tau_1,\tau_2,\tau_3} 2\sigma^2-2\tau_3\sigma^2-2\tau_2\omega_2\rho\sigma^2-2\tau_2\omega_1\rho^2\sigma^2-2\tau_1\rho^2\sigma^2\nonumber\\\\
&&\text{s.t.}:\;\;\tau_3^2 (1-2^{-2R_3}(\rho^42^{-2R_1-2R_2}+\rho^2(1-\rho^2)2^{-2R_2}\nonumber\\&&\hspace{2cm}-\rho^2))\leq (1-2^{-2R_3})(1-\tau_1^2-\tau_2^2\nonumber\\&&\hspace{3cm} -2\tau_1\tau_2\omega_1-2\tau_1\tau_2\omega_2\rho\nonumber\\&&\hspace{3cm}-2\tau_2\tau_3\omega_1\rho^2-2\tau_2\tau_3\omega_2\rho-2\tau_1\tau_3\rho^2),\nonumber\\\label{R-JD-4th}\\
&&\hspace{1cm}\rho^2=\tau_1+\tau_2\rho+\tau_3\rho^2,\label{perception1-R1inf-JD}\\
&&\hspace{1cm}\rho=\tau_1\rho+\tau_2+\tau_3(\omega_1\rho^2+\rho\omega_2).\label{perception2-R1inf-JD}
\end{IEEEeqnarray}
\end{subequations}

\textbf{Case of $R_3\to \infty$}: In this case, the optimization problem in~\eqref{JD-optimization-3rd-step-program} simplifies to the following:
\begin{subequations}
\begin{IEEEeqnarray}{rCl}
&&\min_{\tau_1,\tau_2,\tau_3} 2\sigma^2-2\tau_3\sigma^2-2\tau_2\omega_2\rho\sigma^2-2\tau_2\omega_1\rho^2\sigma^2-2\tau_1\rho^2\sigma^2\nonumber\\\\
&&\text{s.t.}:\;\;\tau_3^2 \leq (1-\tau_1^2-\tau_2^2 -2\tau_1\tau_2\omega_1-2\tau_1\tau_2\omega_2\rho\nonumber\\&&\hspace{2cm}-2\tau_2\tau_3\omega_1\rho^2-2\tau_2\tau_3\omega_2\rho-2\tau_1\tau_3\rho^2),\label{rate-inf-eps-inf}\\
&&\hspace{1cm}\rho^2=\tau_1+\tau_2\rho+\tau_3\rho^2,\label{P-loss-inf1}\\
&&\hspace{1cm}\rho=\tau_1\rho+\tau_2+\tau_3(\omega_1\rho^2+\rho\omega_2).\label{P-loss-inf2}
\end{IEEEeqnarray}
\end{subequations}
We write $\tau_1$, $\tau_2$ and $\tau_3$ as $\tau_1=K_1+\delta_1\sqrt{2\epsilon\ln 2}$, $\tau_2=K_2+\delta_2\sqrt{2\epsilon\ln 2}$ and $\tau_3=K_3+\delta_3\sqrt{2\epsilon\ln 2}$, and plug them into~\eqref{P-loss-inf1}--\eqref{P-loss-inf2} to get
\begin{IEEEeqnarray}{rCl}
\rho^2&=& K_1+\rho K_2+\rho^2 K_3,\\
\rho&=& \rho K_1+K_2+\rho^3K_3,\\
0&=&\delta_1+\rho\delta_2+\rho^2\delta_3,\\
0&=&\rho \delta_1+\delta_2+K_3(\rho-\rho^3)+\delta_3\rho^3.
\end{IEEEeqnarray}
The above set of equations yields the following
\begin{IEEEeqnarray}{rCl}
K_2&=&\rho,\label{K2-inf-eps-inf}\\
K_1&=&-\rho^2 K_3,\label{K1-inf-eps-inf}\\
\delta_2&=&-\rho K_3.\label{D2-inf-eps-inf}
\end{IEEEeqnarray}
Also, the constraint in~\eqref{rate-inf-eps-inf} yields the following for the first-order terms:
\begin{IEEEeqnarray}{rCl}
K_3^2\leq 1-K_1^2-K_2^2-2K_1K_2\rho-2K_2K_3\rho^3-2K_1K_3\rho^2.\nonumber\\
\end{IEEEeqnarray}
Plugging~\eqref{K2-inf-eps-inf} and~\eqref{K1-inf-eps-inf} into the above equation and considering the fact the solution of optimization problem is given when the above inequality is satisfied with equality, we get
\begin{IEEEeqnarray}{rCl}
K_3=\frac{1}{\sqrt{1+\rho^2}},
\end{IEEEeqnarray}
and 
\begin{IEEEeqnarray}{rCl}
K_1=-\frac{\rho^2}{\sqrt{1+\rho^2}}.
\end{IEEEeqnarray}
Also, from~\eqref{D2-inf-eps-inf}, we get
\begin{IEEEeqnarray}{rCl}
\delta_2=-\frac{\rho}{\sqrt{1+\rho^2}}.
\end{IEEEeqnarray}
Notice that, in this optimization problem, all first-order terms (i.e., $K_1, K_2, K_3$) are non-zero, we can write the third reconstruction as follows
\begin{IEEEeqnarray}{rCl}
\hat{X}_3 &=& (-\frac{\rho^2}{\sqrt{1+\rho^2}}+O(\sqrt{\epsilon}))\hat{X}_1+(\rho-O(\sqrt{\epsilon}))\hat{X}_2\nonumber\\&&+ (\frac{1}{\sqrt{1+\rho^2}}+O(\sqrt{\epsilon}))X_3+Z_{3,\text{JD}},
\end{IEEEeqnarray}
where $Z_{3,\text{JD}}\sim \mathcal{N}(0,O(\sqrt{\epsilon})\sigma^2)$. Notice that since $\hat{X}_1=X_1$ due to high-rate, the above reconstruction can be further simplified as follows
\begin{IEEEeqnarray}{rCl}
\hat{X}_3&=&(\rho-O(\sqrt{\epsilon})) \hat{X}_2\nonumber\\&&+\frac{1}{\sqrt{1+\rho^2}}(\rho N_1+N_2+O(\sqrt{\epsilon}))+Z_{3,\text{JD}}.\nonumber\\
\end{IEEEeqnarray}
  
\textbf{Case of $R_3=R_2=\epsilon$}: Similar to~\eqref{opt-AR-4th-program}, we consider the dominant terms of the constraint in~\eqref{R-JD-4th} and get the following upper bound on the above optimization problem,
\begin{subequations}\label{4th-JD-program-asym}
\begin{IEEEeqnarray}{rCl}
&&\min_{\tau_1,\tau_2,\tau_3} 2\sigma^2-2\tau_3\sigma^2-2\tau_2\omega_2\rho\sigma^2-2\tau_2\omega_1\rho^2\sigma^2-2\tau_1\rho^2\sigma^2\nonumber\\\\
&&\text{s.t.}:\;\;(1-\rho^4)\tau_3^2\leq (1-\tau_1^2-\tau_2^2)(2\epsilon\ln 2),\label{rate-R1inf-JD-asym}\\
&&\hspace{1cm}\rho^2=\tau_1+\tau_2\rho+\tau_3\rho^2,\label{perception1-R1inf-JD-asym}\\
&&\hspace{1cm}\rho=\tau_1\rho+\tau_2+\tau_3(\omega_1\rho^2+\rho\omega_2).\label{perception2-R1inf-JD-asym}
\end{IEEEeqnarray}
\end{subequations}
We write $\tau_1$, $\tau_2$ and $\tau_3$ as $\tau_1=K_1+\delta_1\sqrt{2\epsilon\ln 2}$, $\tau_2=K_2+\delta_2\sqrt{2\epsilon\ln 2}$ and $\tau_3=\delta_3\sqrt{2\epsilon\ln 2}$, and plug them into~\eqref{perception1-R1inf-JD-asym}--\eqref{perception2-R1inf-JD-asym} to get the following equations
\begin{subequations}
\begin{IEEEeqnarray}{rCl}
\rho^2&=& K_1+\rho K_2,\label{equation1-JD}\\
\rho &=& K_1\rho + K_2,\label{equation2-JD}\\
0&=& \delta_1+\rho\delta_2+\rho^2\delta_3,\label{equation3-JD}\\
0&=&\rho\delta_1+\delta_2+\rho^3\delta_3.\label{equation4-JD}
\end{IEEEeqnarray}
\end{subequations}
Equations~\eqref{equation1-JD} and~\eqref{equation2-JD} yield $K_1=0$ and $K_2=\rho$. Notice that the constant terms of $\{\tau_j\}_{j=1}^2$ (i.e., $\{K_j\}_{j=1}^2$) contribute to the dominant terms of the inequality~\eqref{rate-R1inf-JD-asym}. Thus, we have the following condition
\begin{IEEEeqnarray}{rCl}
\delta_3\leq \frac{1}{\sqrt{1+\rho^2}}.
\end{IEEEeqnarray}
The optimization program in~\eqref{4th-JD-program-asym} further simplifies as follows
\begin{subequations}
\begin{IEEEeqnarray}{rCl}
&&\min_{\delta_1,\delta_2,\delta_3} 2\sigma^2(1-\rho^4\nonumber\\&&\hspace{2cm}-(\delta_3+\delta_1\rho^2+\delta_2\rho^3+\rho^2-\rho^4)\sqrt{2\epsilon\ln 2}),\nonumber\\\\
&&\text{s.t.}:\;\;\delta_3\leq \frac{1}{\sqrt{1+\rho^2}},\\
&&\hspace{1cm}0= \delta_1+\rho\delta_2+\rho^2\delta_3,\\
&&\hspace{1cm}0=\rho\delta_1+\delta_2+\rho^3\delta_3.
\end{IEEEeqnarray}
\end{subequations}
Solving the above optimization program, we get
\begin{IEEEeqnarray}{rCl}
\delta_2&=& 0,\;\;\delta_1= -\frac{\rho^2}{\sqrt{1+\rho^2}},\;\; \delta_3=\frac{1}{\sqrt{1+\rho^2}}.
\end{IEEEeqnarray}
Thus, we have
\begin{IEEEeqnarray}{rCl}
\hat{X}_3&=&\rho\hat{X}_2-\frac{\rho^2}{\sqrt{1+\rho^2}}\sqrt{2\epsilon\ln 2}\hat{X}_1\nonumber\\&&\hspace{0.5cm}+\frac{1}{\sqrt{1+\rho^2}}\sqrt{2\epsilon\ln 2}X_3+Z_{3,\text{JD}},
\end{IEEEeqnarray}
where $Z_{3,\text{JD}}\sim\mathcal{N}(0,(1-\rho^2+O(\epsilon))\sigma^2)$ is independent of $(\hat{X}_1,\hat{X}_2,X_3)$.
Plugging~\eqref{2nd-step-reconstruction} into the above expression yields the following
\begin{IEEEeqnarray}{rCl}
\hat{X}_3&=&\left(\rho^2-(\rho^2+\frac{\rho^2}{\sqrt{1+\rho^2}})\sqrt{2\epsilon\ln 2}\right)\hat{X}_1\nonumber\\&&+\rho\sqrt{2\epsilon\ln 2}X_2+\frac{\sqrt{2\epsilon\ln 2}}{\sqrt{1+\rho^2}}X_3+\rho Z_{2,\text{JD}}+Z_{3,\text{JD}},\nonumber\\\label{JD-3rd-step-reconstruction-app}
\end{IEEEeqnarray}
where the distortion is given as follows
\begin{IEEEeqnarray}{rCl}
D_{3,\text{JD}}^{\infty}&:=&2(1-\rho^4-(1-\rho^2)(\rho^2+\sqrt{1+\rho^2})\sqrt{2\epsilon\ln 2})\sigma^2\nonumber\\&&+O(\epsilon).
\end{IEEEeqnarray}
Using~\eqref{Gaus-def2}, \eqref{JD-3rd-step-reconstruction-app} can be further simplified as follows
\begin{IEEEeqnarray}{rCl}
\hat{X}_3&=&\rho^2\hat{X}_1+\left(\rho+\frac{\rho}{\sqrt{1+\rho^2}}\right)\sqrt{2\epsilon\ln2}N_1\nonumber\\&&+\frac{1}{\sqrt{1+\rho^2}}\sqrt{2\epsilon\ln2}N_2+\rho Z_{2,\text{JD}}+Z_{3,\text{JD}}.
\end{IEEEeqnarray}

\underline{\textit{Fourth Step:}}

The optimization program of the fourth step for $0$-PLF-JD is similar to that in Proposition~\ref{4th-step-AR-propostion} but when conditions~\eqref{P1-4th}--\eqref{P3-4th} are replaced by the corresponding conditions of $0$-PLF-JD which are 
\begin{IEEEeqnarray}{rCl}
\mathbbm{E}[\hat{X}_4\hat{X}_3]&=& \mathbbm{E}[X_4\hat{X}_3],\; \mathbbm{E}[\hat{X}_4\hat{X}_2]= \mathbbm{E}[X_4\hat{X}_2], \nonumber\\&& \mathbbm{E}[\hat{X}_4\hat{X}_1]= \mathbbm{E}[X_4\hat{X}_1].
\end{IEEEeqnarray}
The above conditions are further simplified as follows 
\begin{IEEEeqnarray}{rCl}
\rho^3 &=& \lambda_1+\rho \lambda_2+\rho^2 \lambda_3+\rho^3\lambda_4,\\
 \rho^2&=& \rho\lambda_1+\lambda_2+\rho\lambda_3+\rho^2(\rho\omega_1+\omega_2)\lambda_4,\\
\rho &=&\rho^2\lambda_1+\rho\lambda_2+\lambda_3+\rho(\rho^2\tau_1+\rho(\rho\omega_1+\omega_2)\tau_2\nonumber\\&&\hspace{4cm}+\tau_3)\lambda_4.
\end{IEEEeqnarray}
We study the case of $R_2=R_3=R_4=\epsilon$ for a sufficiently small $\epsilon>0$. Thus, considering the dominant terms, we have the following optimization problem for the fourth step of $0$-PLF-JD when the first frame is compressed at a high rate
\begin{subequations}\label{4th-JD-opt-asym}
\begin{IEEEeqnarray}{rCl}
&&\min_{\lambda_1,\lambda_2,\lambda_3,\lambda_4}2\sigma^2-2\lambda_4\sigma^2\nonumber\\&&\hspace{1.5cm}-2\lambda_3\rho\tau_3\sigma^2-2\lambda_3\rho^2\tau_2\omega_2\sigma^2-2\lambda_3\rho^3\tau_2\omega_1\sigma^2\nonumber\\[-1ex]&&\hspace{1.5cm}-2\lambda_3\rho^3\tau_1\sigma^2-2\lambda_2\rho^3\omega_1\sigma^2-2\lambda_2\rho^2\omega_2\sigma^2\nonumber\\&&\hspace{1.5cm}-2\lambda_1\rho^3\sigma^2\\[1.5ex]
&&\text{s.t.}: \lambda_4^2(1-\rho^6)\leq (1-\lambda_1^2-\lambda_2^2-\lambda_3^2+O(\epsilon))(2\epsilon\ln 2),\nonumber\\\label{R-4th-JD}\\
&& \hspace{0.6cm}\rho^3 = \lambda_1+\rho \lambda_2+\rho^2 \lambda_3+\rho^3\lambda_4,\label{P1-4th-JD}\\
&& \hspace{0.6cm}\rho^2= \rho\lambda_1+\lambda_2+\rho\lambda_3+\rho^2(\rho\omega_1+\omega_2)\lambda_4,\label{P2-4th-JD}\\
&&\hspace{0.6cm}\rho =\rho^2\lambda_1+\rho\lambda_2+\lambda_3+\rho(\rho^2\tau_1+\rho(\rho\omega_1+\omega_2)\tau_2\nonumber\\&&\hspace{5cm}+\tau_3)\lambda_4.\label{P3-4th-JD}
\end{IEEEeqnarray}
\end{subequations}
We proceed with solving the above optimization program. We write $\lambda_j=K_j+\delta_j\sqrt{2\epsilon\ln 2}$ for $j\in\{1,2,3\}$ and $\lambda_4=\delta_4\sqrt{2\epsilon\ln 2}$ and plug them into~\eqref{P1-4th-JD}--\eqref{P3-4th-JD} to get 
\begin{IEEEeqnarray}{rCl}
\rho^3&=& K_1+\rho K_2+\rho^2 K_3,\\
\rho^2 &=& \rho K_1+K_2+\rho K_3,\\
\rho &=& \rho^2 K_1+\rho K_2 +K_3,\\
0&=& \delta_1+\rho \delta_2+\rho^2 \delta_3+\rho^3\delta_4,\\
0&=& \rho\delta_1+\delta_2+\rho\delta_3+\rho^4\delta_4,\\
0&=&\rho^2\delta_1+\rho\delta_2+\delta_3+\rho^5\delta_4.
\end{IEEEeqnarray}
Thus, we have $K_1=K_2=0$, $K_3=\rho$. Considering the fact that the constant terms of $\{\lambda_j\}_{j=1}^3$ (i.e., $\{K_j\}_{j=1}^3$) contribute to the dominant terms of~\eqref{R-4th-JD} which simplifies to the following
\begin{IEEEeqnarray}{rCl}
\delta_4\leq \sqrt{\frac{1-\rho^2}{1-\rho^6}}.
\end{IEEEeqnarray}
The optimization program in~\eqref{4th-JD-opt-asym} further reduces to the following
\begin{subequations}
\begin{IEEEeqnarray}{rCl}
&&\min_{\delta_1,\delta_2,\delta_3,\delta_4}2(1-\rho^6-(\delta_1\rho^3+\delta_2\rho^4+\delta_3\rho^5+\delta_4+\rho^2\nonumber\\&&\hspace{2cm}-\rho^6)\sqrt{2\epsilon\ln 2})\sigma^2,\\
&&\text{s.t.}: \delta_4\leq \sqrt{\frac{1-\rho^2}{1-\rho^6}},\\
&& \hspace{0.7cm}0=\delta_1+\rho \delta_2+\rho^2 \delta_3+\rho^3\delta_4,\\
&& \hspace{0.7cm}0=\rho\delta_1+\delta_2+\rho\delta_3+\rho^4\delta_4,\\
&&\hspace{0.7cm}0=\rho^2\delta_1+\rho\delta_2+\delta_3+\rho^5\delta_4.
\end{IEEEeqnarray}
\end{subequations}
Solving the above optimization program, we get
\begin{IEEEeqnarray}{rCl}
\delta_1=-\rho^3\sqrt{\frac{1-\rho^2}{1-\rho^6}},\;\;\; \delta_2=\delta_3=0,\;\;\; \delta_4=\sqrt{\frac{1-\rho^2}{1-\rho^6}}.\nonumber\\
\end{IEEEeqnarray}
In summary, we get the following achievable reconstruction
\begin{IEEEeqnarray}{rCl}
\hat{X}_4&=&-\rho^3\sqrt{\frac{1-\rho^2}{1-\rho^6}}\sqrt{2\epsilon\ln 2}\hat{X}_1+\rho \hat{X}_3\nonumber\\&&\hspace{0.5cm}+\sqrt{\frac{1-\rho^2}{1-\rho^6}}\sqrt{2\epsilon\ln 2}X_4+Z_{4,\text{JD}},
\end{IEEEeqnarray}
where $Z_{4,\text{JD}}\sim \mathcal{N}(0,(1-\rho^2+\rho^4-\rho^6+O(\epsilon))\sigma^2)$ is a Gaussian random variable independent of $(\hat{X}_1,\hat{X}_3,X_4)$.
Now, we plug~\eqref{2nd-step-reconstruction} and~\eqref{CP-3rd-step-reconstruction} into the above expression and we get
\begin{IEEEeqnarray}{rCl}
\hat{X}_4&=&\rho^3\hat{X}_1+\left(\rho^2+\rho^2\sqrt{\frac{1-\rho^2}{1-\rho^6}}\right)\sqrt{2\epsilon\ln 2}N_1\nonumber\\&&\hspace{0.5cm}+\left(\rho+\rho\sqrt{\frac{1-\rho^2}{1-\rho^6}}\right)\sqrt{2\epsilon\ln 2}N_2\nonumber\\&&\hspace{0.5cm}+\sqrt{\frac{1-\rho^2}{1-\rho^6}}\sqrt{2\epsilon\ln 2}N_3+\rho^2\sqrt{2\epsilon\ln 2}Z_{2,\text{JD}}\nonumber\\&&\hspace{0.5cm}+\rho Z_{3,\text{JD}}+Z_{4,\text{JD}},
\end{IEEEeqnarray}
where the distortion is given by
\begin{IEEEeqnarray}{rCl}
D_{4,\text{JD}}^{\infty}&:=&2\sigma^2\Bigg(1-\rho^{6}\nonumber\\&&\hspace{0.1cm}-\sqrt{2\epsilon\ln 2}(1-\rho^2)\left(\sqrt{\frac{1-\rho^{6}}{1-\rho^2}}+\rho^2-\rho^6\right)\Bigg)\nonumber\\&&\hspace{0.1cm}+O(\epsilon).
\end{IEEEeqnarray}

\underline{\textit{$j$th Step:}}

Using induction and extension of the above analysis for the $j$-th frame yields the following achievable reconstruction
\begin{IEEEeqnarray}{rCl}
\hat{X}_j&=& \rho^{j-1}\hat{X}_1\nonumber\\&&\hspace{0.5cm}+\sqrt{2\epsilon\ln 2}\left(1+\sqrt{\frac{1-\rho^2}{1-\rho^{2(j-1)}}}\right)\sum_{i=1}^{j-2}\rho^{j-1-i}N_i\nonumber\\&&\hspace{0.5cm}+\sqrt{\frac{1-\rho^2}{1-\rho^{2(j-1)}}}\sqrt{2\epsilon\ln 2}N_{j-1}\nonumber\\&&\hspace{0.5cm}+\sqrt{2\epsilon\ln 2}\sum_{i=2}^{j-2}\rho^iZ_{j-i,\text{JD}}+\rho Z_{j-1,\text{JD}}+Z_{j,\text{JD}},\nonumber\\
\end{IEEEeqnarray}
where $Z_{j,\text{JD}}$ is a Gaussian random variable independent of $(\{N_i\}_{i=1}^{j-1},\{Z_{i,\text{JD}}\}_{i=2}^{j-1})$ with mean zero and the following variance
\begin{IEEEeqnarray}{rCl}
&&\mathbbm{E}[Z_{j,\text{JD}}^2]\nonumber\\&&=\left\{\begin{array}{ll} ((1-\rho^2)\sum_{i=0}^{\frac{j}{2}-1}\rho^{4i}+O(\epsilon))\sigma^2 & \text{if}\;j\;\text{is even},\\((1-\rho^2)\sum_{i=0}^{\frac{j-1}{2}-1}\rho^{4i}+O(\epsilon))\sigma^2 & \text{if}\;j\;\text{is odd},\end{array}\right. \nonumber\\
\end{IEEEeqnarray}
and the distortion is given by
\begin{IEEEeqnarray}{rCl}
D_{j,\text{JD}}^{\infty}&:=&2\sigma^2\Bigg(1-\rho^{2(j-1)}\nonumber\\&&\hspace{-1.2cm}-\sqrt{2\epsilon\ln 2}(1-\rho^2)\left(\sqrt{\frac{1-\rho^{2(j-1)}}{1-\rho^2}}+\sum_{i=1}^{j-2}\rho^{2(j-1-i)}\right)\Bigg)\nonumber\\&&+O(\epsilon).\nonumber\\
\end{IEEEeqnarray}

\begin{table*}[ht]
\centering
\caption{Achievable reconstructions and distortions for $R_1\to\infty$ and $R_2=R_3=\epsilon$. }
\label{table-ach-recons}
\begin{center}
\begin{tiny}
\begin{sc}
\begin{minipage}{\linewidth}
\begin{tabular}{|l|l|l|}
\hline
& Second Step  & Third Step  \\
\hline 
$0$-PLF-FMD & $\hat{X}_2=(1-O(\epsilon))\hat{X}_1+O(\epsilon)X_2+Z_{2,\text{FMD}}$ & $\hat{X}_3= (1-O(\epsilon))\hat{X}_1+O(\epsilon)X_2+O(\epsilon)X_3+Z_{3,\text{FMD}}$\\
(\!$\sqrt{\epsilon}\!\ll\!\! \rho\!\! <\! 1$\!)& $Z_{2,\text{FMD}}\sim\mathcal{N}(0,O(\epsilon)\sigma^2)$ & $Z_{3,\text{FMD}}\sim\mathcal{N}(0,O(\epsilon)\sigma^2)$\\
& $D_{2,\text{FMD}}^{\infty}=2(1-\rho-O(\epsilon))\sigma^2$ Table 2 in \cite{Jun-Ashish2023} & $D_{3,\text{FMD}}^{\infty}=2(1-\rho^2-O(\epsilon))\sigma^2$ \qquad (Appendix~\ref{FMd-high-R1-low-R23})\\
\hline 
$0$-PLF-FMD & $\hat{X}_2= O(\sqrt{\epsilon})X_2+Z'_{2,\text{FMD}}$ & $\hat{X}_3= O(\sqrt{\epsilon})X_3+Z'_{3,\text{FMD}}$\\
($0\!<\!\! \rho\!\! \ll \! \sqrt{\epsilon}$) & $Z'_{2,\text{FMD}}\sim\mathcal{N}(0,(1-O(\epsilon))\sigma^2)$ & $Z'_{3,\text{FMD}}\sim\mathcal{N}(0,(1-O(\epsilon))\sigma^2)$\\
& $D_{2,\text{FMD}}^{\infty}=2\sigma^2(1-O(\sqrt{\epsilon}))$\;\;\; (Appendix~\ref{FMd-high-R1-low-R23}) & $D_{3,\text{FMD}}^{\infty}=2\sigma^2(1-O(\sqrt{\epsilon}))$  \qquad\qquad (Appendix~\ref{FMd-high-R1-low-R23})\\
\hline
$0$-PLF-JD & $\hat{X}_2=(\rho-O(\sqrt{\epsilon}))\hat{X}_1+O(\sqrt{\epsilon})X_2+Z_{2,\text{JD}}$ & $\hat{X}_3\!=\!\rho^2\hat{X}_1+O(\sqrt{\epsilon})N_1+O(\sqrt{\epsilon}) N_2+\rho Z_{2,\text{JD}}+Z_{3,\text{JD}}$\\
&$Z_{2,\text{JD}}\sim \mathcal{N}(0,(1-\rho^2+O(\epsilon))\sigma^2)$ & $Z_{3,\text{JD}}\sim \mathcal{N}(0,(1-\rho^2+O(\epsilon))\sigma^2)$\\
&$D_{2,\text{JD}}^{\infty}= 2\sigma^2(1-\rho^2-O(\sqrt{\epsilon}))$  Table 2 in \cite{Jun-Ashish2023} & $D_{3,\text{JD}}^{\infty}= 2\sigma^2(1-\rho^4-O(\sqrt{\epsilon}))$\qquad (Appendix~\ref{JD-high-R1-low-R23})\\
\hline
$0$-PLF-SA  & $\hat{X}_2=(\rho-O(\sqrt{\epsilon}))\hat{X}_1+O(\sqrt{\epsilon})X_2+Z_{2,\text{SA}}$ & $\hat{X}_3\!=\!\rho^2\hat{X}_1+O(\sqrt{\epsilon})N_1+O(\sqrt{\epsilon}) N_2+O(\sqrt{\epsilon})Z_{2,\text{SA}}+Z_{3,\text{SA}}$\\
& $Z_{2,\text{SA}}=Z_{2,\text{JD}}$ & $Z_{3,\text{SA}}\sim \mathcal{N}(0,(1-\rho^4+O(\epsilon))\sigma^2)$\\
& $D_{2,\text{SA}}^{\infty}= D^{\infty}_{2,\text{JD}}$\qquad\qquad\qquad\;\;\; (Appendix~\ref{CP-high-R1-low-R23}) & $D_{3,\text{SA}}^{\infty}= 2\sigma^2(1-\rho^4-O(\sqrt{\epsilon}))$ \qquad (Appendix~\ref{CP-high-R1-low-R23})\\
\hline
\end{tabular}
\end{minipage}
\end{sc}
\end{tiny}
\end{center}
\vskip -0.1in
\end{table*}

\subsection{$0$-PLF-FMD}\label{FMd-high-R1-low-R23} 

In this section, we provide the optimization programs for the second and third steps of $0$-PLF-FMD and solve them. These results were presented in the first and second rows of Table~\ref{table-ach-recons}. Recall that for the Gauss-Markov source model, the reconstructions exploit the structure in~\eqref{X1hat-Gaussian-ach}--\eqref{X3hat-Gaussian-ach}. 

\underline{\textit{Second Step:}}

For the second step, similar to~\eqref{X2hat-Gaussian-ach}, we write the achievable reconstruction as 
\begin{IEEEeqnarray}{rCl}
\hat{X}_2=\omega_1\hat{X}_1+\omega_2X_2+Z_{2,\text{FMD}},
\end{IEEEeqnarray} 
where $Z_{2,\text{FMD}}$ is independent of $(\hat{X}_1,X_2)$ and notice that $\hat{X}_1=X_1$ since we have high compression rate for the first frame. The optimization program of the second step is similar to that of Proposition~\ref{0-PLF-AR-2nd-step}, but with $\nu=1$ and when the perception constraint in~\eqref{perception-derivation} (which preserves the joint distribution of $(\hat{X}_1,\hat{X}_2)$) is removed and only the marginal distribution is fixed. Thus, we have the following optimization program for the second step of $0$-PLF-FMD 
\begin{subequations}\label{FMD-high-rate-optimization-program}
\begin{IEEEeqnarray}{rCl}
&&\hspace{0cm}\min_{\substack{\omega_1,\omega_2}}\;\; 2\sigma^2-2 \omega_1\rho\sigma^2-2\omega_2\sigma^2,\\
&&\text{s.t.}\; \omega_2^2(1-\rho^22^{-2R_2})\leq (1-\omega_1^2-2\omega_1\omega_2\rho)(1-2^{-2R_2}).\nonumber\\
\end{IEEEeqnarray}
\end{subequations}
The solution of the above program when $R_2=\epsilon$ (for a sufficiently small $\epsilon$) is given by (see Table 2 in \cite{Jun-Ashish2023})
\begin{IEEEeqnarray}{rCl}
\hat{X}_2=(1-\frac{(1+\rho^2)2\epsilon\ln 2}{2\rho^2})\hat{X}_1+\frac{2\epsilon\ln 2}{\rho}X_2+Z_{2,\text{FMD}},\nonumber\\\label{FMD-high-rate-a}
\end{IEEEeqnarray}
where $Z_{2,\text{FMD}}\sim\mathcal{N}(0,(\frac{1-\rho^2}{\rho^2})2\sigma^2\epsilon\ln 2 )$ is independent of $(\hat{X}_1,X_2)$.

Notice that when $\rho=\Theta(\sqrt{\epsilon})$, the term $\frac{(1+\rho^2)2\epsilon\ln 2}{2\rho^2}$ becomes a constant. In this case, the approximation in~\eqref{FMD-high-rate-a} is not valid anymore. This case should be handled separately as follows.

\textbf{Case of $0<\rho\ll \sqrt{\epsilon}$:} In this case, considering the dominant terms of~\eqref{FMD-high-rate-optimization-program}, this program reduces to the following
\begin{subequations}
\begin{IEEEeqnarray}{rCl}
&&\hspace{0cm}\min_{\substack{\omega_1,\omega_2}}\;\; 2\sigma^2-2\omega_2\sigma^2,\\
&&\text{s.t.}\qquad \omega_2^2\leq (1-\omega_1^2)(2\epsilon\ln 2).
\end{IEEEeqnarray}
\end{subequations}
The solution of the above program is as follows
\begin{IEEEeqnarray}{rCl}
\omega_1&=& 0,\\
\omega_2&=&\sqrt{2\epsilon\ln 2}. 
\end{IEEEeqnarray}
Thus, the reconstruction of the second step can be written as follows
\begin{IEEEeqnarray}{rCl}
\hat{X}_2=\sqrt{2\epsilon\ln 2}X_2+Z'_{2,\text{FMD}},\label{FMD-high-rate}
\end{IEEEeqnarray}
where $Z'_{2,\text{FMD}}\sim\mathcal{N}(0,(1-2\epsilon\ln 2)\sigma^2)$ is independent of $X_2$. 

\underline{\textit{Third Step:}}

For the third step, similar to~\eqref{X3hat-Gaussian-ach}, we write the achievable reconstruction as 
\begin{IEEEeqnarray}{rCl}
\hat{X}_3=\tau_1\hat{X}_1+\tau_2\hat{X}_2+\tau_3X_3+Z_{3,\text{FMD}},
\end{IEEEeqnarray} 
where $Z_{3,\text{FMD}}$ is a Gaussian random variable independent of $(\hat{X}_1,\hat{X}_2,X_3)$. 
The optimization program of the third step is similar to that of Proposition~\ref{3rd-step-prop-AR} but with $\nu=1$ and when the constraints in~\eqref{perception_3rd_frame} and~\eqref{perception_3rd_frameb} which preserve the joint distribution of $P_{\hat{X}_1\hat{X}_2\hat{X}_3}$ are removed and only the marginal distributions are fixed. Thus, we get the following optimization program
\begin{subequations}
\begin{IEEEeqnarray}{rCl}
&&\min_{\tau_1,\tau_2,\tau_3} 2\sigma^2-2\tau_3\sigma^2-2\tau_2\omega_2\rho\sigma^2-2\tau_2\omega_1\rho^2\sigma^2-2\tau_1\rho^2\sigma^2\nonumber\\\\
&&\text{s.t.}:\;\;\tau_3^2\sigma^2 (1-2^{-2R_3}(\rho^42^{-2R_1-2R_2}+\rho^2(1-\rho^2)2^{-2R_2}\nonumber\\&&\hspace{1cm}-\rho^2))\leq (1-2^{-2R_3})(1-\tau_1^2-\tau_2^2-2\tau_1\tau_2\omega_1\nonumber\\&&\hspace{3cm} -2\tau_1\tau_2\omega_2\rho-2\tau_2\tau_3\omega_1\rho^2\nonumber\\&&\hspace{3cm}-2\tau_2\tau_3\omega_2\rho-2\tau_1\tau_3\rho^2)\sigma^2.\label{R-3rd-high-rate-FMD}
\end{IEEEeqnarray}
\end{subequations}

\textbf{Case of $R_3\to \infty$}: In this case, the solution of the optimization is trivially given by $\hat{X}_3=X_3$ since it satisfies the $0$-PLF-FMD condition in the third frame which is $P_{\hat{X}_3}=P_{X_3}$.

\textbf{Case of $R_3=R_2=\epsilon$}:
We use the following approximation
\begin{IEEEeqnarray}{rCl}
1-2^{-2R_j}=2\epsilon\ln 2 +O(\epsilon^2),\qquad j\in\{2,3\}. 
\end{IEEEeqnarray}
Thus, considering the dominant terms of the constraint in~\eqref{R-3rd-high-rate-FMD}, we have
\begin{IEEEeqnarray}{rCl}
&&(1-\tau_1^2-\tau_2^2-2\tau_1\tau_2\omega_1-2\tau_1\tau_2\omega_2\rho\nonumber\\&&\hspace{0.5cm}-2\tau_2\tau_3\omega_1\rho^2-2\tau_2\tau_3\omega_2\rho-2\tau_1\tau_3\rho^2)(2\epsilon\ln 2)\nonumber\\&&\hspace{0.5cm}\geq (1-\rho^4)\tau_3^2.\label{rate-3rd-frame}
\end{IEEEeqnarray}
For the third frame, we have the following optimization program,
\begin{subequations}\label{opt-3rd-FMD-high}
\begin{IEEEeqnarray}{rCl}
&&\min_{\tau_1,\tau_2,\tau_3}2\sigma^2-2\tau_3\sigma^2-2\tau_2\omega_2\rho\sigma^2-2\tau_2\omega_1\rho^2\sigma^2-2\tau_1\rho^2\sigma^2,\nonumber\\\label{R1-inf-T-frame}\\
&&\text{s.t.}\;(1-\tau_1^2-\tau_2^2-2\tau_1\tau_2\omega_1-2\tau_1\tau_2\omega_2\rho\nonumber\\&&\hspace{0.7cm}-2\tau_2\tau_3\omega_1\rho^2-2\tau_2\tau_3\omega_2\rho-2\tau_1\tau_3\rho^2)(2\epsilon\ln 2)\nonumber\\&&\hspace{1cm}\geq (1-\rho^4)\tau_3^2.
\end{IEEEeqnarray}
\end{subequations}
We write $\tau_1$ and $\tau_2$ as follows
\begin{IEEEeqnarray}{rCl}
\tau_1 &=&\frac{1}{2}-\delta_1(2\epsilon\ln 2),\\
\tau_2 &=& \frac{1}{2}-\delta_2(2\epsilon\ln 2),\\
\tau_3 &=& \delta_3 (2\epsilon\ln 2).
\end{IEEEeqnarray}
for some $\delta_1$, $\delta_2$ and $\delta_3$. Plugging the above into \eqref{rate-3rd-frame}, we have
\begin{IEEEeqnarray}{rCl}
(3\delta_1+3\delta_2-2\delta_3\rho^2-\frac{1}{4}+\frac{1}{4\rho^2})\geq(1-\rho^4)\delta_3^2.\label{rate3-simplified}
\end{IEEEeqnarray}
Thus, the optimization program in~\eqref{opt-3rd-FMD-high} reduces to the following
\begin{IEEEeqnarray}{rCl}
&&\min_{\delta_1,\delta_2,\delta_3}\;\;2\sigma^2-2\rho^2\sigma^2\nonumber\\&&\hspace{0.7cm}-(2\delta_3+1-2(\delta_1+\delta_2)\rho^2-\frac{1-\rho^2}{2})(2\epsilon\ln 2)\nonumber\\\\
&&\text{s.t.}\;\;(3\delta_1+3\delta_2-2\delta_3\rho^2-\frac{1}{4}+\frac{1}{4\rho^2})\geq(1-\rho^4)\delta_3^2.\nonumber\\\label{D3-simplified_new}
\end{IEEEeqnarray}
Optimizing over $\delta_1,\delta_2, \delta_3$, we get
\begin{IEEEeqnarray}{rCl}
\delta_3=\frac{1-\frac{2}{3}\rho^4}{\frac{2}{3}\rho^2(1-\rho^4)},
\end{IEEEeqnarray}
and 
\begin{IEEEeqnarray}{rCl}
\delta_1=\delta_2=\frac{3-4\rho^8}{8\rho^4(1-\rho^4)}+\frac{1-\rho^2}{24\rho^2}.
\end{IEEEeqnarray}
Thus, we have
\begin{IEEEeqnarray}{rCl}
\hat{X}_3&=&(\frac{1}{2}-\delta_1(2\epsilon\ln 2))\hat{X}_1+(\frac{1}{2}-\delta_1(2\epsilon\ln 2))\hat{X}_2\nonumber\\&&\hspace{0.5cm}+\delta_3(2\epsilon\ln 2)X_3 +Z_{3,\text{FMD}},
\end{IEEEeqnarray}
where $Z_{3,\text{FMD}}\sim \mathcal{N}(0,O(\epsilon)\sigma^2)$ is independent of $(\hat{X}_1,\hat{X}_2,X_3)$, where the optimal distortion is given by 
\begin{IEEEeqnarray}{rCl}
&&D_{3,\text{FMD}}^{\infty}:=\nonumber\\&&\hspace{0.2cm}2\left(1-\rho^2-\left(\delta_3+\frac{1-\rho^2}{4}-(\delta_1+\delta_2)\rho^2\right)2\epsilon\ln 2\right)\sigma^2\nonumber\\&&\hspace{0.2cm}+O(\epsilon^2).
\end{IEEEeqnarray}

\textbf{Case of $0<\rho\ll \sqrt{\epsilon}$:} In this case, considering the dominant terms of~\eqref{opt-3rd-FMD-high}, the program reduces to the following: 
\begin{subequations}
\begin{IEEEeqnarray}{rCl}
&&\min_{\tau_1,\tau_2,\tau_3}\;\;2\sigma^2-2\tau_3\sigma^2,\\
&&\text{s.t.}\;(1-\tau_1^2-\tau_2^2)(2\epsilon\ln 2)\geq \tau_3^2.
\end{IEEEeqnarray}
\end{subequations}
The solution of the above program is simply given by
\begin{IEEEeqnarray}{rCl}
\tau_1&=& 0,\\
\tau_2 &=& 0,\\
\tau_3 &=& \sqrt{2\epsilon\ln 2}.
\end{IEEEeqnarray}
Thus, the reconstruction is given by 
\begin{IEEEeqnarray}{rCl}
\hat{X}_3=\sqrt{2\epsilon\ln 2}X_3+Z'_{3,\text{FMD}},
\end{IEEEeqnarray}
where $Z'_{3,\text{FMD}}\sim\mathcal{N}(0,(1-2\epsilon\ln 2)\sigma^2)$ is independent of $X_3$.

The achievable reconstructions derived in this section are summarized in Table~\ref{table-ach-recons} for the first three frames.

\section{Experimental Setup Details}\label{experiment-app}

As described in Section~\ref{experimental}, our experimental setup is based on the one proposed in \cite{Jun-Ashish2023}. We briefly describe our setup as follows.

\textit{Neural Video Compressor. } In this work, we use the version of the scale-space flow model \cite{agustsson2020scale} presented in \cite{Jun-Ashish2023} to compress each P-frame. This architecture allows us to efficiently learn the statistical
characteristics of the source distribution without using any pre-trained module such as an optical flow estimator. To control the bit rate, we adjust the dimension of the latent representation while fixing the quantization interval to $2$. We use dithered quantization to simulate the common randomness in our setting~\cite{Jun-Ashish2021}. For each frame $X_j$, we optimize its corresponding encoder-decoder by using the representation from the optimized encoder-decoder pairs of previous frames.

\textit{Distortion and Perception Measurement. } Our theoretical results require solving a constrained optimization, which is intractable in practice due to the complexity of neural networks. Instead, we optimize the Lagrange approximations:
$$\min \mathbbm{E}[\|X_j-\hat{X}_j\|^2] + \lambda \phi_j(P_{\hat{X}_{1}\ldots\hat{X}_{j-1} X_{j}},P_{\hat{X}_{1}\ldots\hat{X}_{j-1}\hat{X}_{j}}), $$
where each $\lambda$ is adjusted to characterize different constraint levels on the perceptuality. Similar to previous works, we use WGAN \cite{gulrajani2017improved} to approximate this perception function. 

\textit{Training Details. }  
The neural architectures tested on UVG are trained on $256\times256$ patches from the Vimeo-90K dataset \cite{xue2019video}. 
For each MNIST encoder-decoder pair, training takes about one day on a single NVIDIA A100 GPU, with Vimeo-90K training procedures taking around two days. 
For each rate regime, we first pre-train a model to optimize the MMSE loss before fine-tuning the model with the joint distortion-perception loss, which we found to be more stable than training everything end-to-end. We utilize the \textit{rmsprop} optimizer \cite{graves2014generating} for our MovingMNIST training procedures and the \textit{Adam} optimizer \cite{kingma2017adam} for Vimeo-90K training runs.

As described in Section~\ref{sec:exp_details}, DCVC-HEM is adopted for comparisons. 
For the MovingMNIST dataset, since DCVC-HEM is not specifically trained for low-bitrate scenarios ($R_1 = \epsilon$) on this dataset, we fine-tune the pre-trained DCVC-HEM to ensure a fair comparison. 
During fine-tuning, the quantization scales are adjusted to $\{q_{\text{I-frame}} = 3.5, q_{\text{P-frame}} = 1.5\}$ to enhance its compression performance under low-bitrate settings. 
In the high-bitrate scenario ($R_1=\infty$) on MovingMNIST, where the bitrate for second frame is fixed at $R_2 = 2$ bits across all PLF models, achieving this bitrate with DCVC-HEM is challenging. 
To address this, we directly input the second-frame reconstruction results from $0$-PLF-SA into DCVC-HEM to produce the reconstruction of the third frame.
For UVG dataset, we use the pre-trained DCVC-HEM checkpoint without additional fine-tuning. 
All results presented in Section~\ref{experimental} ensure that the average per-frame bitrate of DCVC-HEM is slightly greater than or equal to the bitrate settings of the proposed PLF-SA models.

\textit{Perceptual Quality Evaluations}
To evaluate the perceptual quality of different compressors, we use the widely adopted LPIPS~\cite{zhang2018unreasonable}, computed as $\| f(X_i) - f(\hat{X}_i) \|_2^2$ where $f(\cdot)$ is a pretrained, fixed deep network.
For the UVG dataset, we follow standard setting and use an ImageNet-pretrained VGG net as the feature extractor. 
For MovingMNIST, the domain gap between ImageNet and MNIST datasets causes the VGG net to overlook meaningful feature differences, even when digit identities change. 
To address this, we train two $4$-layer convolutional networks on the second and third frames of MovingMNIST, each for 10 epochs using the \textit{Adam} optimizer. 
Once trained, these models serve as feature extractors and we compute LPIPS metric between embeddings of source and reconstructed frames on MovingMNIST.

\end{document}